\documentclass[Afour,sageh,times]{sagej}

\usepackage{moreverb}
\usepackage[hyphens]{url}
\usepackage[colorlinks,breaklinks,bookmarksopen,bookmarksnumbered,citecolor=red,urlcolor=red]{hyperref}

\newcommand\BibTeX{{\rmfamily B\kern-.05em \textsc{i\kern-.025em b}\kern-.08em
T\kern-.1667em\lower.7ex\hbox{E}\kern-.125emX}}

\usepackage{graphics} 
\usepackage{epsfig} 
\usepackage{amsmath} 
\usepackage{silence}
\usepackage{svg}
\usepackage{amssymb}  
\usepackage{pifont}
\usepackage{multirow}
\usepackage{booktabs}
\usepackage{tabularx}
\usepackage{comment}
\usepackage{lipsum} 
\usepackage{graphicx,subcaption}
\usepackage[font={footnotesize,it}]{caption}

\usepackage{algorithm}

\usepackage[noend]{algpseudocode}

\definecolor{blue}{RGB}{0,0,0}
\definecolor{blue1}{RGB}{0,0,0}

\usepackage{fancyhdr}
\usepackage{ragged2e}
\fancypagestyle{custom_notice}
{
    \setlength{\headheight}{32pt}
    \addtolength{\topmargin}{-5pt}
    \fancyhead[L]{\fbox{\begin{minipage}{\textwidth}
    \noindent\footnotesize\justifying
    © 2025 SAGE. This paper is accepted to IJRR and will be published with open access. Personal use of this material is permitted. Permission from SAGE must be obtained for all other uses, in any current or future media, including reprinting/republishing this material for advertising or promotional purposes, creating new collective works, resale, redistribution to servers or lists, or reuse of any copyrighted component of this work in other works.
    \end{minipage}}}
    \fancyfoot{} 
}

\makeatletter
\newcommand{\thickhline}{%
    \noalign {\ifnum 0=`}\fi \hrule height 1pt
    \futurelet \reserved@a \@xhline
}
\newcolumntype{"}{@{\hskip\tabcolsep\vrule width 1pt\hskip\tabcolsep}}
\makeatother

\def\volumeyear{2024}
\begin{document}

\runninghead{A. Waltersson and Karayiannidis}

\title{Perception, Control and Hardware for In-Hand Slip-Aware Object Manipulation with Parallel Grippers}

\author{Gabriel Arslan Waltersson \affilnum{1} and Yiannis Karayiannidis \affilnum{2}}

\affiliation{\affilnum{1} Department of Electrical Engineering, Chalmers University of Technology, Sweden\\
\affilnum{2}  Department of Automatic Control, Lund University, Sweden. The author is a member of the ELLIIT Strategic Research Area at Lund University. }

\corrauth{Gabriel Arslan Waltersson, Chalmers University of Technology
 Department of Electrical Engineering,
 SE-412 96 Gothenburg, Sweden.}

\email{gabwal@chalmers.se}

\begin{abstract}
Dexterous in-hand manipulation offers significant potential to enhance robotic manipulator capabilities. 
\textcolor{blue}{This paper presents a sensori-motor \textcolor{blue1}{architecture} for in-hand slip-aware control, being embodied in a sensorized gripper.}
The gripper \textcolor{blue}{in our \textcolor{blue1}{architecture}} features rapid closed-loop, low-level force control, and is equipped with sensors capable of independently measuring contact forces and sliding velocities. Our system can quickly estimate essential object properties during pick-up using only in-hand sensing, without relying on prior object information. We introduce four distinct slippage controllers:  gravity-assisted trajectory following for both rotational and linear slippage, a hinge controller that maintains the object's orientation while the gripper rotates, and a slip-avoidance controller. The \textcolor{blue}{gripper} is mounted on a robot arm and validated through extensive experiments involving a diverse range of objects, demonstrating \textcolor{blue}{the \textcolor{blue1}{architecture's}} novel capabilities \textcolor{blue1}{for manipulating objects with flat surfaces.} 
\end{abstract}

\keywords{In-Hand Manipulation, Manipulation and Grasping, Grippers, Sensor-Based Control, Contact Modelling}

\maketitle
\thispagestyle{custom_notice}

\section{Introduction}
\begin{figure}
    \centering
    \smallskip 
    \includegraphics[width=0.65\columnwidth]{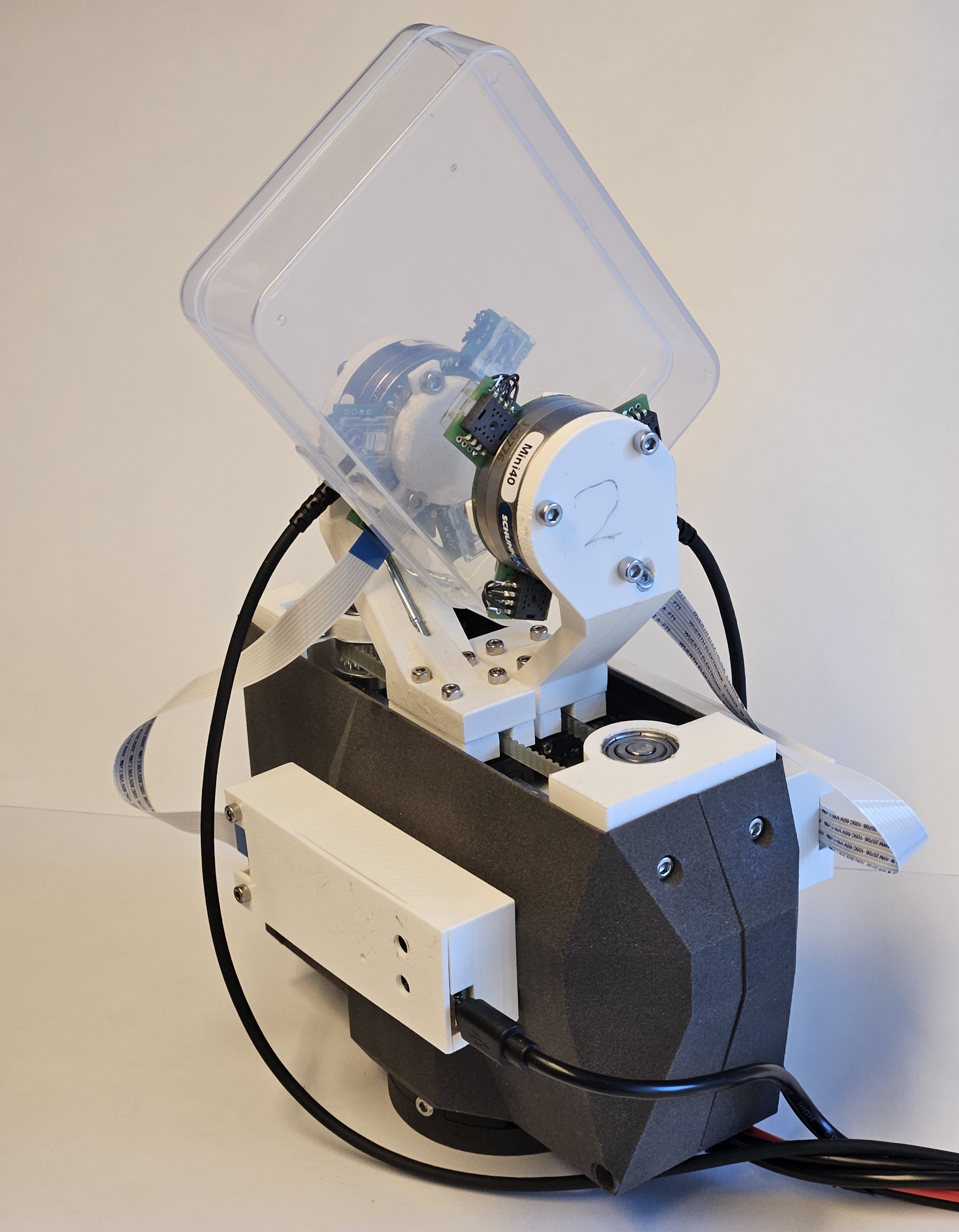}
    \caption{Picture of the custom gripper mounted with F/T and relative velocity sensors.}
    \label{fig:gripper}
    \vspace*{-0.5cm}
\end{figure}

Humans have the remarkable ability to pick up unfamiliar objects and quickly understand their surface properties, such as friction, and dynamics. This knowledge enables us not only to reorient objects using our arms but also to manipulate them within our hands, extending our capabilities beyond what is typically seen in traditional robotics. \textcolor{blue}{In this paper, we introduce a sensorized parallel gripper (see Fig. \ref{fig:gripper}), for in-hand slip-aware control that relies solely on in-hand sensing. 
This work is the first to combine both planar velocity and contact forces from independent in-hand sensing modalities for the purpose of slip-aware control in a parallel gripper, introducing new opportunities for intricate robotic manipulation. 
Each finger is equipped with a commercial 6-degree-of-freedom (DoF) force-torque (F/T) sensor and a custom relative velocity sensor.} This hardware combination enables rapid estimation of friction and contact surface properties without the need for external sensors, thus facilitating for precise in-hand manipulation of objects \textcolor{blue1}{with flat surfaces} in both rotational and translational movements.

Slip-aware control significantly enhances the functionality of robotic manipulators by enabling the object-end-effector relative pose to adapt during grasping, thereby extending the operational workspace. This adaptability is particularly valuable in constrained environments, where the manipulator's movement is limited, or for intelligent human-robot interaction, enabling for instance more intuitive and safe handovers. 
Furthermore, in-hand slippage control opens up new opportunities for multi-arm manipulation of single objects, allowing for the repositioning of grasps without releasing the object, thereby enabling more efficient and flexible handling of larger items. Our system has been rigorously tested across a wide range of experiments, demonstrating its effectiveness and versatility. 
\endnote{Hardware build guides and the code developed for this paper is available at: \url{https://github.com/Gabrieleenx/Slip-Aware-Object-Manipulation-with-Parallel-Grippers}.}

\section{Related Work}

\subsection{Tactile Sensors}
In-hand perception and manipulation are intrinsically tied to hardware. The gripper's structure and the number of DoF dictate its dexterity. 
The design of tactile sensors is closely linked to the contact properties, which play a crucial role in influencing both sliding and grasping behavior.
Numerous tactile sensors have been developed, and comprehensive reviews can be found in
\cite{reviewTactileSensorsChen2018, reviewTactileSensorsZhanat2015, Dahiya2010tactile, YOUSEF2011171, s18040948, fiber_optics_review_2025}. A notable example is the GelSight sensor \cite{GelSightYuan2017}, which evolved from the GelForce sensor \cite{GelForceKamiyama}. The GelForce sensor utilized two layers of markers within an elastomer to detect a 3D force field over a 2D surface. Building on this, the GelSight sensor \cite{GelSightYuan2017}, an advancement of the predecessor described in \cite{GelsightPredecessorJohnson2009}, features a clear elastic polymer with a reflective coating. A camera and multi-angled LEDs behind a rigid transparent plate capture deformations, using colored shading to reconstruct the 3D imprint. Surface markers enable estimation of shear forces and the reconstruction of a 6-axis force vector. The GelSight sensor has been commercialized and widely adopted in research. Another commercially available sensor is the BioTac sensor \cite{Wettels2014}, which has a finger-like appearance and is capable of measuring 3-axis force, vibrations, and temperature. Another approach is detailed in \cite{SUNTouchCostanzo2019, DEMARIA201260}, introducing a tactile sensor later named SUNTouch \cite{controlledSlipCostanzo2023}. This sensor is based on a deformable layer with cavities containing reflectors, where LED-phototransistors estimate deformation, allowing estimation of both the 6-axis F/T and surface deformation.

Several array-like tactile sensors have been proposed to measure the deflection of nibs, as demonstrated in \cite{PapillArray, nibArrayYao2022, NibArrayWang2019, HuhSlipTack2020}. \cite{HuhSlipTack2020} introduced a tactile sensor specifically designed for simple in-hand manipulations, which dynamically clusters nib measurements, allowing for either faster data acquisition or higher spatial resolution. This concept of faster measurements was further expanded in \cite{GloumakovFastInHandSlip2024}, where nib vibrations were captured to estimate sliding velocity. Alternatively, \cite{ridgesDamian2015} explored the use of ridges on a flexible sensing resistor (FSR) sensor to estimate both velocity and position of sliding objects. Earlier approaches include the work of  \cite{SkinAccHowe1989}, who developed a tactile sensor with a rubber skin embedded with an accelerometer to detect vibrations during slippage.
The TacTip sensor, originally developed by \cite{OrignialTacTipChorley2009}, has been adapted into various versions that can be easily manufactured using rapid prototyping techniques, as described in \cite{tacTipFamilyWard2018}. \textcolor{blue1}{Recent studies have explored fiber-optic methods for in-hand motion tracking \cite{TRIPICCHIO2023102990, FBG_Qian_18, FBG_s22218390}, showing promise for pressure, position, and force sensing free from magnetic interference. }

Our approach differs from previous work by focusing on measuring in-hand force and \textcolor{blue}{planar} velocity using independent sensing modalities. We achieve this by equipping the gripper fingers with both F/T sensors and custom relative velocity sensors based on optical mouse sensors. Optical mouse sensors have previously been explored for in-hand object manipulation to a limited extent, prior studies include \textcolor{blue}{\cite{maldonado12improving}, which integrated single optical mouse sensors into the fingertips of the DLR/HIT hand (\cite{DLR_hand}) for slip avoidance and object recognition.} \cite{OpticalMouseSensorSani2011} used a single optical mouse sensor for slip detection, and \cite{OpticalMouseSensorHöver2010}, which employed an optical mouse sensor to capture tool displacement.

\subsection{Grippers}
Parallel grippers typically possess only a single DoF of control, which limits their capability for dexterous in-hand manipulation. As a result, these grippers often rely on external factors such as gravity, external contacts, or the inherent dynamics of the object to achieve complex tasks. Given the single DoF control and the rapid onset of slip events, precise and fast control becomes essential for effective manipulation. However, commercially available grippers often lack accessible, fast low-level force control, which is critical for such tasks.

Given access to sufficient bandwidth, velocity or position controlled grippers can be used to indirectly control grasping force with additional sensors \cite{Francisco2016adaptive}, however, few commercial grippers offers this functionality. 
One \textcolor{blue1}{discontinued} commercial option is the WSG50 gripper, as utilized in \cite{controlledSlipCostanzo2023}, which, with custom software modifications, achieves an external control rate of 50 Hz.  In contrast, custom grippers using Dynamixel servo motors have been employed in various studies \cite{bi2021zeroshotsimtorealtransfertactile, GloumakovFastInHandSlip2024, ZhangVisuotactile2022}, with a reported response time of 0.12 seconds \cite{GloumakovFastInHandSlip2024}. A different example is a custom gripper integrated with force and vibration sensors, as described in \cite{Zaki2010_slip_sensor}. Additionally, \cite{Wang2016_fual_motor} introduced a dual-motor parallel gripper, which combines a position controller with a closed-loop force controller. Another approach, described in \cite{Nicola2023_adaptive_force_control}, involves an adaptive force controller for low-resolution, position-controlled parallel grippers, demonstrating sliding manipulation with external object tracking.

A two-phase finger for a parallel gripper was proposed by \cite{TwophaseGripperChavan2015}, allowing easier reorientation by switching between free-spinning contact and firm grasp. At lower grasp forces, the finger acts as a point contact, while at higher forces, it provides a stable contact surface.  \cite{threeFingerDafle2014} employed a three-finger, single-motor gripper to perform in-hand reorientation of objects using external forces.

\textcolor{blue}{
Anthropomorphic hands represent another approach to in-hand manipulation. Vision-based object reorientation has been demonstrated using the Shadow Dexterous Hand with reinforcement learning \cite{openai2019learningdexterousinhandmanipulation}, and an overview of related learning-based methods is provided in \cite{weinberg2024surveylearningbasedapproachesrobotic}. The RBO Hand 3, a soft hand with 16 degrees of actuation, achieves in-hand manipulation using simple controllers \cite{Puhlmann_2022, oliver_brock_finger_gate}. However, the mechanical complexity of anthropomorphic hands presents challenges for reliability, durability, and cost-effectiveness \cite{HUANG2025100212}. Recent efforts to address these limitations include low-cost designs enabled by rapid prototyping \cite{YANG2021104210} and actuator-reduction strategies to simplify control and reduce cost \cite{KontoudisHumanHand2019, Odhner2014Hand, Santina2018Hand}.
}

\subsection{Slip Control and Estimation }
Estimation of friction and contact properties plays a crucial role in enabling effective in-hand slip-aware control. For instance, in \cite{controlledSlipCostanzo2023}, the authors advanced their research on planar in-hand sliding control using the SUNTouch sensor. Their approach involves estimating angular velocity based on the limit surface concept and the LuGre friction model \cite{LuGre2008}, utilizing estimated friction parameters in conjunction with F/T measurements. By estimating the center of rotation (CoR), they infer the linear velocity from angular velocity estimates. The estimated CoR is constrained to be within 1.5 times the contact radius, which limits the planar velocity estimation to be primarily rotational. The friction estimation method employed in \cite{controlledSlipCostanzo2023} is detailed in \cite{tactileSensingMaria2015}. Our approach does not impose such constraints on planar velocity estimation, and allows for pure linear slippage as well as rotational.

Understanding the contact properties between the gripper and the object is essential for achieving controlled slippage. In previous work, we proposed a LuGre-based planar friction model \textcolor{blue}{for simulating} in-hand slippage \cite{PlanarFrictionMe2024}.  \cite{xydas1999modeling} modelled the contact mechanics for soft finger contacts based on the limit surface concept \cite{goyal1989limit, goyal1991planar} and linear elastic contact models, originally described by \cite{hertz1882contact}. Additionally, in \cite{FrictionEstimatopmLe2021} combined visual and haptic data to estimate friction and assign friction coefficients to material segmentation based on camera input.

In \cite{gravityFrancisco2015} and \cite{Francisco2016adaptive}, a trajectory-following controller was introduced for controlled rotational slippage in a parallel gripper, using an external camera for object pose \textcolor{blue}{tracking}. The optoforce tactile sensor \cite{OptoForce_paper} was employed in \cite{Francisco2016adaptive} alongside an adaptive controller to compensate for errors in the friction coefficient. A framework for picking up and pivoting objects from a flat surface was presented in \cite{FrameworkPivotHolladay2015}.  \cite{HuhSlipTack2020} used a custom sensor to stand objects upright on a flat surface, and the same sensor was employed in \cite{GloumakovFastInHandSlip2024} to estimate slipping velocities and accelerations, enabling fast in-hand linear slippage.  \cite{LinearVelControlChen2021} employed neural networks with the BioTac sensor to estimate linear sliding velocity and implemented a closed-loop controller to maintain a consistent sliding velocity.

In other in-hand slippage research, \cite{bi2021zeroshotsimtorealtransfertactile} utilized custom tactile sensors for swing-up control. The GelSight sensor was employed by \cite{wang2021swingbotlearningphysicalfeatures} to identify physical features through exploratory movements aimed at dynamically swinging up objects.  \cite{MotionConesNikhil2020} used motion cones with a gripper and an external pusher to plan the reorientation and repositioning of objects to achieve a desired final pose.  \cite{3dPoseReorientHou2018} demonstrated the ability to reorient objects using two motion primitives, pivoting and rolling on a table, with a gripper featuring two-phase fingers that enable both a firm grasp and pivoting.

\subsection{Contributions and Overview}
\begin{figure}
    \centering
    \smallskip     \includegraphics[width=1\columnwidth]{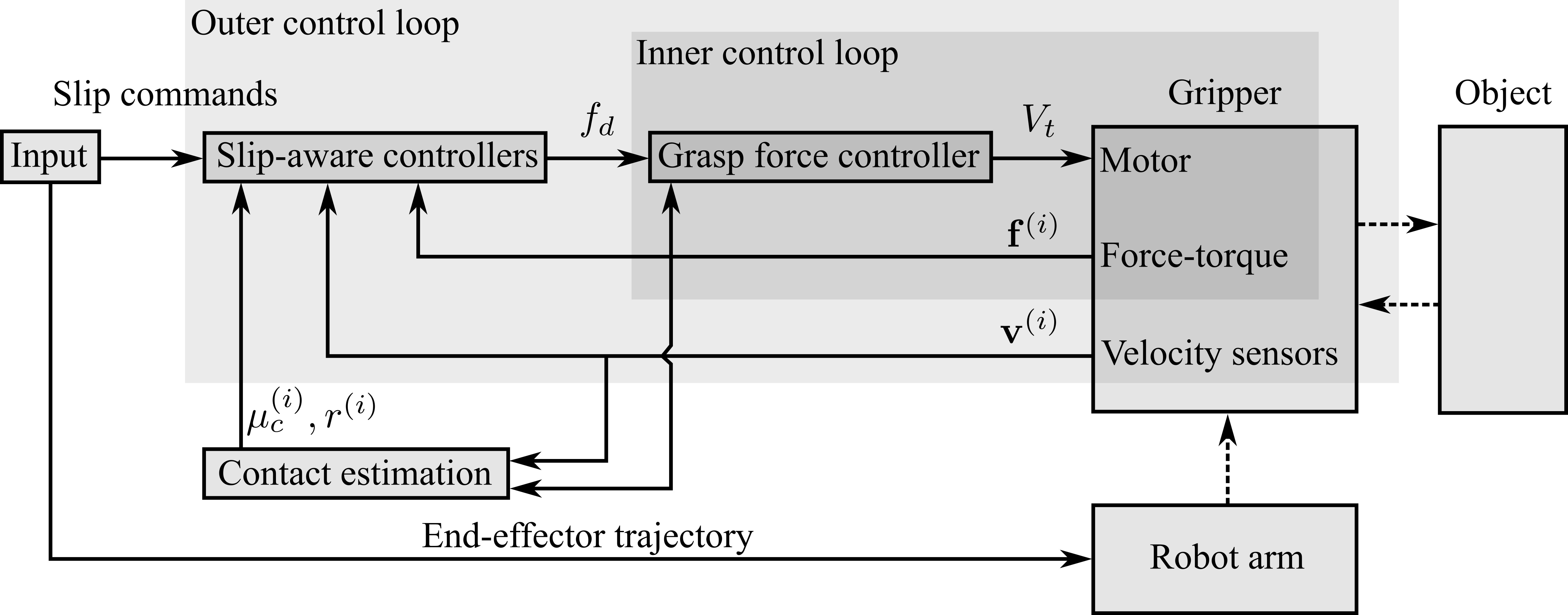}
    \caption{Overview of the \textcolor{blue1}{architecture} with the inner and outer control loops, the mechanical connections are marked with dashed lines and information pathways are marked as solid lines.}
    \label{fig:overview}
    \vspace*{-0.5cm}
\end{figure}

\textcolor{blue}{This paper presents an \textcolor{blue1}{architecture} for in-hand slip-aware control, featuring a high-performance sensorized gripper. The \textcolor{blue1}{architecture} facilitates in-hand slippage in both linear and rotational directions using simple controllers.}
\textcolor{blue}{These slip-aware controllers enables us to extend the capabilities of parallel grippers and relax the assumption of a rigid grasp. Initial estimation of Stribeck friction parameters \cite{Stribeck1902} and the contact radius allows for  manipulation of unfamiliar non-fragile planar objects.}
The key contributions \textcolor{blue}{are:} 
\begin{enumerate}
\vspace{-.5cm}
\item \textbf{High-performance parallel gripper:} \textcolor{blue}{Designed} for in-hand slip-aware control with rapid, precise force \textcolor{blue}{regulation. Unlike \cite{Francisco2016adaptive} and \cite{controlledSlipCostanzo2023}, which use velocity-controlled grippers for force control, we develop a direct force actuation-based gripper. In contrast to \cite{Arimotoetal1999}, which uses deformation derivatives and adaptive PI control, we employ a simple PI controller without deformation derivatives, relying on standard techniques like saturation and gain scheduling, and validate its performance for time-varying references.}

\item \textbf{Planar velocity sensors:} \textcolor{blue}{In-hand sensors that accurately measure planar sliding velocity at the contact -- both linear and rotational -- and can be mounted atop of F/T sensors. In contrast to prior approaches to in-hand slip-aware control \cite{controlledSlipCostanzo2023, Francisco2016adaptive} and tactile optical tracking \cite{maldonado12improving, OpticalMouseSensorSani2011, OpticalMouseSensorHöver2010}, our novel sensor set-up enables planar slip-aware control for parallel grippers through independent in-hand measurements of contact forces and sliding velocities.}

\item \textbf{In-hand estimation of contact properties:} \textcolor{blue}{The proposed \textcolor{blue1}{architecture}, based on independent measurements of contact forces and velocities, enable rapid estimation of contact properties at the moment of object pickup—relying solely on in-hand sensing, in contrast to other approaches for in-hand sliding manipulation \cite{tactileSensingMaria2015, Francisco2016adaptive}.}
\item \textbf{Slip-aware controllers:}  \textcolor{blue}{Leveraging the custom gripper, sensors, and contact estimation, this work introduces—unlike prior approaches \cite{controlledSlipCostanzo2023, Francisco2016adaptive, GloumakovFastInHandSlip2024, LinearVelControlChen2021}—the first \textcolor{blue1}{architecture} demonstrating  both pure linear and rotational gravity-assisted in-hand slippage. We present four simple slip-aware controllers: trajectory-following for gravity-assisted linear and rotational slip, hinge control with linear slip-avoidance, and slip-avoidance.}
\end{enumerate}

An overview of the system is shown in Fig. \ref{fig:overview}. 
The gripper, equipped with F/T and velocity sensors, is mounted on a robot arm. \textcolor{blue}{A grasp force controller -- based on the F/T sensors $\mathbf{f}^{(i)}$ -- takes as input the desired grasp force $f_d$ and produce the output voltage $V_t$ that drives the motor controller.} During a brief initial exploration phase at object pick-up, contact properties 
are estimated based on readings from the velocity $\mathbf{v}^{(i)}$ and F/T sensors. The slip-aware controllers then use these contact estimations and sensor data to execute various slip commands. \textcolor{blue1}{Notably, the slip-aware controllers are independent of the specific gripper dynamics, so in principle a different high-performance gripper could be used; however, the authors have not found a suitable commercial alternative.}

The paper is structured as follows: Section \ref{sec:gripper} presents the design of the gripper and the grasp force controller. Section \ref{sec:vel_sensors} details the relative velocity sensors and their calibration. The method for estimating contact properties is described in Section \ref{sec:contact_est}. Section \ref{sec:controlled_slippage} introduces the four slip-aware controllers. The experiments and results are discussed and analyzed in Section \ref{sec:results}. Finally, the conclusions are presented in Section \ref{sec:conclusions}.

\section{Gripper design} \label{sec:gripper}
\begin{figure}
    \centering
    \smallskip 
    \includegraphics[width=0.8\columnwidth]{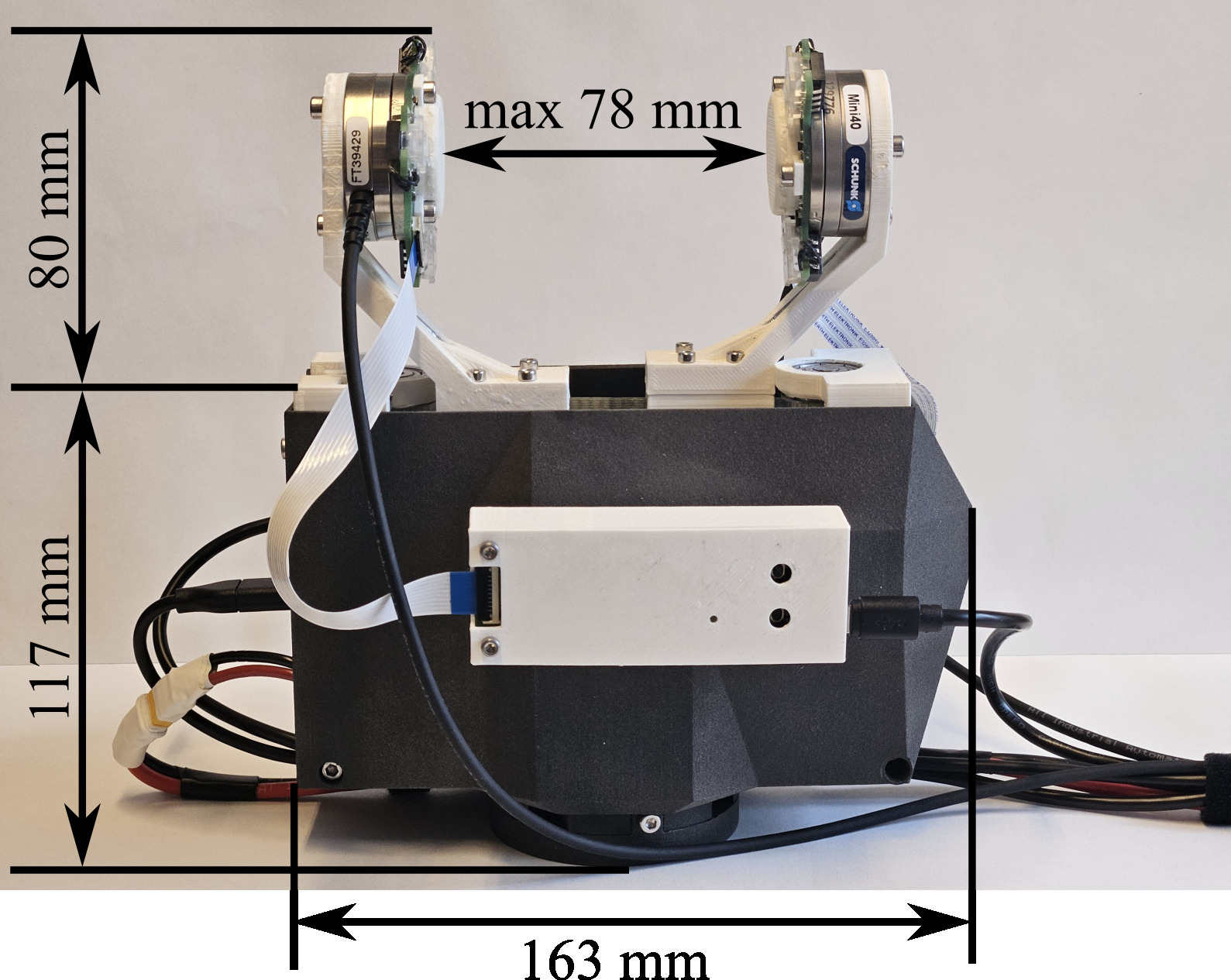}
    \caption{Gripper viewed from the side with the overall dimensions.}
    \label{fig:gripper_side}
    \vspace*{-0.5cm}
\end{figure}

Commercial grippers are generally not designed for in-hand sliding manipulation tasks and often lack accessible low-level force control. To address this limitation, we present a high-performance, force-controlled gripper. \textcolor{blue1}{This permits treating the force response as near-perfect in the design of the slip-aware controllers.} This section details the hardware design of the gripper and the grasp force controller. The gripper features a high-performance motor and efficient field-oriented control (FOC), with the low-level FOC controller running on a ESP32-S3 microcontroller. The design incorporates commercially available components and rapid prototyping techniques, such as 3D printing.

\subsection{Hardware}

\begin{figure}
    \centering
    \smallskip 
    \includegraphics[width=0.99\columnwidth]{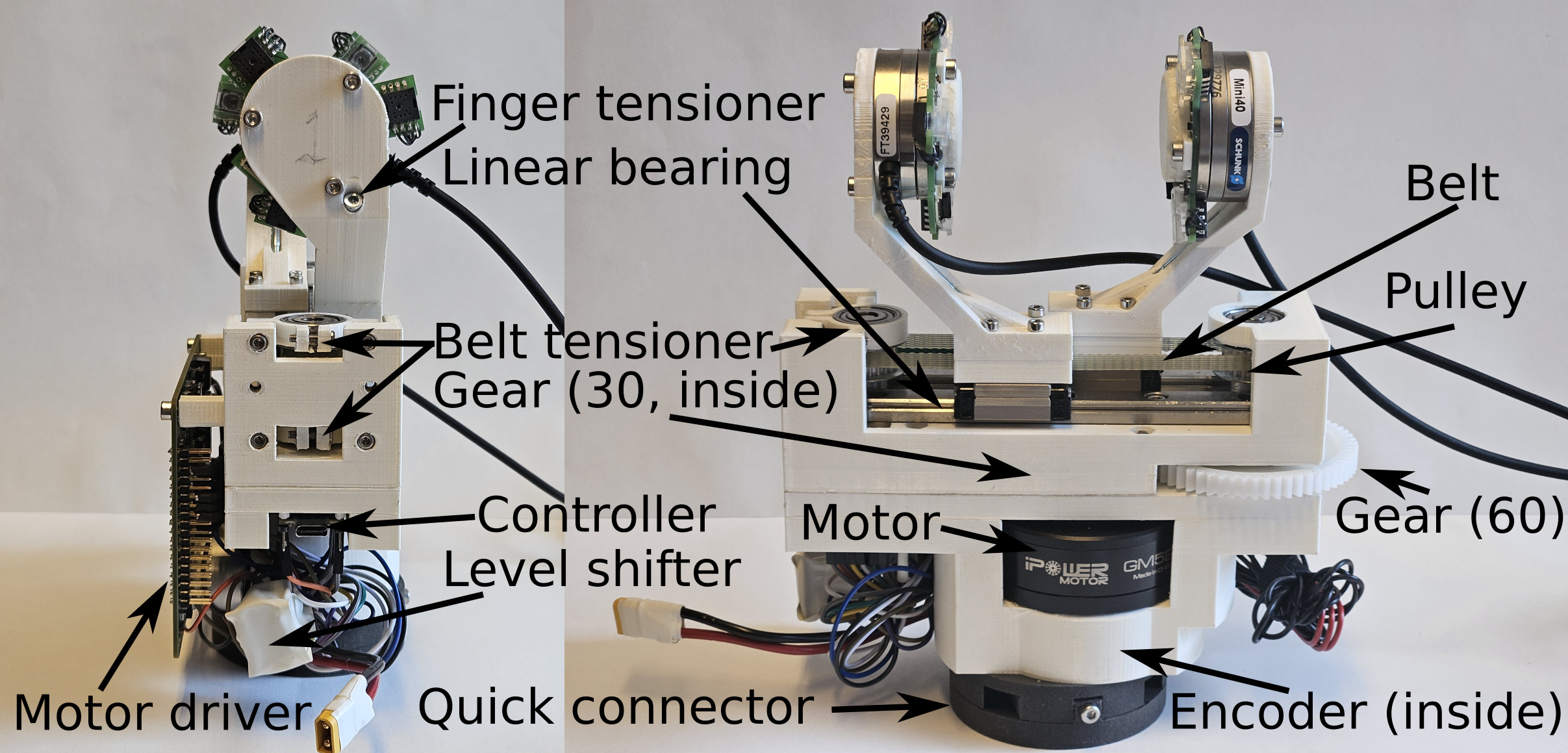}
    \caption{The internal components of the gripper.}
    \label{fig:gripper_inside}
    \vspace*{-0.5cm}
\end{figure}

\begin{figure}
    \centering
    \smallskip 
    \includegraphics[width=0.7\columnwidth]{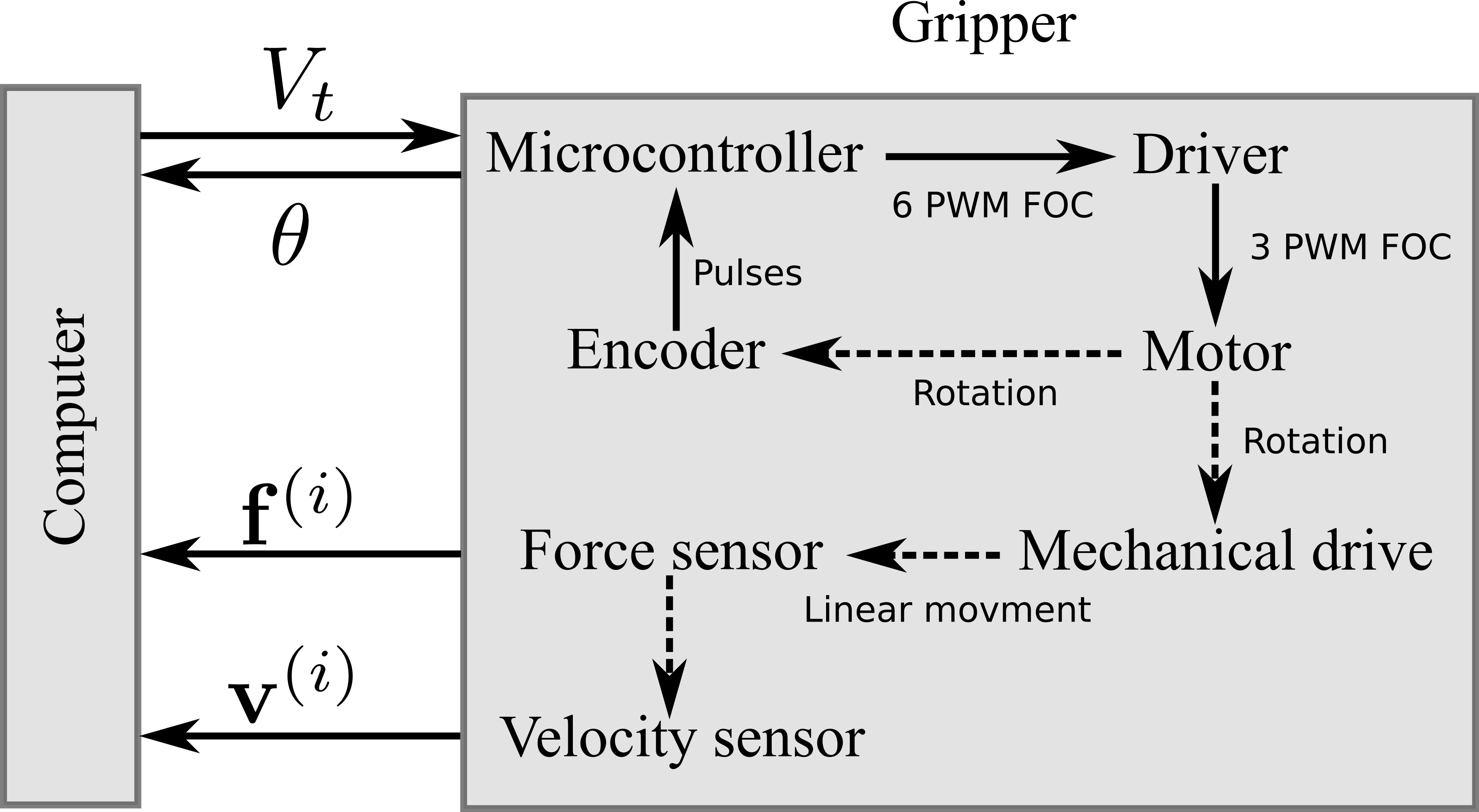}
    \caption{The communication inside the gripper and with the computer.}
    \label{fig:Low_level_communication_overview}
    \vspace*{-0.5cm}
\end{figure}

\textcolor{blue}{The gripper is designed to be quasi-direct drive for rapid force control and its overall dimensions be seen in Fig. \ref{fig:gripper_side},} while the construction and internal components are detailed in Fig. \ref{fig:gripper_inside}. The electrical components are listed in Tab.  \ref{tab:electrical_components}.
The gripper is powered by a high-torque, low Kv BLDC motor, originally designed for camera gimbals and optimized for operation under stalled conditions, \textcolor{blue}{e.g. when grasping objects}. This motor is controlled using FOC based on the SimpleFOC library \cite{simplefoc2022}. An encoder, directly attached to the motor output, provides real-time feedback to the FOC algorithm, ensuring constant torque regardless of the rotor orientation. The encoder also allows for the calculation of finger position and velocity.
The motor is driven by a DRV8302 motor driver, with a software-limited maximum output of 1 A at 20V. \textcolor{blue}{The encoder connected to the} ESP32 microcontroller via a voltage level shifter, offering an effective resolution of 4096 counts per revolution using the ESP32's hardware pulse counters. The communication between the gripper and a computer is handled over USB serial at 500 Hz, with motor angle data $\theta$ transmitted and target voltage commands $V_t$ received. Fig. \ref{fig:Low_level_communication_overview} illustrates the communication pathways, while Tab. \ref{tab:rates} summarizes the system's operational rates.

{
\renewcommand{\arraystretch}{1.2}
\begin{table}[pbb]
\scriptsize
\centering
\caption{Gripper electrical components \label{tab:electrical_components}}
\begin{tabular}{c|c}
\thickhline
 Microcontroller  & ESP32  \\ \hline
 Motor driver & DRV8302\\ \hline
 Encoder & AMT103-V\\ \hline
 Motor & GM5208-24\\ \hline
 Level shifter & BSS138 \\ \hline
\end{tabular}
\end{table}
}

\begin{figure}
    \centering
    \smallskip 
    \includegraphics[width=0.7\columnwidth, angle =0]{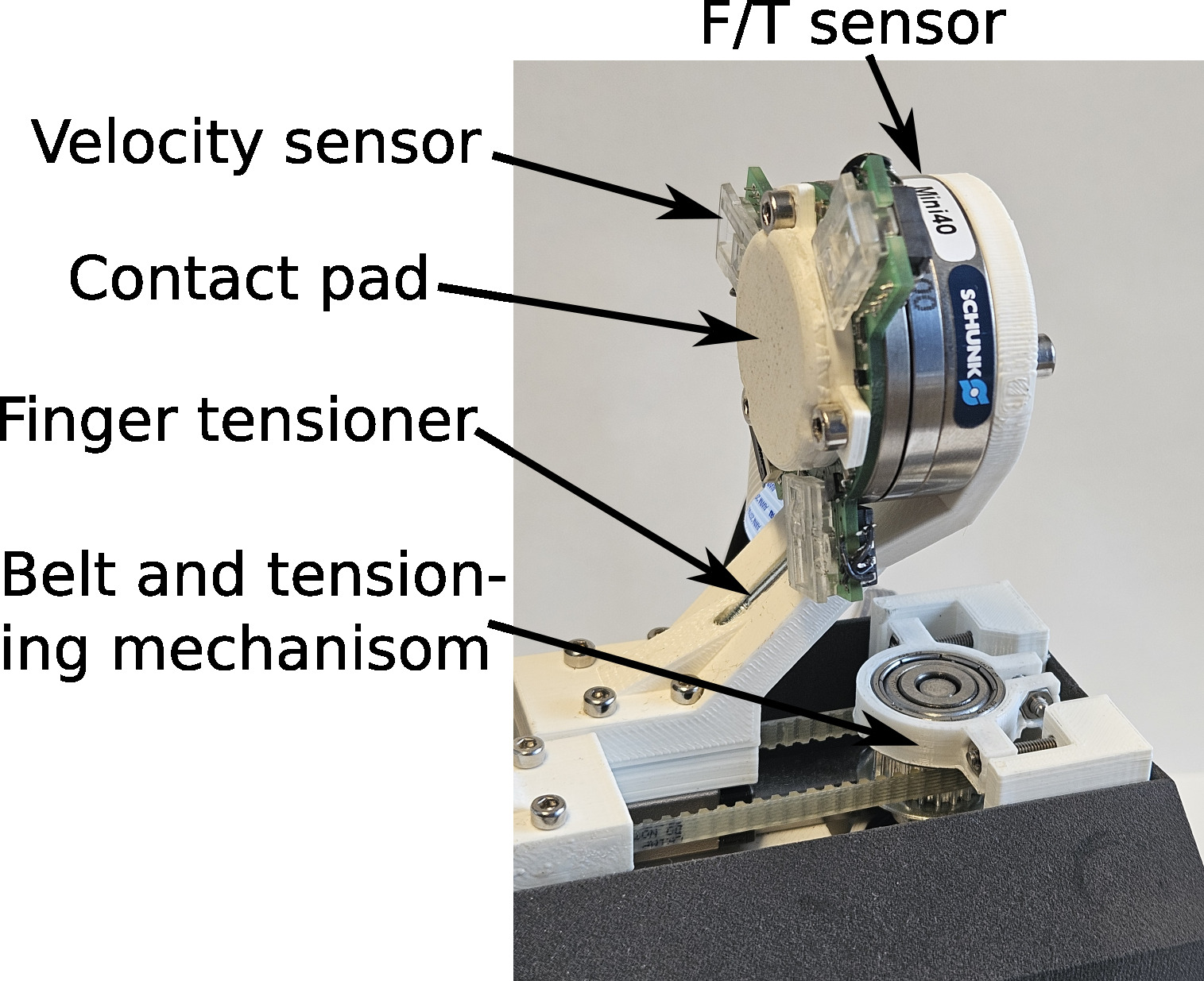}
    \caption{Sensors, finger and belt tensioning mechanism.}
    \label{fig:finger_closeup}
    \vspace*{-0.5cm}
\end{figure}

The gripper's body is primarily constructed using rapid prototyping techniques. As shown in Fig. \ref{fig:gripper_inside}, all white components, except the gears, are 3D-printed in PLA using an FDM printer, while the outer shell and quick connector are produced via SLS printing. The motor is coupled to a 2:1 gear ratio, effectively doubling its torque output \textcolor{blue}{while maintaining a quasi-direct-drive configuration}. This output gear drives a belt pulley system that converts rotational motion into linear motion. Each belt pulley has 22 teeth and a diameter of 17.51 mm, resulting in \textcolor{blue}{27.5} mm of linear travel per motor revolution for each finger. The gripper achieves a measured maximum grasp force of approximately 45 N. The pulley and gear shafts are mounted on rotational bearings, while the fingers move along two linear bearings and are connected to the belt on either side. The finger design allows for a maximum grasp opening of 78 mm. Belt tensioning is achieved through adjustable screws, as depicted in Figs. \ref{fig:gripper_inside} and \ref{fig:finger_closeup}. The total weight of the gripper, including the sensors, is 1.23 kg.

The contact pad (see Fig. \ref{fig:finger_closeup}) is flat and rigid to accommodate the velocity sensors’ sensitivity to surface proximity. Designed for easy interchangeability, the contact pad is circular with a 15 mm radius and is 3D printed from both PLA and TPU95. The TPU95 surface increases friction compared to PLA alone. The fingers, printed in solid PLA, tend to flex under load. To counteract this, a long M3 screw runs along the inner side of each finger, preloading one side to create an adjustable counter-flex, as partially visible in Fig.  \ref{fig:finger_closeup}.

\subsection{Grasp Force Control}\label{sec:closed_loop_control}

{
\renewcommand{\arraystretch}{1.2}
\begin{table}[pbb]
\scriptsize
\centering
\caption{Loop and communication rates  \label{tab:rates}}
\begin{tabular}{c|c}
\thickhline
 FOC  & $> 1e4$ Hz \\ \hline
Force-torque sensor & 1e3 Hz \\ \hline
 Gripper serial communication & 500 Hz \\ \hline
 Grasp force controller & 500 Hz \\ \hline
 Velocity sensors & 120 Hz\\ \hline
 Slip-aware controllers & 120 Hz\\ \hline
\end{tabular}
\end{table}
}

\begin{figure}
    \centering
    \smallskip 
    \includegraphics[width=0.8\columnwidth]{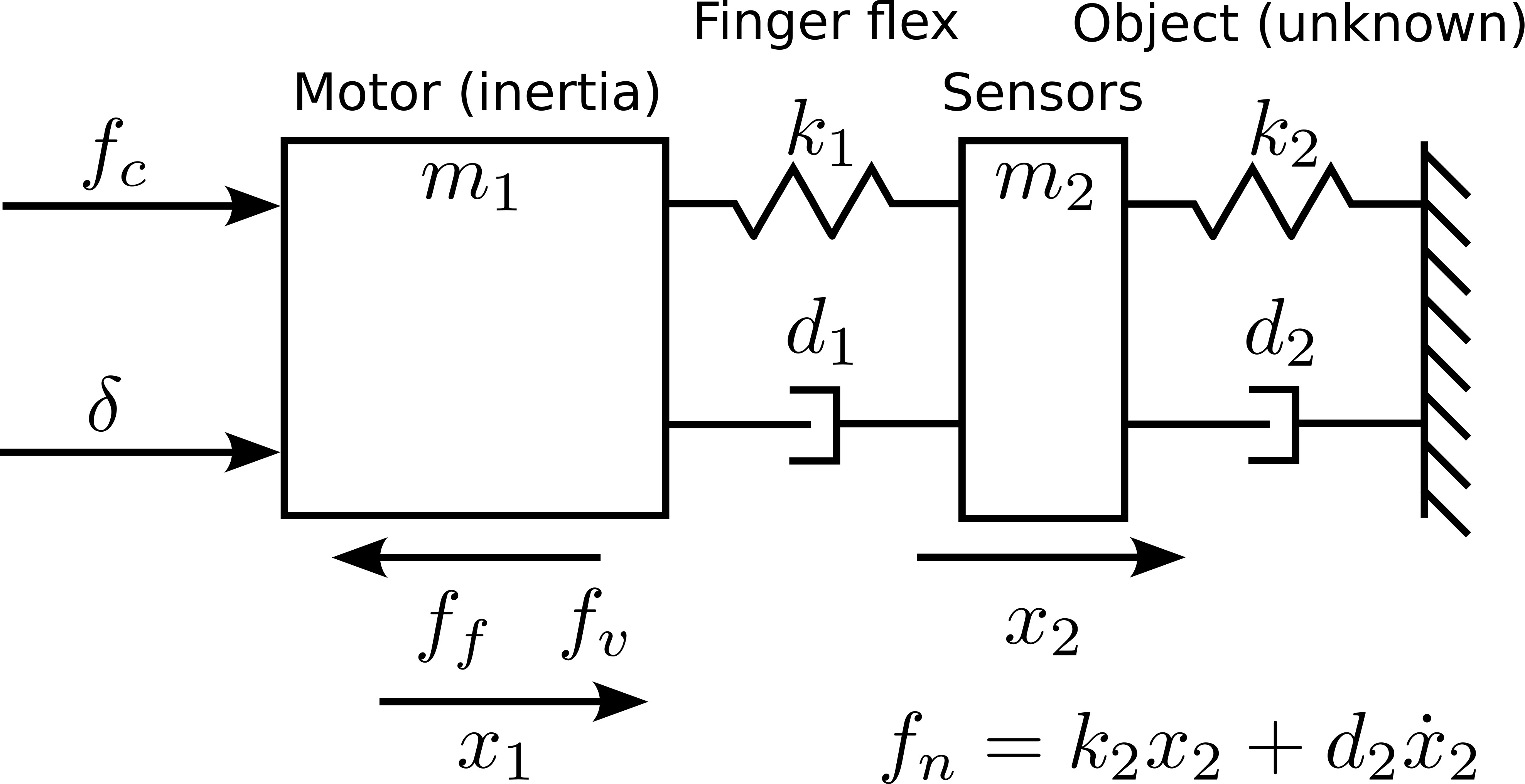}
    \caption{Simplified model of gripper dynamics.}
    \label{fig:Model}
    \vspace*{-0.5cm}
\end{figure}

The grasp controller is responsible for \textcolor{blue}{rapidly} achieving a desired grasp force $f_d$\textcolor{blue}{, with the feedback $f_n$ from the F/T sensors. We model the gripper and analyze it to motivate our grasp force controller.} The controller outputs a commanded force, $f_c$, which is translated into a target voltage $V_t$ for the FOC algorithm as follows:
\begin{equation}
    V_t = -\frac{1}{k_f}f_c
\end{equation}
where the sign of $V_t$ determines the direction of the force: positive $V_t$ opens the gripper, while negative $V_t$ closes it. However, the actual grasp force exerted by the gripper may not exactly match $f_c$ due to factors such as friction $f_f$, damping force from the motor's back EMF $f_v$ and other unmodeled disturbances $\delta$, like motor winding temperature. The gripper's dynamics during object grasping can be simplified and modeled as shown in Fig. \ref{fig:Model}. In this model, the motor's rotational motion is expressed as an equivalent linear motion, leading to the following gripper dynamics:
\begin{equation}\label{eq:SS1}
     m_1 \ddot{x}_1  = f_c - f_f - f_v - k_1(x_1 - x_2) - d_1 (\dot{x}_1 - \dot{x}_2) + \delta
\end{equation}
where $m_1$ is the mass equivalent of the motor inertia, and $x_1$ is the equivalent linear displacement. The forces acting on the sensors can be modeled similarly as:
\begin{equation}\label{eq:SS2}
    m_2 \ddot{x}_2 =  k_1(x_1 - x_2) + d_1 (\dot{x}_1 - \dot{x}_2) - k_2 x_2 - d_2 \dot{x}_2
\end{equation}
where $x_2$ represents the displacement of the finger after it has made contact with the object, with  $x_2 \geq 0$. The measured normal force \textcolor{blue}{by the F/T sensors} can be modeled as:
\begin{equation}\label{eq:Fn}
f_n = k_2 x_2  + d_2 \dot{x}_2.    
\end{equation}

To analyze stability of the system \textcolor{blue}{with states $[x_1, \dot{x}_1, x_2, \dot{x}_2]^T$} consider the Lyapunov function candidate:

\begin{equation}\label{eq:V}
\begin{split}
    V =& \frac{1}{2} m_1 \dot{x}_1^2 + \frac{1}{2} m_2 \dot{x}_2^2 + \frac{1}{2}k_1\left(x_1 - x_2 - \frac{f_c+\delta}{k_1}\right)^2 \\ &+ \frac{1}{2}k_2 \left(x_2 - \frac{f_c+\delta}{k_2}\right)^2
\end{split}
\end{equation}
\textcolor{blue}{where the first two terms represent the kinetic energy, and the last two terms represent the potential energy stored in the springs, with the equilibrium position offset by the disturbances.} The time derivative of $V$ given by (\ref{eq:V}) along the systems trajectories (\ref{eq:SS1}), (\ref{eq:SS2}) is given by:
\begin{equation}
\begin{split}
     \dot{V} =  - d_2 \dot{x}_2^2 - (f_f + f_v )\dot{x}_1 - d_1(\dot{x}_1 - \dot{x}_2)^2.
\end{split}
\end{equation}

The friction $f_f$ and back emf $f_v$ always act in the opposite direction of $\dot{x}_1$, where $f_v$ naturally dampens the system\textcolor{blue}{, i.e. $- (f_f + f_v )\dot{x}_1 \leq 0, \, \forall \dot{x}_1$}, and thus: 
\begin{equation}
\begin{split}
     \dot{V} \leq - d_2 \dot{x}_2^2   - d_1(\dot{x}_1 - \dot{x}_2)^2. 
\end{split}
\end{equation}
As $V$ is positive definite and $\dot{V}\leq 0$, we can readily prove, relying on LaSalle's theorem for asymptotic stability,  that $\dot{x}_1=0$, $\dot{x}_2=0$ is invariant and subsequently that the equilibrium point $\left(\frac{k_1+k_2}{k_1k_2}(f_c+\delta),0,\frac{f_c+\delta}{k_2},0\right)$ is asymptotically stable, thus implying  $f_n \rightarrow f_c+\delta$. Indeed extensive experiments have shown that if we apply a feedforward force control $f_c=f_d$ system is asymptotically stable but with a bias affected by disturbances. To motivate the choice of the controller we will consider a reduced state system by assuming that the dynamics of convergence to  the invariant set $\dot{x}_1-\dot{x}_2\equiv0$ is faster as compared to the dynamics of $\dot{x}_2$. In fact, substituing (\ref{eq:SS2}) into (\ref{eq:SS1}) and setting $\ddot{x}_1=\ddot{x}_2$ we get:
\begin{equation}\label{eq:SS_reduced}
     (m_1+m_2) \ddot{x}_2  = f_c - f_f - f_v - f_n + \delta
\end{equation}
Adding  $\textcolor{blue}{\frac{k_2}{d_2}}(m_1+m_2)\dot{x}_2$ in both sides of (\ref{eq:SS_reduced}) we get the first-order model with state $f_n$:
\begin{equation}\label{eq:CLSF}
     T_f\dot{f}_n = f_c  - f_n +\Delta(t),
\end{equation}
 time constant $T_f=\frac{m_1+m_2}{d_2}$ and disturbance input  $\Delta(t)=- f_f - f_v+ \frac{k_2}{d_2}(m_1+m_2)\dot{x}_2+\delta$.
System (\ref{eq:CLSF}) calls for a PI controller with a feedforward term for the desired force $f_d$. In particular, the proposed controller is given by:
\begin{equation}
    f_c = f_d + \frac{k_P}{\eta} e + k_I I_c(e)
\end{equation}
where $k_P$ and $k_I$ are positive control gains for the proportional and integral gain respectively, $e = f_d - f_n$ the force error, and $\eta = \gamma  {\Delta e}^2 + 1$ an error dependent scaling factor, where $\Delta e = e(t) - e(t-t_s)$ for a discrete time step size $t_s$. The integral term $I_c$ is given by:
\begin{equation}
    I_c = \min\left(\max\left(\int^t_0 e(\sigma) d\sigma, I_b\right), -I_b\right) 
\end{equation}
bounded by $
    I_b = \frac{I_\textrm{max}}{\eta k_I }$ to alleviate wind-up.
Note that $f_n$ is the force measured by the F/T sensors. 
The scaling factor $\eta$ aims at reducing  the overshoot by  limiting the P and I terms when $\Delta e$ is large. The real-world stability and performance of the controller are tested using step signals and varying sinusoidal profiles, as detailed in \ref{sec:gripper_results}.




\section{Velocity sensor design} \label{sec:vel_sensors}

\begin{figure}
    \centering
    \smallskip 
    \includegraphics[width=0.9\columnwidth]{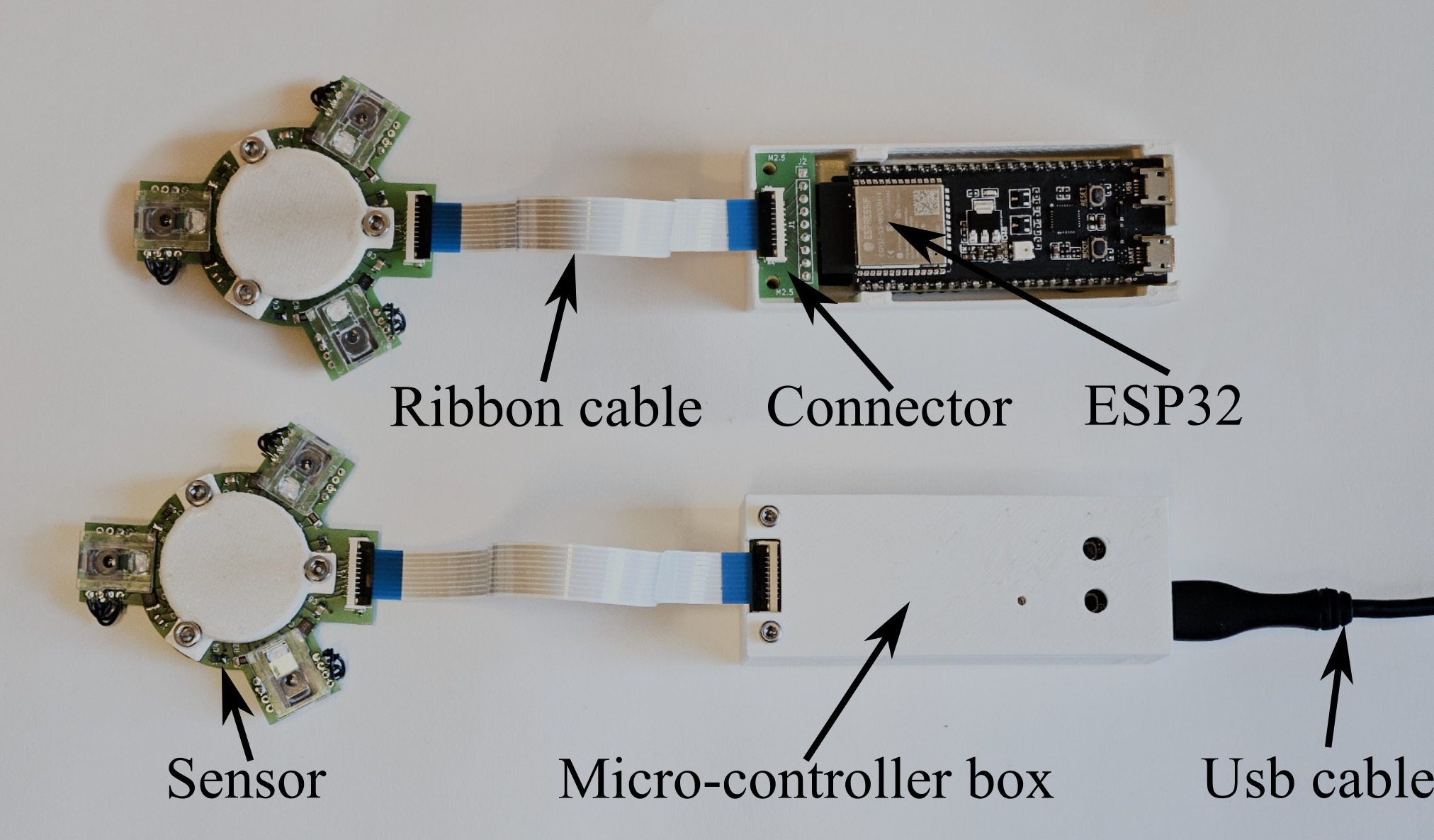}
    \caption{Overview of velocity sensor.}
    \label{fig:vel_sensors_overview}
    \vspace*{-0.5cm}
\end{figure}

In this section, we present the design of a planar velocity sensor, which is mounted between the F/T sensor and the contact pad, see Fig.  \ref{fig:finger_closeup}. \textcolor{blue}{This allows the overall sensor setup to accurately measure both the planar sliding velocity and the forces using in-hand sensing.} An overview of the sensor and its associated microcontroller is provided in Fig. \ref{fig:vel_sensors_overview}.

\subsection{Planar Velocity Sensor}

\begin{figure}
    \centering
    \smallskip 
    \includegraphics[width=0.8\columnwidth]{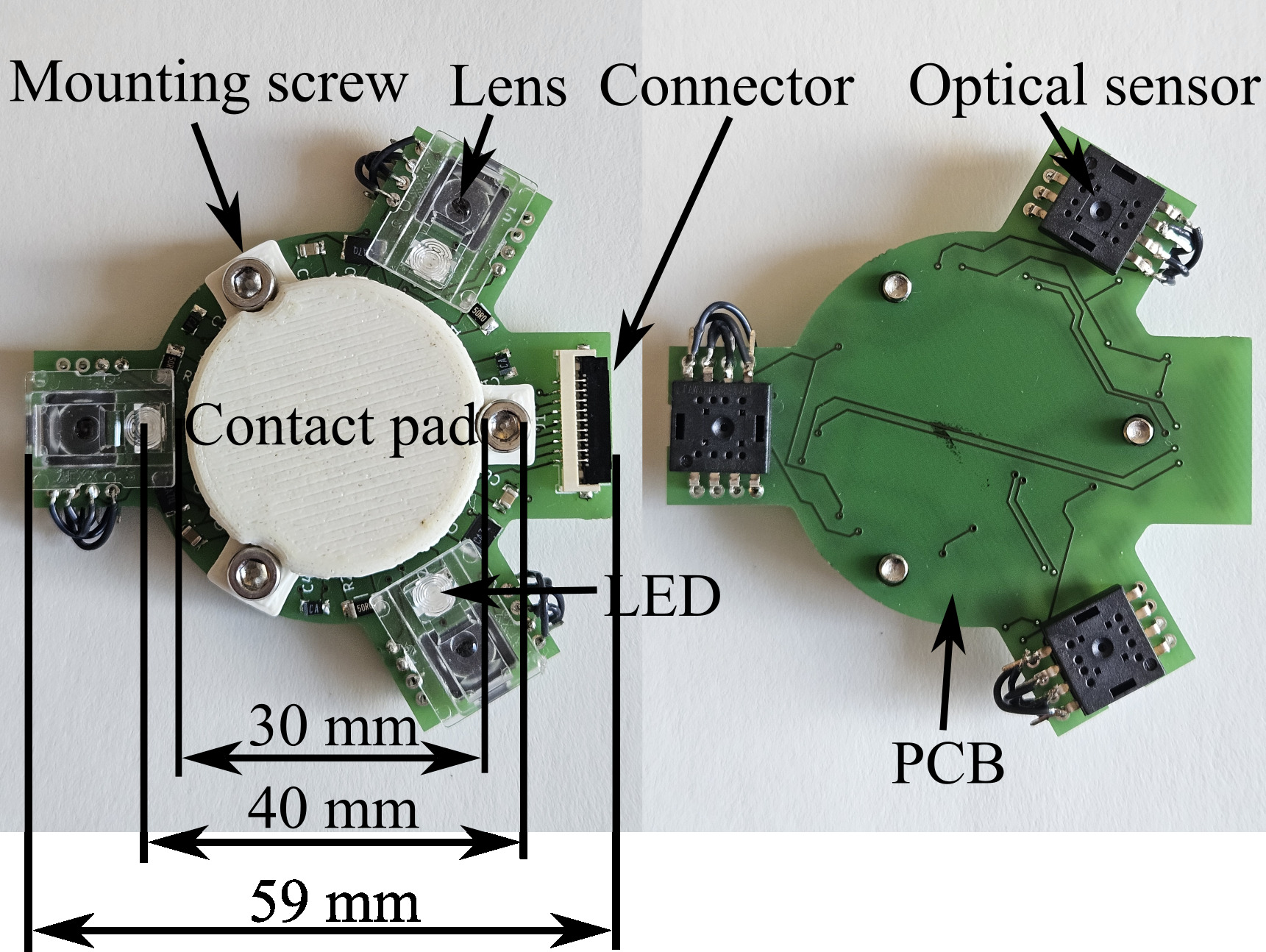}
    \caption{PCB assembly of velocity sensor.}
    \label{fig:vel_sensors}
    \vspace*{-0.3cm}
\end{figure}

 The proposed sensor consists of three optical mouse sensors arranged $120^\circ$ apart in a circular pattern, allowing it to measure linear and rotational relative velocities with redundancy. The sensor is connected via a flexible ribbon cable to an ESP32-S3 microcontroller, as shown in Fig. \ref{fig:vel_sensors_overview}. The microcontroller communicates with a computer over USB serial, see Tab. \ref{tab:rates} communication rates. The sensor assembly is built on a custom PCB, depicted in Fig. \ref{fig:vel_sensors}. The optical mouse sensors (PAW3205DB-TJNT), along with the lenses and LEDs, are repurposed from a Rapoo M300 mouse. The optical sensors are mounted on the back of the PCB, while the LEDs are mounted under the lenses. The lenses are press-fitted into place, and the microcontroller enclosures are mounted on the sides of the gripper, as shown in Fig. \ref{fig:gripper}. 

\begin{figure}
    \centering
    \smallskip 
    \includegraphics[width=0.6\columnwidth]{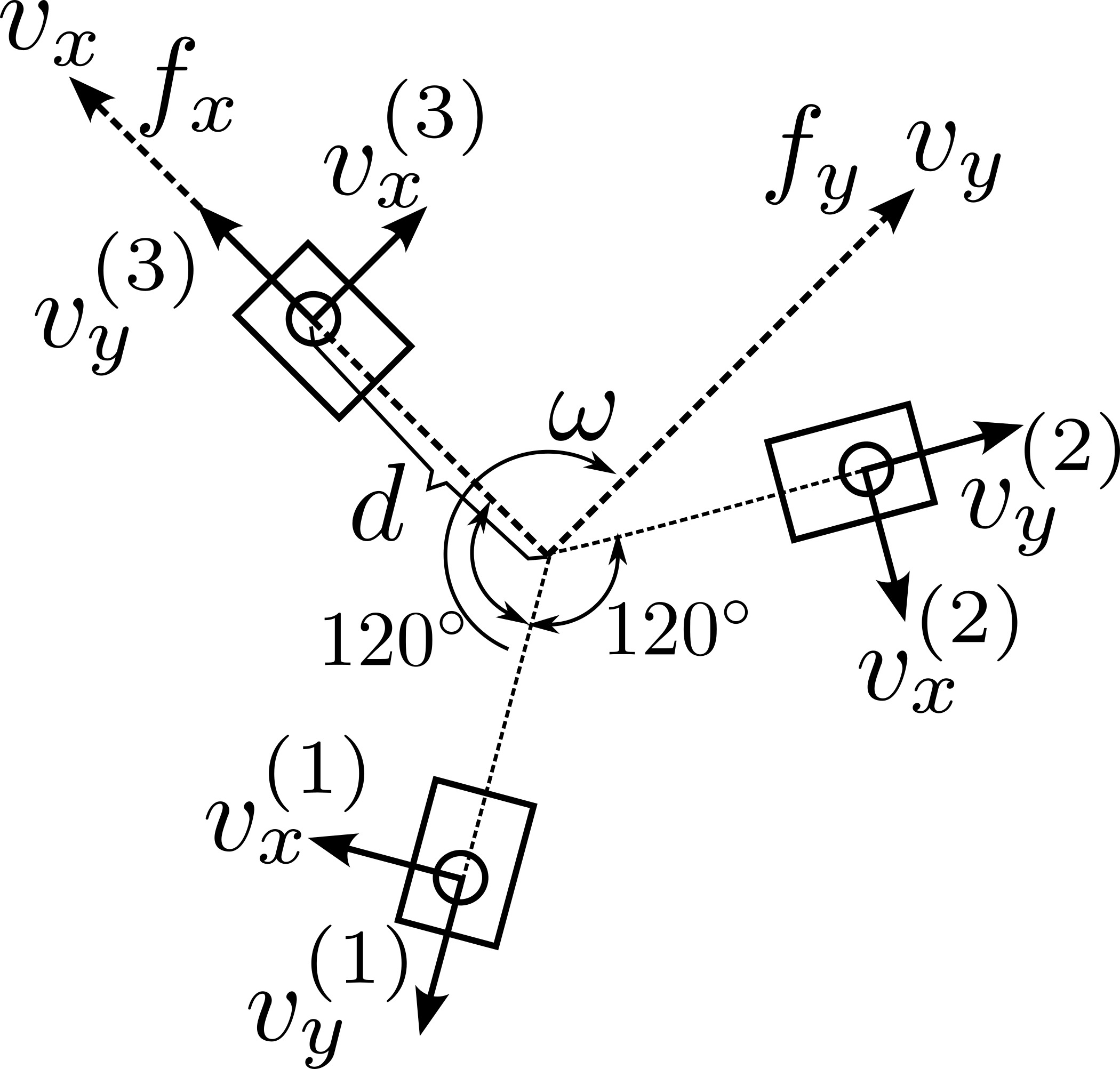}
    \caption{Frames of the optical mouse sensors and the combined velocity sensor in regards to the F/T sensor. The velocities are expressed in $m/s$ or $\textit{rad}/s$.}
    \label{fig:sensor_vel_model}
    \vspace*{-0.5cm}
\end{figure}

Experimental results show that each optical sensor operates at 3200 counts per inch (CPI), providing a spatial resolution of approximately 0.008 mm. Each sensor outputs the measured displacement in the $x$ and $y$ directions, with the sensor frames illustrated in Fig. \ref{fig:sensor_vel_model}. The ESP32 microcontroller is configured with one core dedicated to half-duplex communication with the optical sensors, while the other core handles filtering and USB serial communication. The filtering and communication loop runs at 120 Hz, balancing low latency with accuracy.

The filtering process is carried out in multiple stages: First, the average velocity within a time window is calculated for the $x$ and $y$ directions of each sensor. Next, each velocity is processed through a calibration filter. Finally, the velocity components are fused to output the sensor's planar velocity. The filtering rate is crucial, as a rate that's too high results in noisy velocity measurements, while a rate that's too low introduces unnecessary delays. Minimizing delay is essential for accurately capturing slip and stick events. Common filtering techniques, such as tracking loops, Kalman filters, or other recursive or running average filters, often results in higher delay but lower noise. The velocity $v_{k}^{(j)}$ of the optical mouse sensor $j$ in direction $k$ is processed through a calibration filter:
\begin{equation}\label{eq:calibration}
v_{c,k}^{(j)} = \frac{v_{k}^{(j)}}{b + a v_{k}^{(j)}}    
\end{equation}
where $a$ and $b$ are calibration parameters. The calibration process is further detailed in Section \ref{sec:vel_calibration}.

The velocity output of the planar velocity sensor is aligned with the F/T sensor, as illustrated in Fig. \ref{fig:sensor_vel_model}. The output velocity $\mathbf{v} = [v_x, v_y, \omega]^T$ includes both linear and rotational components. The velocity from each sensor $\mathbf{v}_{c}^{(j)} = [v_{c,x}^{j}, v_{c,y}^{j}]^T$ is incorporated into a measurement model for the planar sensor:
\begin{equation}\label{eq:sensor_model}
    \mathbf{v}_o = \begin{bmatrix}
        \mathbf{v}_{c}^{(1)} \\
        \mathbf{v}_{c}^{(2)} \\
        \mathbf{v}_{c}^{(3)}
    \end{bmatrix} = \begin{bmatrix}
        \mathbf{R}(-30^\circ) & \mathbf{d} \\
        \mathbf{R}(-150^\circ) & \mathbf{d} \\
        \mathbf{R}(90^\circ) & \mathbf{d} 
    \end{bmatrix} \begin{bmatrix}
        v_x\\
        v_y \\
        \omega 
    \end{bmatrix} = \mathbf{A}\mathbf{v}
\end{equation}
where $\mathbf{R}(\cdot) \in SO(2)$ is a rotation matrix, and $\mathbf{d} = [d, 0]^T$, with $d$ representing the distance from the center to the point where the optical sensor measures velocity, see Fig. \ref{fig:sensor_vel_model}. The planar velocity  $\mathbf{v}$ is then estimated using the least squares solution.
\begin{equation}\label{eq:v_hat}
     \hat{\mathbf{v}} = \mathbf{A}^+ \mathbf{v}_o
\end{equation}
where $\cdot^+$ is the Moore-Penrose pseudo inverse. 

There is redundancy in the calculation of $\mathbf{v}$, as only two of the three optical sensors are strictly necessary. This redundancy is exploited by an outlier rejection system, which addresses the risk of one sensor being outside the object or failing to track the surface correctly. When tracking fails, optical mouse sensors typically register zero or significantly reduced velocity. The outlier rejection system operates as follows: First, the estimated velocity $\hat{\mathbf{v}}$ is substituted into \eqref{eq:sensor_model} to calculate the least squares error $\varepsilon = ||\mathbf{v}_o - \hat{\mathbf{v}}_o||_2$.
Let $\varepsilon_t = 0.2$ be a threshold. If the condition $\frac{\varepsilon}{||\mathbf{v_o}||_2 + 1\mathrm{e}{-3}} > \varepsilon_{t}$ is met, the sensor with the lowest velocity  $||v_i||_2$  is discarded.  A new matrix $\bar{\mathbf{A}}$ and vector $\bar{\mathbf{v}}_o$ are then constructed from the remaining two sensors. The updated estimated velocity is given by:
\begin{equation}
     \hat{\mathbf{v}} = \bar{\mathbf{A}}^+ \bar{\mathbf{v}}_o.
\end{equation}
This approach allows for one of the three sensors to be outside the object, with minimal impact on tracking accuracy.

\begin{figure} 
    \centering
    \smallskip 
    \includegraphics[width=0.9\columnwidth]{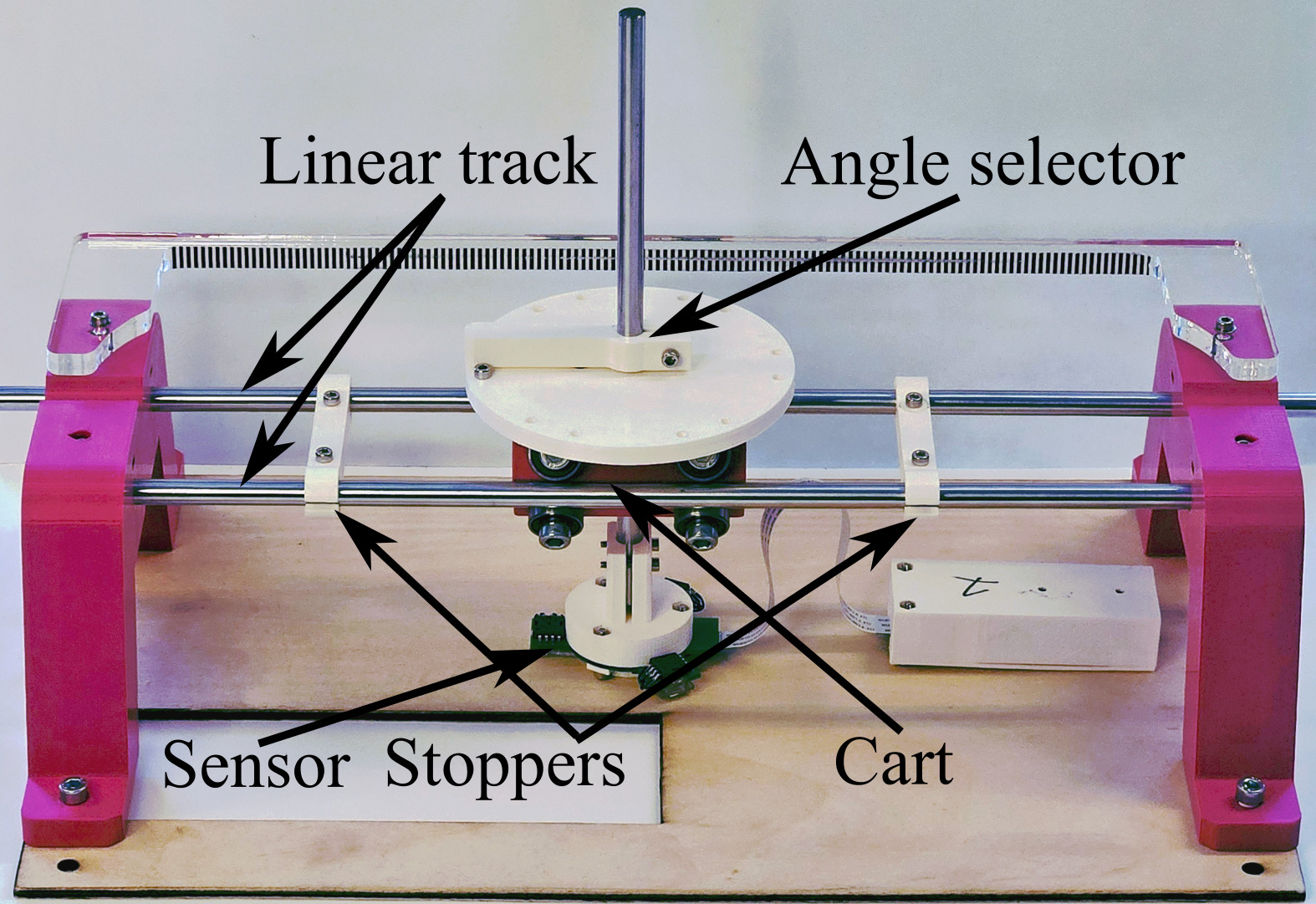}
    \caption{Testing and calibration system for the velocity sensors. The test rig can be set up to test the sensor outlier system.}
    \label{fig:testRig}
    \vspace*{-0.5cm}
\end{figure}

\subsection{Velocity Sensor Calibration}\label{sec:vel_calibration}

The planar velocity sensors are calibrated and tested using the test rig shown in Fig. \ref{fig:testRig}. The rig features a cart that moves along linear tracks, with the sensor mounted on a rotatable shaft. Two adjustable stops are installed on the linear tracks, and the cart includes an angle selector, enabling controlled displacement during testing.

\begin{figure}
    \centering
    \smallskip 
    \includegraphics[width=0.9\columnwidth]{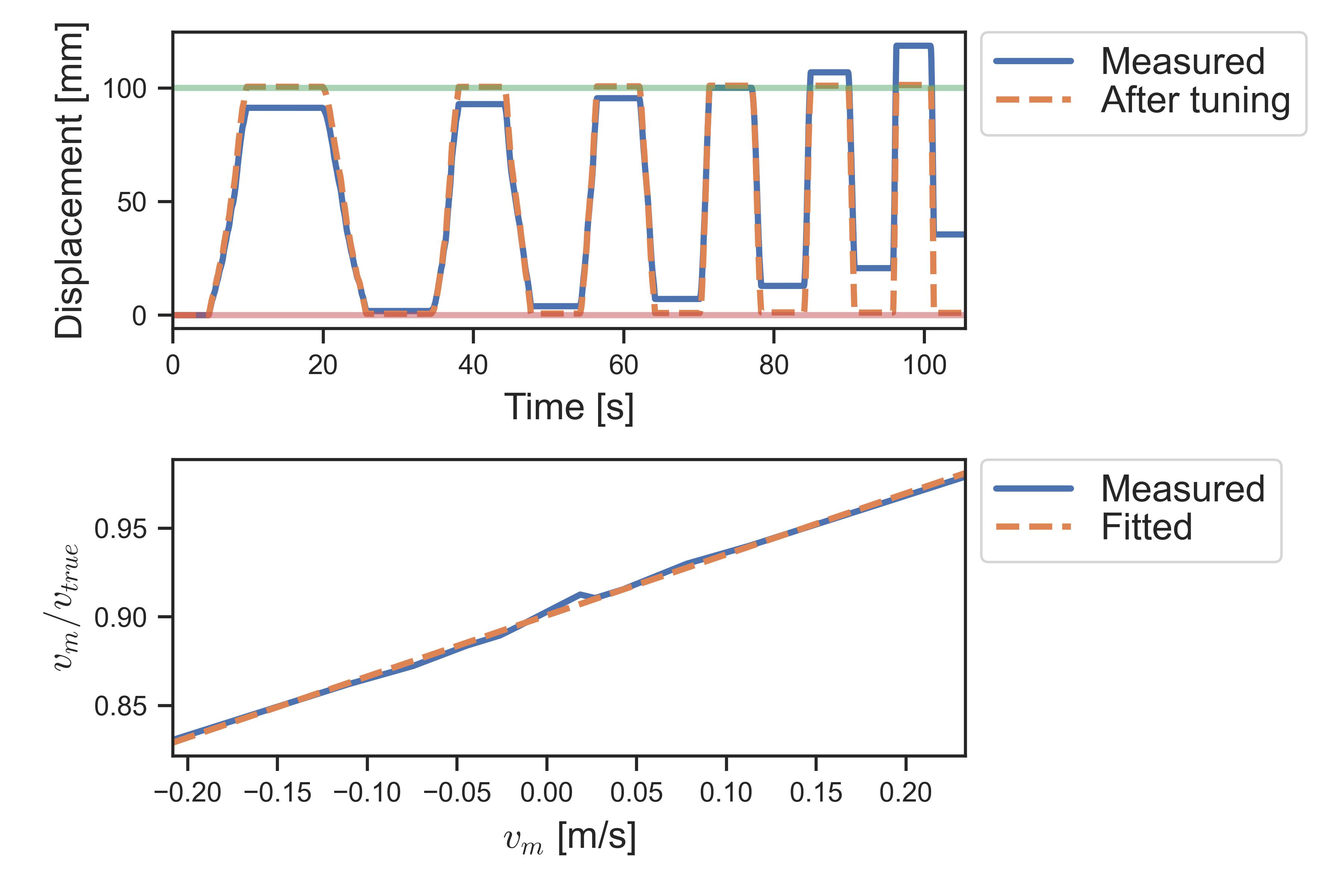}
    \caption{Typical calibration example, we call the average true velocity $v_{\textrm{true}}$ and the raw measured velocity $v_m$}
    \label{fig:vel_calibration}
    \vspace*{-0.5cm}
\end{figure}

To calibrate the optical mouse sensors and determine the calibration parameters  $a$ and $b$ from \eqref{eq:calibration} for both $x$ and $y$ directions, the stoppers are set to allow 100 mm of linear travel for the cart. The angle selector is adjusted according to the specific sensor and direction being calibrated. Data is then collected by manually moving the cart between the stoppers at varying speeds. By knowing the displacement between the stops and calculating the average time taken, the true average velocity is obtained. The ratio between the measured and estimated velocities is then calculated and plotted across different speeds, as shown in Fig. \ref{fig:vel_calibration}. 

Linear regression is used to fit a line to the data, providing an initial estimate for  $a$ and $b$. These coefficients are then manually fine-tuned if necessary, as the automated parameter identification assumes a constant velocity, which is not always the case during manual actuation. The results before and after calibration are displayed in Fig. \ref{fig:vel_calibration}. Additionally, the variable $d$ is manually tuned by rotating the sensor $180^\circ$ and adjusting $d$ until the correct rotation is estimated.

\section{Estimation of contact properties} \label{sec:contact_est}

\begin{figure}
    \centering
    \smallskip 
    \includegraphics[width=0.99\columnwidth]{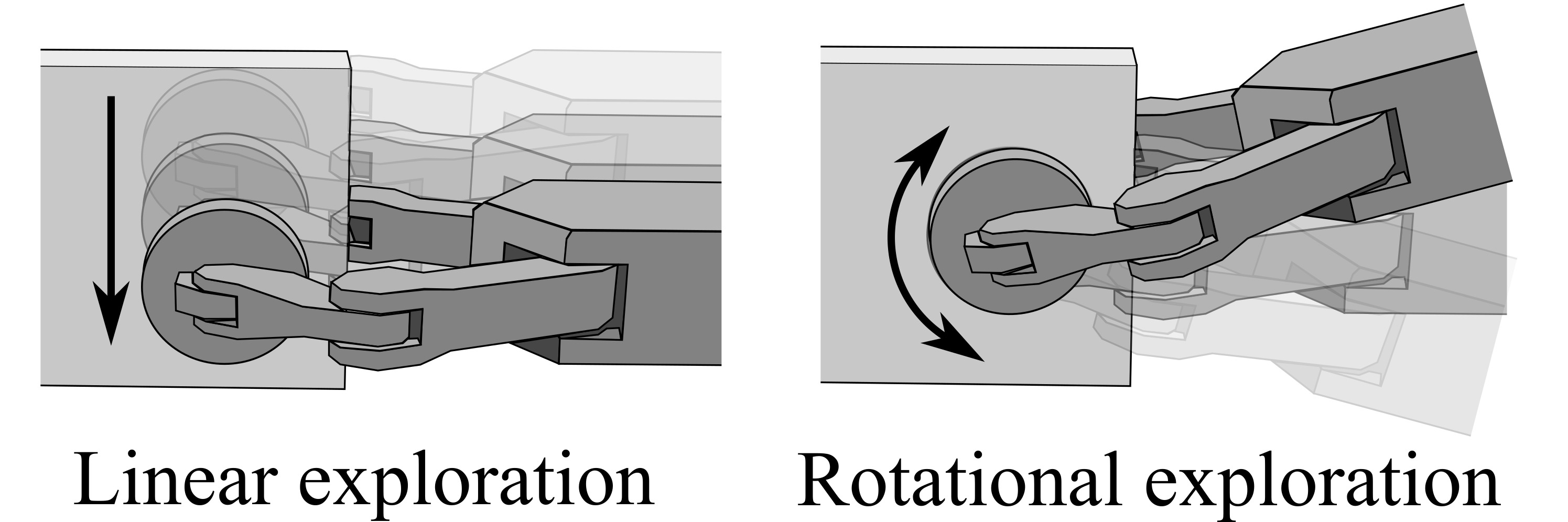}
    \caption{Linear and rotational exploration for estimation of contact properties.}
    \label{fig:friction_explore}
    \vspace*{-0.5cm}
\end{figure}

This section outlines a two-step process for estimating key contact properties: the static, Coulomb, and viscous friction coefficients $\mu_s$, $\mu_c$, and $\mu_v$, as well as the radius $r$ in a rim contact model, which approximates the limit surface as an ellipsoid \cite{PlanarFrictionMe2024}. \textcolor{blue}{The estimations of $\mu_c$ and $r$ are later used in Section \ref{sec:controlled_slippage} for slip-aware control.} First, a brief linear slippage exploration is performed during the pickup phase while the object rests on a surface, as illustrated in Fig. \ref{fig:friction_explore}. The process begins by grasping the object with a normal force 
$f_{g}$. After time $t_{1}$, 
 a linear motion with displacement
$d_e$ downwards towards the surface is initiated, lasting for 
a duration of $t_{2}$. The data collected from this linear exploration is processed using Alg. \ref{alg:frictionEstimation} to estimate the friction coefficients. Each finger is evaluated independently, which allows the method to accommodate objects with different materials on either side. Forces and velocities are measured at the contact point, eliminating the need for precise velocity tracking by the robot manipulator. The algorithm discards data points with excessive angular velocity, as these tend to skew the results toward lower friction coefficient estimates according to limit surface theory \cite{goyal1989limit}.

\begin{algorithm}
\caption{Friction estimation from linear exploration}
\label{alg:frictionEstimation}
\begin{algorithmic}[1]
    \State \textbf{Input:} Force data $f_{x}, f_{y}, f_{n}$ and velocity data $v_{x}, v_{y}, \omega$
    \State \textbf{Output:} The friction coefficients $\mu_c$, $\mu_s$ and $\mu_v$
    \State $\mu_{\textrm{List}} \gets []$, $v_{\textrm{List}} \gets []$,  $\mu_{s, \textrm{List}} \gets []$
    \State $v_{\textrm{min}} \gets 2\mathrm{e}{-3}$, $\omega_\textrm{max} \gets 0.1$,  $f_{n,\textrm{min}} \gets 1$
    \For{$i \gets 1$ to length(data)}
        \If{$|\omega[i]| > \omega_\textrm{max}$ or $f_n[i] < f_{n,\textrm{min}}$}
            \textbf{continue}
        \EndIf
        \State $v \gets ||(v_x[i], v_y[i])||_2$
        \State $\mu \gets ||(f_x[i], f_y[i])||_2/f_n[i]$
        
        \If{$v < v_{\textrm{min}}$}
            \State append $\mu$ to $\mu_{s, \textrm{List}}$
        \Else 
            \State append $\mu$ and $v$ to $\mu_{\textrm{List}}$ and $v_{\textrm{List}}$
        \EndIf
    \EndFor
    \State $\mu_c, \mu_v \gets \textrm{LinearRegression}(v_{\textrm{List}}, \mu_{\textrm{List}})$ \Comment{$\mu \approx \mu_c + \mu_v v$ if $v > v_s$}
    \State $\mu_s \gets \textrm{max}(\textrm{max}(\mu_{s, \textrm{List}}), \mu_c)$
    \State \textbf{return} $\mu_c$, $\mu_s$ and $\mu_v$
\end{algorithmic}
\end{algorithm}

Following the estimation of the friction coefficients, a rotational exploration phase is conducted to estimate the contact radius of an equivalent rim contact. This phase is illustrated in Fig. \ref{fig:friction_explore}. During this phase, force and velocity data are collected while the gripper rotates $\theta_e$ over a period of $t_\theta$ and then returns to its original orientation. The algorithm used to estimate the contact radius $r$ is detailed in Alg. \ref{alg:radiusEstimation}. The estimated radius $r$ will subsequently be used to estimate the limit surface of the contact area. Overall, the friction coefficients and contact radius can be estimated within 2 s of grasping an object.

\begin{algorithm}
\caption{Estimation of contact radius}
\label{alg:radiusEstimation}
\begin{algorithmic}[1]
    \State \textbf{Input:} Torque data $\tau$, friction coefficients $\mu_c$ and velocity data $v_x, v_y, \omega$
    \State \textbf{Output:} Estimated contact rim contact radius $r$
    
    \State $\mu_{\tau,\textrm{List}} \gets []$, $\omega_{\textrm{List}} \gets []$
    
    \State $\omega_{\textrm{min}} \gets 0.1$, $v_\textrm{max} \gets 5\mathrm{e}{-3}$,  $f_{n,\textrm{min}} \gets 1$

    \For{$i \gets 1$ to length(data)}
        \If{$|\omega[i]| < \omega_\textrm{min}$ or $f_n[i] < f_{n,\textrm{min}}$}
            \textbf{continue}
        \EndIf

        \If{$||(v_x[i], v_y[i])||_2 > v_\textrm{max}$}
            \textbf{continue}
        \EndIf
        \State $\mu_\tau \gets |\tau_z[i]|/f_n[i]$
    
        \State append $\mu_\tau$ to $\mu_{\tau,\textrm{List}}$
        \State append $|\omega[i]|$ to $\omega_{\textrm{List}} $
    \EndFor
    \State $a, b \gets \textrm{LinearRegression}(\omega_{\textrm{List}}, \mu_{\tau,\textrm{List}})$ \Comment{$\mu_\tau \approx a + b\omega$ if $ \omega > v_s/r$}
    \State $r \gets a/\mu_c$
    \State \textbf{return} $r$
\end{algorithmic}
\end{algorithm}

\section{Controlled slippage} \label{sec:controlled_slippage}

\begin{figure}
    \centering
    \smallskip 
    \includegraphics[width=0.99\columnwidth]{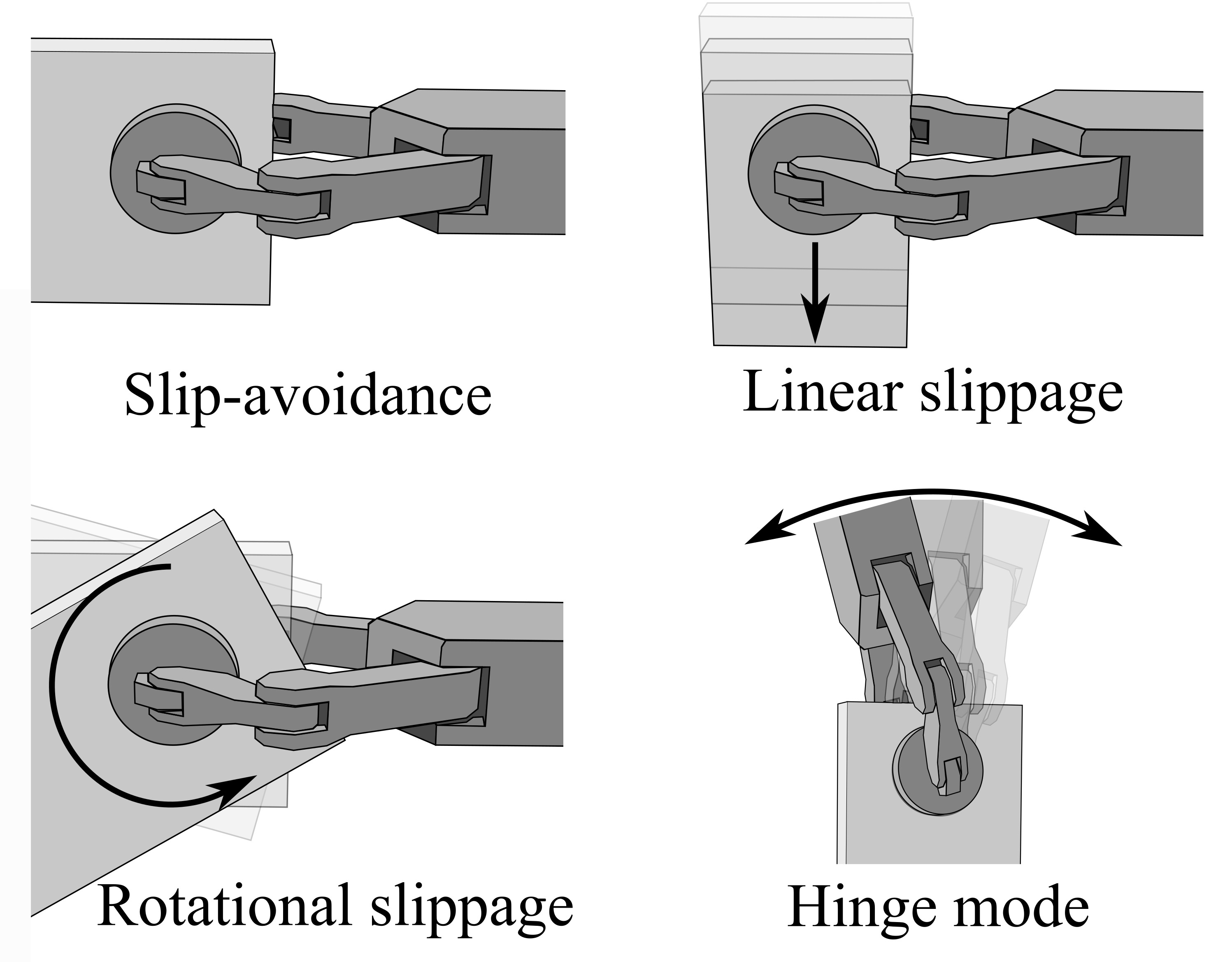}
    \caption{Four different slip control modes}
    \label{fig:slip_modes}
    \vspace*{-0.5cm}
\end{figure}

\begin{figure}
    \centering
    \smallskip 
    \includegraphics[width=0.7\columnwidth]{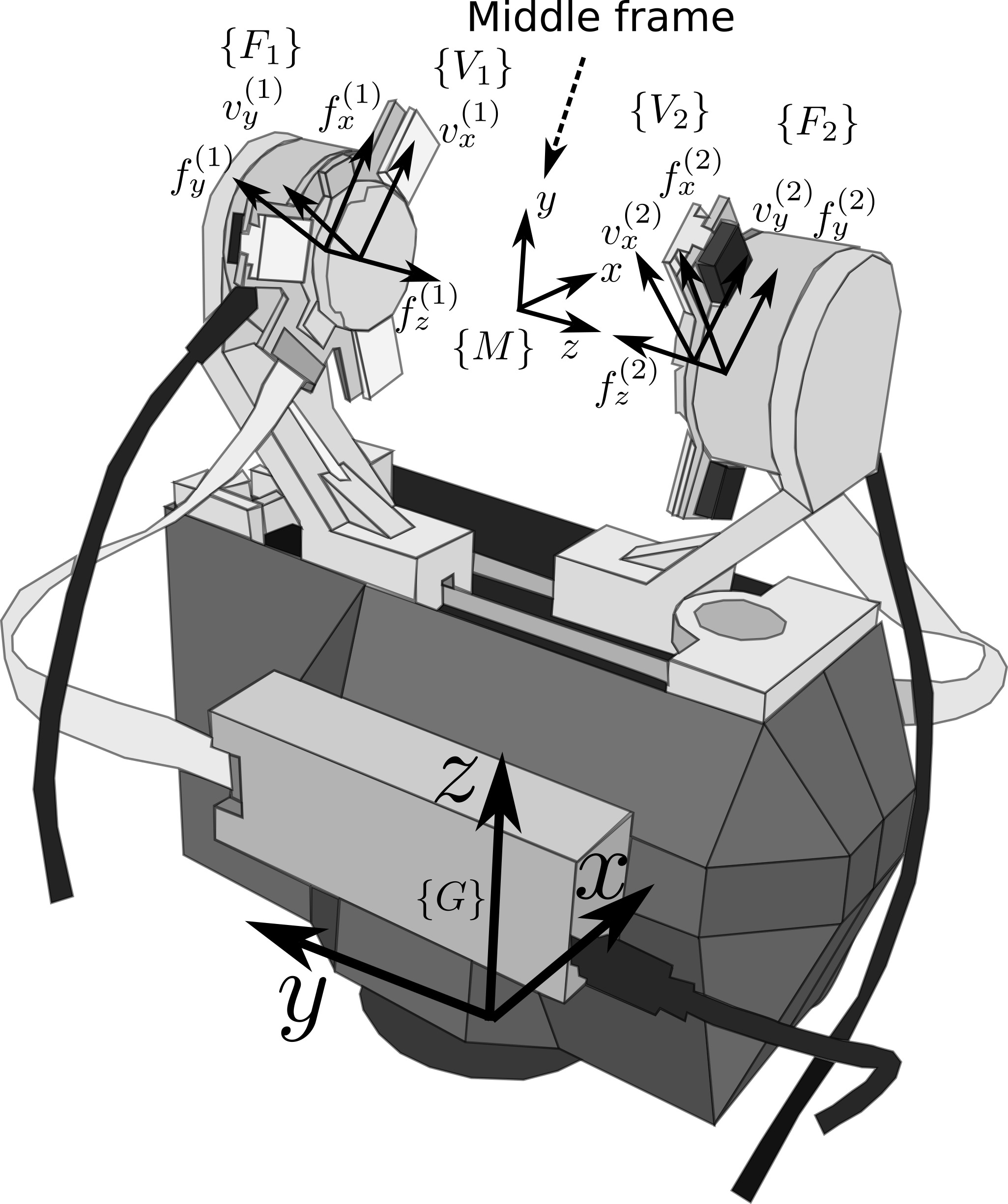}
    \caption{The frames of the gripper.}
    \label{fig:gripper_frames}
    \vspace*{-0.5cm}
\end{figure}

In this section, we present four slip-aware controllers: linear slippage, rotational slippage, hinge control, and slip-avoidance, as illustrated in Fig. \ref{fig:slip_modes}. These controllers are designed to be simple, \textcolor{blue}{and to be used as control primitives for more complex tasks.} \textcolor{blue}{They build on-top of the grasp controller from Section. \ref{sec:closed_loop_control} and are experimentally verified.} The notation $\bar{\cdot}$ denotes the average of both sensors, expressed in the middle frame $\{M\}$ of the gripper, as shown in Fig. \ref{fig:gripper_frames}. For example, $\bar{f}_x$ represents the average force measured by the two F/T sensors in the $x$ direction of the middle frame. All slip-aware controllers described here operate at $\sim 120$ Hz.

\subsection{Slip-Avoidance}

\begin{figure}
    \centering
    \smallskip 
    \includegraphics[width=0.7\columnwidth]{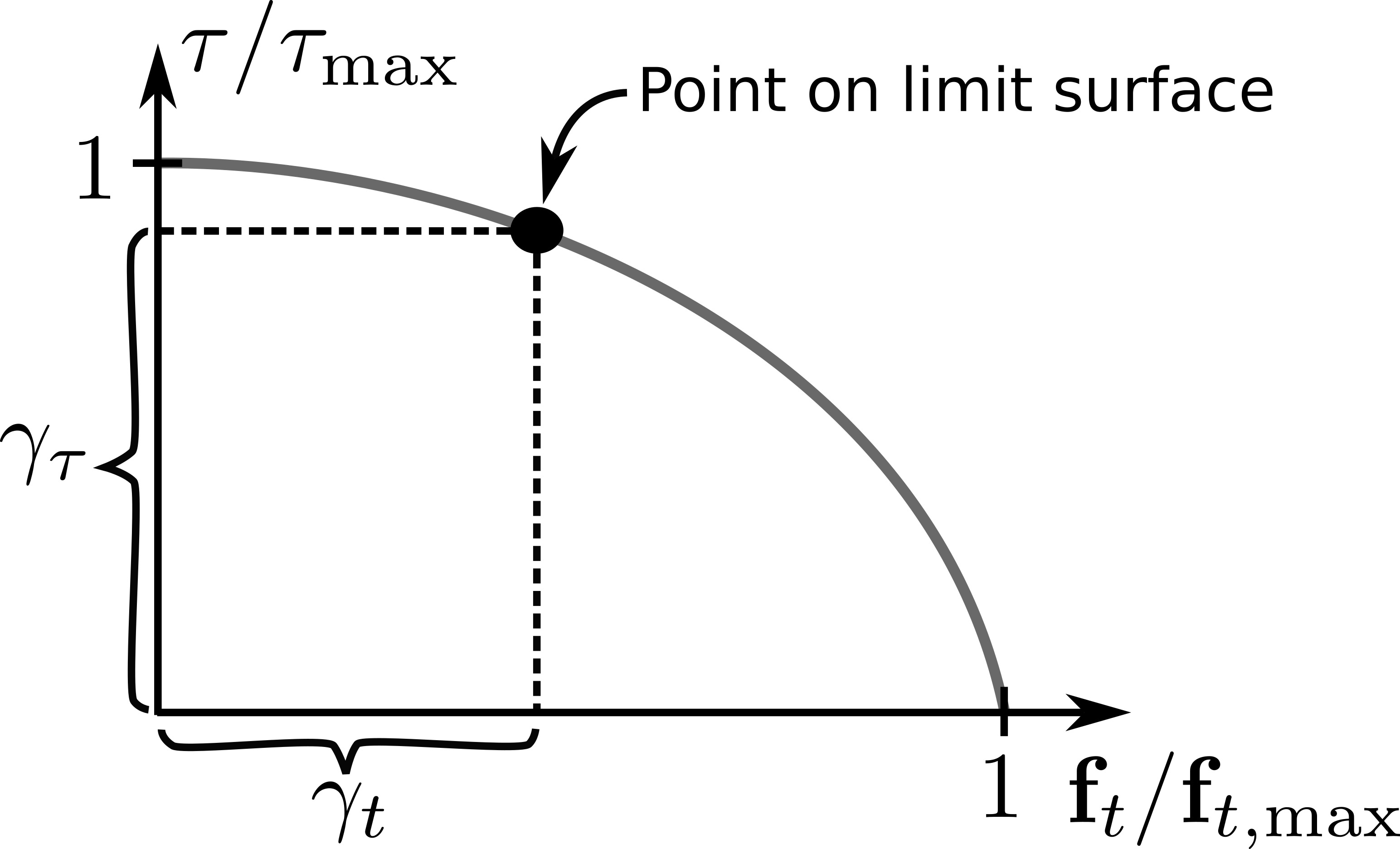}
    \caption{Illustration of the normalized limit surface.}
    \label{fig:limit_surf}
    \vspace*{-0.5cm}
\end{figure}

The primary focus of the slip-avoidance controller is to adjust the grasping force to securely hold an object without excessive grasp force. Let $\bar{f}_t$ be the Euclidean norm of the average tangential forces from the two F/T sensors: 
\begin{equation}
    \bar{f}_t = ||\left[\bar{f}_x, \bar{f}_y\right]^T||_2.
\end{equation}
This is used to calculate the minimum grasp force to counter the measured tangential forces:
\begin{equation}
    f_{n, t, \textrm{min}} = \frac{\bar{f}_t}{\bar{\mu}_c \gamma_t}
\end{equation}
where $\bar{\mu}_c$ is the average estimated Coulomb friction coefficient between the two fingers. The ratio $\gamma_t$ is derived from an ellipsoid approximation of the limit surface (see Fig. \ref{fig:limit_surf}), adjusting the grasp based on the measured torque versus tangential forces. Despite the limitations of the ellipsoid approximation \cite{PlanarFrictionMe2024}, it is chosen for simplicity. The average force wrench between the sensors  $\bar{\mathbf{f}} = \left[\bar{f}_x, \bar{f}_y, \bar{\tau} \right]^T$ and the average rim contact radius $\bar{r}$ are used to calculate $\gamma_t$ as:
\begin{equation}
    \gamma_t = \frac{\bar{f}_t +  \epsilon}{||\mathbf{S}^{-1}\bar{\mathbf{f}}||_2 +\epsilon}
\end{equation}
where $\mathbf{S} = \textrm{diag}([1,1,\bar{r}])$ and $\epsilon 
= 1\mathrm{e}{-5}$ for numerical stability and feasibility. 
The minimum grasp force to counteract the measured torque is given by:
\begin{equation}
    f_{n, \tau, \textrm{min}} = \frac{|\bar{\tau}|}{\bar{\mu}_c \bar{r} \gamma_\tau}
\end{equation}
where:
\begin{equation}
    \gamma_\tau = \frac{|\bar{\tau}|/\bar{r} +  \epsilon}{||\mathbf{S}^{-1}\bar{\mathbf{f}}||_2 +  \epsilon}.
\end{equation}

The overall grasp force, considering both torque and tangential forces, includes a tunable safety margin $\gamma_s$ and a velocity component to account for slippage:
\begin{equation}
    f_{n,c} = \gamma_s \textrm{max}(f_{n, t, \textrm{min}}, f_{n, \tau, \textrm{min}}, f_{n, s, \textrm{min}}) + k_{P, s} ||\mathbf{S}\bar{\mathbf{v}}||_2
\end{equation}
where $\bar{\mathbf{v}} = \left[\bar{v}_x, \bar{v}_y, \bar{\omega} \right]^T$ and $f_{n,s,\textrm{min}}$ is the lower limit on the normal force. To mitigate oscillations, an exponential decay filter is applied at time $t$ with a time step of $t_h$:
\begin{equation}
\scalebox{0.87}{%
    $f_d(t) = \begin{cases}
        f_{n,c}(t) & \textrm{if } f_{n, c}(t) > f_d(t-t_h) \\
        \alpha_s f_d (t-t_h) + (1 - \alpha_s) f_{n, c}(t) & \textrm{otherwise}
    \end{cases}$
}
\end{equation}
where $f_d$ is the control signal to be sent to the grasp force controller described in Section \ref{sec:closed_loop_control}.

\subsection{Linear Slip Control}

\textcolor{blue}{This controller allows the object to be repositioned within the gripper along the gravity vector.}
Let $p_d(t)$ and $v_d(t)=\dot{p}_d(t)$ represent the desired displacement and velocity trajectories, respectively, and define a  reference velocity:
\begin{equation}
    v_c = v_d + K_{p,v} (p_d - p)
\end{equation}
where $p$  is the measured linear distance traveled. The value of $p$ is obtained by integrating the tangential velocity $\bar{v}_t = ||\left[ \bar{v}_x, \bar{v}_y \right]^T ||_2$ \textcolor{blue}{measured by the planar velocity sensors}, assuming the object moves in the direction of the gravitational field.



The grasp force controller designed to follow $v_c$ is a superposition of a PI"D" controller for tracking $v_c$ and a force control term based on $f_s$ \eqref{eq:f_s} representing the normal force required to maintain the slip. The control law is given by:
\begin{equation}\label{eq:linear_grasp_control}
\scalebox{0.95}{%
    $f_{n, c} = f_{s}  - k_{P,l} f_{s} (v_c - \bar{v}_t) - k_{I,l} \int_0^t (v_c - \bar{v}_t) \, dt + k_{D,l}f_{s}\dot{\bar{v}}_t$
}
\end{equation}
where $\dot{\bar{v}}_t$ is the time derivative of the tangential velocity, computed using the backward Euler method and applied to dampen rapid accelerations. To accommodate objects with varying masses and frictional properties, the proportional term depends on $f_s$. The integral term compensates for friction estimation errors and unmodeled object inertia. To manage fluctuations in the measured forces during object acceleration, $f_s$ is updated using the following heuristic:
\begin{equation}\label{eq:f_s}
    f_s(t) = \begin{cases}
        \alpha f_s(t-t_h) + (1-\alpha) \frac{\bar{f}_t}{\bar{\mu}_c} &  \dot{\bar{v}}_t < 0.1 \\
        f_s(t-t_h)  & \dot{\bar{v}}_t \geq 0.1 
    \end{cases}.
\end{equation}
The computed grasp force $f_d$ for the inner loop controller is given by:
\begin{equation}
    f_d = \textrm{max}(f_{n, c}, f_{n, \textrm{min}})
\end{equation}
where $f_{n, \textrm{min}}$ is a lower limit on the normal force.
Finally, the controller switches to the slip-avoidance controller whether the target displacement has been reached or if the trajectory time has run out and the object is not moving towards the target.

\subsection{Rotational Slip Control}

In rotational slip control, the gripper remains stationary while the object rotates within the gripper under the influence of gravity, a similar problem formulation was explored in \cite{Francisco2016adaptive}. Let $\theta_d(t)$ and $\omega_d(t) = \dot{\theta}_d(t)$ represent the desired angle and angular velocity trajectories, respectively, and the reference angular velocity:  
\begin{equation}
    \omega_c = \omega_d + k_{P,\omega} (\theta_d - \theta)
\end{equation}
where \(\theta\) is the measured rotational displacement, obtained by integrating the measured angular velocity $\omega_a = |\bar{\omega}|$. Similarly to \eqref{eq:linear_grasp_control}, the grasp force controller to follow $\omega_c$ is given by:
\begin{equation}
    f_{n, c} = f_{\tau} - k_{P,\tau}  f_{\tau} e_\omega - k_{I,\tau}\int_0^t e_\omega \, dt + k_{D,\tau} f_{\tau}\dot{\omega}_a
\end{equation}
where $e_\omega = \omega_c - \omega_a$, and $\dot{\omega}_a$ is computed using the backward Euler method. Here, $f_{\tau}$ represents the normal force required to maintain slip during rotation and is updated using the following heuristic:
\begin{equation}
    f_{\tau}(t) = \begin{cases}
        \alpha f_{\tau}(t-t_h) + (1-\alpha) \frac{|\bar{\tau}|}{\bar{\mu}_c \bar{r} \gamma_\tau} &  |\dot{\bar{\omega}}| < 0.1 \\
        f_{\tau}(t-t_h)  & |\dot{\bar{\omega}}| \geq 0.1 
    \end{cases}
\end{equation}
The grasp force sent to the inner loop controller is given by:
\begin{equation}
    f_d = \textrm{max}(f_{n, c}, \frac{\bar{f}_t}{\bar{\mu}_c}, f_{n, \textrm{min}})
\end{equation}

As with linear slip control, the controller switches to the slip-avoidance controller if either the target displacement has been reached or if the trajectory time has run out and the object is not moving towards the target.

\subsection{Hinge Mode}

In hinge mode, the object is allowed to rotate freely within the gripper while preventing linear slippage by dynamically adjusting the grasp force. If the object's center of gravity (CoG) is directly beneath the grasp point, the object remains stationary as the gripper rotates around it. The grasp force in hinge mode is calculated as follows:
\begin{equation}
    f_{n, c} = \gamma_{h} \frac{\bar{f}_t}{\bar{\mu}_c} + k_{P,h} \bar{v}_t
\end{equation}
where \(\gamma_{h}\) is a tunable parameter that determines how closely the controller operates to the linear slip point.
The desired grasp force $f_d$ to be sent to the grasp force controller is computed as:
\begin{equation}
    f_d = \textrm{max}(f_{n, c}, f_{n, \textrm{min}}).
\end{equation}

{
\renewcommand{\arraystretch}{1.2}
\begin{table}[pbb]
\scriptsize
\centering
\caption{Object dimensions and weight. \label{tab:object_properites}}
\begin{tabular}{c|c|c}
\thickhline
Object type & Dimensions (h,w,d) & Weight \\ \hline
\thickhline
Sponge & (92, 62,27) mm & 6.8 g\\ \hline
Cardboard & (153, 70, 69) mm & 160.6 g\\ \hline
Case (synthetic leather) & (169, 70, 51) mm & 138.7 g\\ \hline
Plastic & (140, 81, 46) mm & 84.1 g\\ \hline
Wood & (180, 60, 28) mm & 141.3 g\\ \hline
\end{tabular}
\end{table}
}

\section{Results}\label{sec:results}

\begin{figure}
    \centering
    \smallskip 
    \includegraphics[width=0.99\columnwidth]{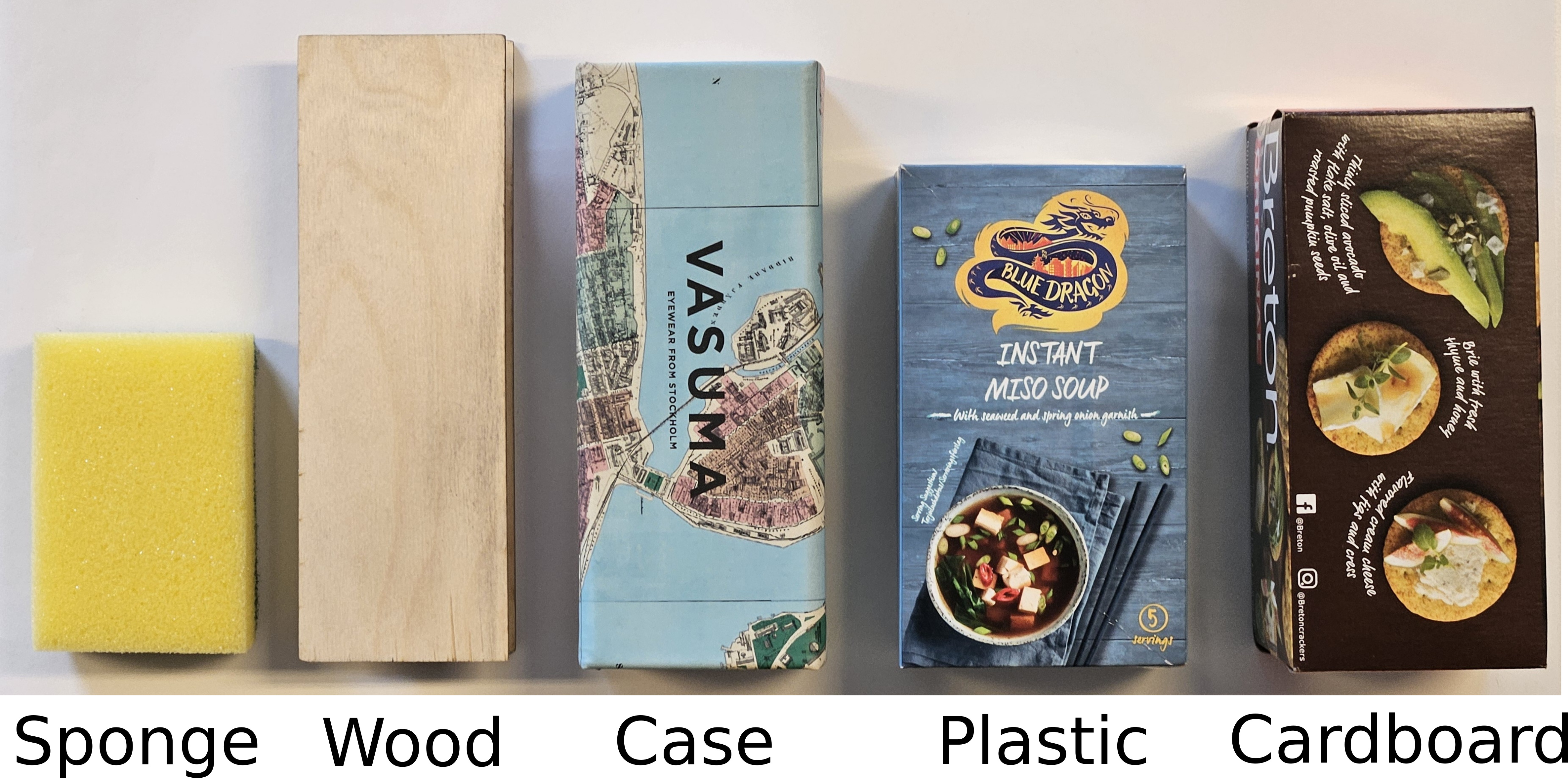}
    \caption{Test objects, the sponge is highly deformable, the wood object has a plywood surface and is rigid. The spectacle case has a synthetic leather material. The plastic object is a plastic covered cardboard box. The cardboard box has a paper finish. }
    \label{fig:objects}
    \vspace*{-0.5cm}
\end{figure}

In this section, we present the experimental validation of the proposed hardware and methods. Various objects detailed in Fig. \ref{fig:objects} are used for the experiments. The dimensions and weights of these objects are listed in Tab. \ref{tab:object_properites}. Depending on the specific experiment, only a subset of these objects may be evaluated. The experiments are organized as follows:
\begin{itemize}
\item \textbf{Section \ref{sec:gripper_results}:} Validation of the grasp controller.
\item \textbf{Section \ref{sec:vel_sensors_results}:} Testing of the relative velocity sensor on different surfaces using the setup shown in Fig. \ref{fig:testRig}.
\item \textbf{Section \ref{sec:friction_est_results}:} Evaluation of the contact estimation procedure.
\item \textbf{Section \ref{sec:slip_avoidance} to   \ref{sec:final_demo}:} Testing of the slip-aware controllers.
\end{itemize}

The experiments were conducted using a laptop equipped with an Intel i7-1185G7 processor and 32 GB of RAM. The software \textcolor{blue1}{and algorithms} was implemented in ROS Noetic, and the UR10 robot was controlled using the Universal Robots ROS driver. The built-in task-based trajectory controllers were utilized for the robot's movement.

\subsection{Gripper Performance}\label{sec:gripper_results}

{
\renewcommand{\arraystretch}{1.2}
\begin{table}[pbb]
\scriptsize
\centering
\caption{Gripper inner loop grasp controller parameters. \label{tab:tunabel}}
\begin{tabular}{c|c||c|c||c|c}
\thickhline
\textbf{Parameter} & \textbf{Value} & \textbf{Parameter} & \textbf{Value} & \textbf{Parameter} & \textbf{Value} \\ \thickhline
$k_P$ & 2 & $k_I$ & 100 & $I_\textrm{max}$ & 10 \\ \hline
$\gamma$ & 5 & $k_f$ & 3.9166 & $t_s$ & $2\mathrm{e}{-3}$\ \\\hline
\end{tabular}
\end{table}
}

\begin{figure}
    \centering
    \smallskip 
    \includegraphics[width=0.9\columnwidth]{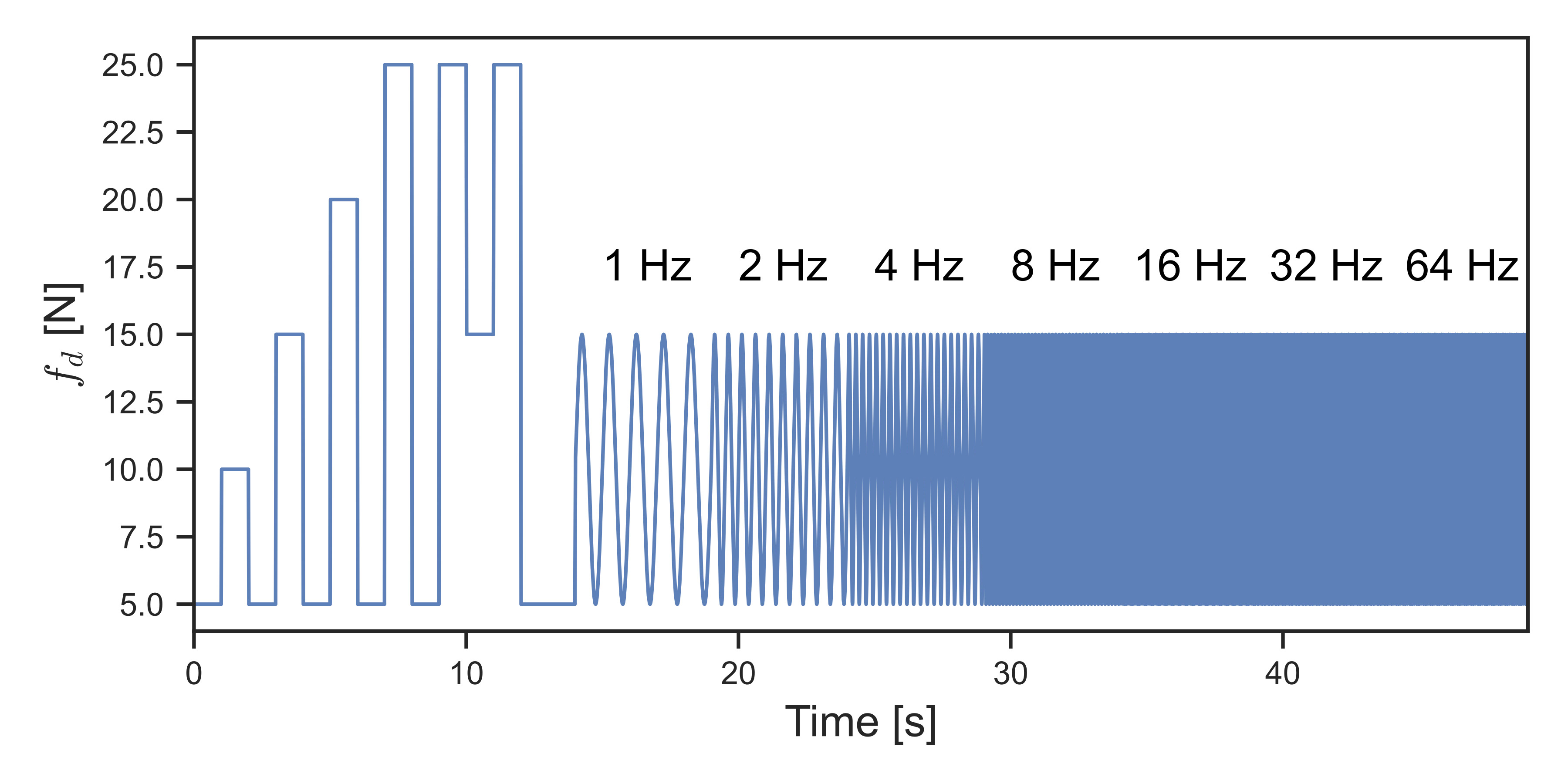}
    \caption{Force trajectory to test gripper performance.}
    \label{fig:force_traj}
    \vspace*{-0.5cm}
\end{figure}

The parameters for the inner loop controller were tuned through experimentation, and the values used are presented in Tab. \ref{tab:tunabel}.
The performance and stability of the gripper were evaluated by following a predefined force trajectory while grasping a variety of objects (see Tab. \ref{tab:object_properites} and Fig. \ref{fig:objects}). The force trajectory consisted of step and sinusoidal signals with varying amplitudes and frequencies, as illustrated in Fig. \ref{fig:force_traj}. 

\begin{figure}
    \centering
    \smallskip 
    \includegraphics[width=0.9\columnwidth]{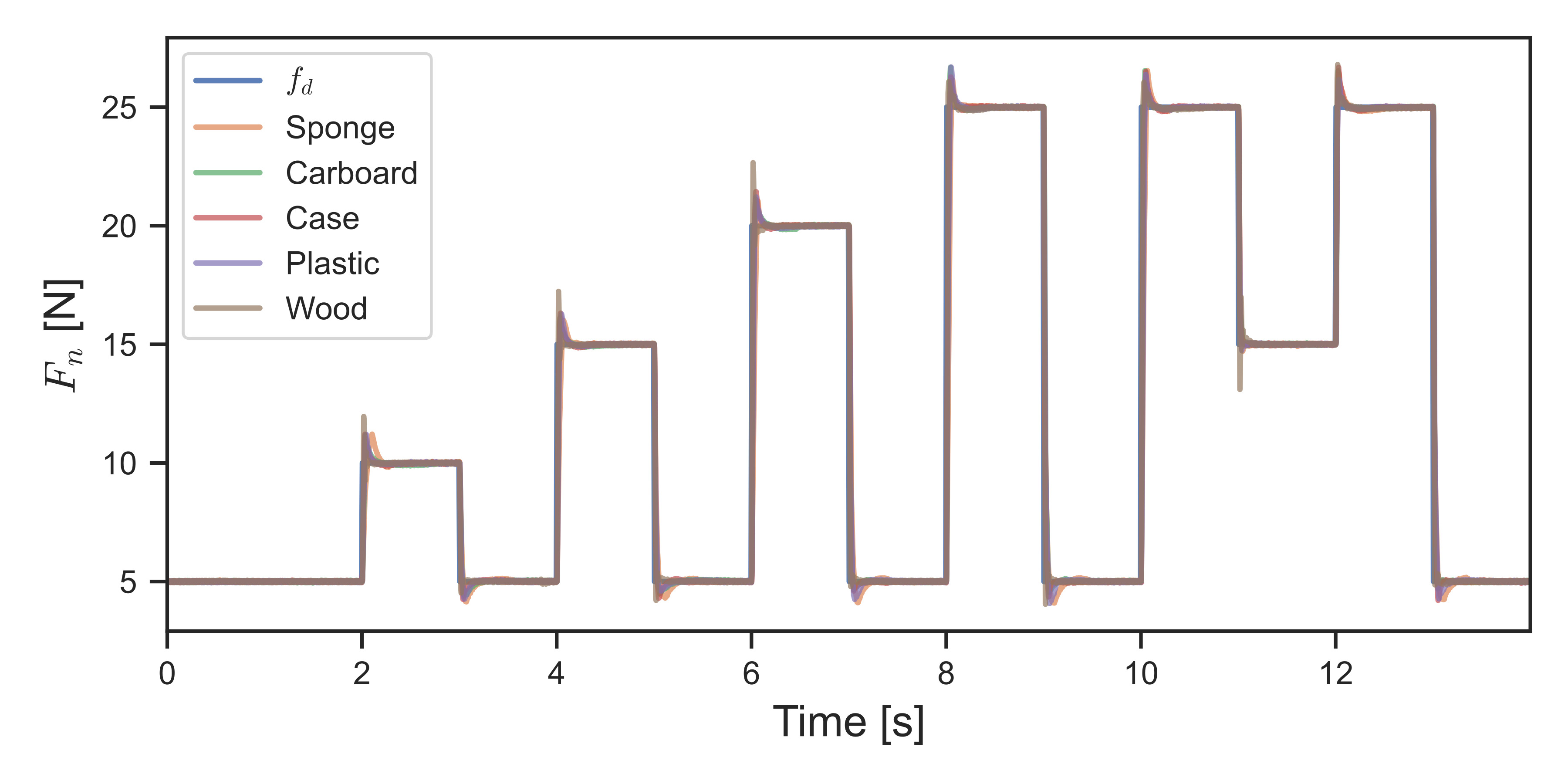}
    \caption{Step response of the gripper, one sample per object.}
    \label{fig:step_resonse}
    \vspace*{-0.5cm}
\end{figure}

The gripper's response time was measured during the transition between a force of 5 N and 25 N, with each object tested 10 times. A representative sample from each object is shown in Fig. \ref{fig:step_resonse}. The key metrics for evaluating the step response are: 
\begin{itemize}
\item \textbf{Rise Time ($t_r$):} The time it takes for the force to rise from 10 \% to 90 \% of the desired force change.  
\item \textbf{Half-Rise Time ($t_{50\%}$):} The time taken to reach 50\% of the desired force change.
\item \textbf{Maximum Overshoot ($M_p$):} The peak force value above the desired value, indicating how much the force exceeded the target.
\item \textbf{Settling Time ($t_s$):} The time required for the force to stabilize within an error band $\pm 2\%$ around the desired force, which corresponds to a range of $\pm 0.4$ N.
\end{itemize}
The statistical results for these metrics across the 10 trials for each object are summarized in Tab. \ref{tab:step_resopnse}. The data presented in Fig. \ref{fig:step_resonse} and Tab. \ref{tab:step_resopnse}  demonstrate that the gripper can quickly achieve the desired force across various object properties. Response times were generally faster for rigid objects compared to soft, deformable ones, as the fingers of the gripper need to physically move and compress the deformable objects further to achieve the desired grasp force. The wood object is much stiffer than the gripper and provides an approximation of the control characteristics where only the gripper's stiffness matters.

{
\renewcommand{\arraystretch}{1.2}
\begin{table}[pbb]
\scriptsize
\centering
\caption{Step response for 5 to 25 N, mean $\mu$ and standard deviation $\sigma$. \label{tab:step_resopnse}}
\resizebox{\columnwidth}{!}{%
\begin{tabular}{c|c|c|c|c}
\thickhline
Object type & $t_r$ $(\mu, \sigma)$ [s]  & $t_{50\%}$ $(\mu, \sigma)$ [s] & $M_p$ $(\mu, \sigma)$ [N] & $t_s$ $(\mu, \sigma)$ [s]\\ \hline
\thickhline
Sponge & (0.0345, 0.0021) & (0.0242, 0.0015) & (1.4556, 0.1166) & (0.0999, 0.0129) \\ \hline
Cardboard box & (0.0175, 0.0018) & (0.0137, 0.0010) & (1.8778, 0.1850) & (0.0747, 0.0088) \\ \hline
Case & (0.0220, 0.0012) & (0.0136, 0.0006) & (1.4398, 0.1479) & (0.0787, 0.0111) \\ \hline
Plastic & (0.0224, 0.0014) & (0.0157, 0.0009) & (1.6043, 0.2162) & (0.0970, 0.0034) \\ \hline
Wood & (0.0086, 0.0010) & (0.0093, 0.0008) & (1.7614, 0.2097) & (0.06877, 0.0154) \\ \hline
\end{tabular}
}
\end{table}
}

\begin{figure}
    \centering
    \smallskip 
    \includegraphics[width=0.9\columnwidth]{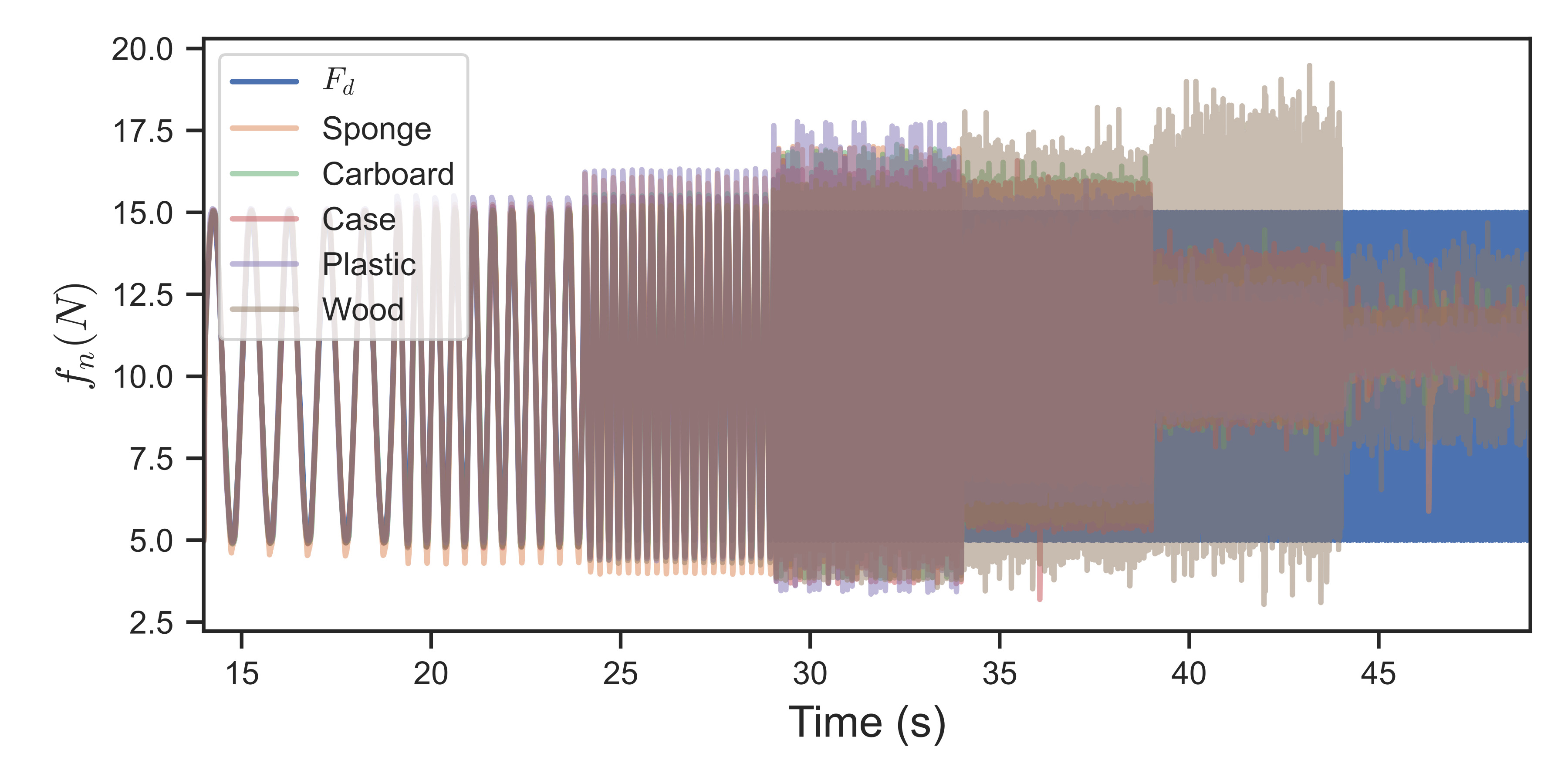}
    \caption{Tracking sinusoidal grasp force trajectory, one sample per object.}
    \label{fig:bode_sin}
    \vspace*{-0.5cm}
\end{figure}

\begin{figure}
    \centering
    \smallskip 
    \includegraphics[width=0.9\columnwidth]{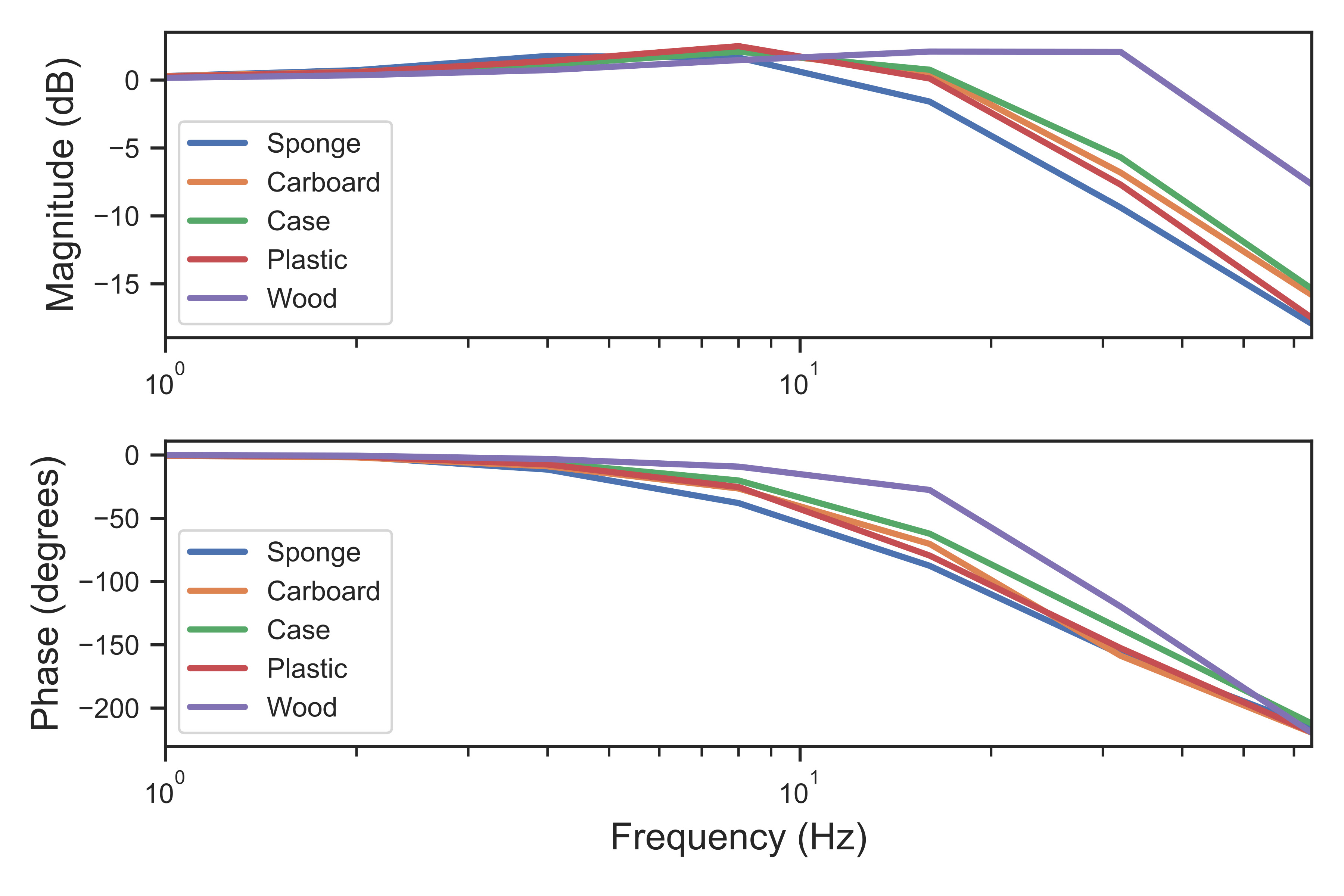}
    \caption{Bode plot, shows an average of 10 runs for each object.}
    \label{fig:bode_plot}
    \vspace*{-0.5cm}
\end{figure}

To evaluate the stability and performance of the gripper under time-varying target forces, the gripper was subjected to a sequence of sinusoidal signals with frequencies ranging from 1 to 64 Hz, as illustrated in Fig. \ref{fig:bode_sin}. The response was analyzed by computing the amplitude and phase for each frequency. The average results from 10 runs are presented in a Bode plot, see Fig. \ref{fig:bode_plot}. 
Figs.  \ref{fig:bode_sin} and \ref{fig:bode_plot} reveal that the magnitude increases at certain frequencies but remains bounded, indicating that stability is maintained throughout the testing range. As expected, the magnitude decreases at higher frequencies, and the phase shifts accordingly. Notably, the phase exceeds $180^\circ$ when the 64 Hz signal is applied. This phase shift can be attributed to the operational rates \textcolor{blue1}{of the controller}. Given that the inner loop controller operates at 500 Hz, a delay of $1/500$ seconds for a 64 Hz signal would result in a phase shift of $46.08^\circ$, which would account for the shift beyond $180^\circ$. Overall, the combined analysis of the step response and Bode plot demonstrates that the inner loop controller is both stable and responsive.

\subsection{Relative Velocity Sensor}\label{sec:vel_sensors_results}

\begin{figure}
    \centering
    \smallskip 
    \includegraphics[width=0.7\columnwidth]{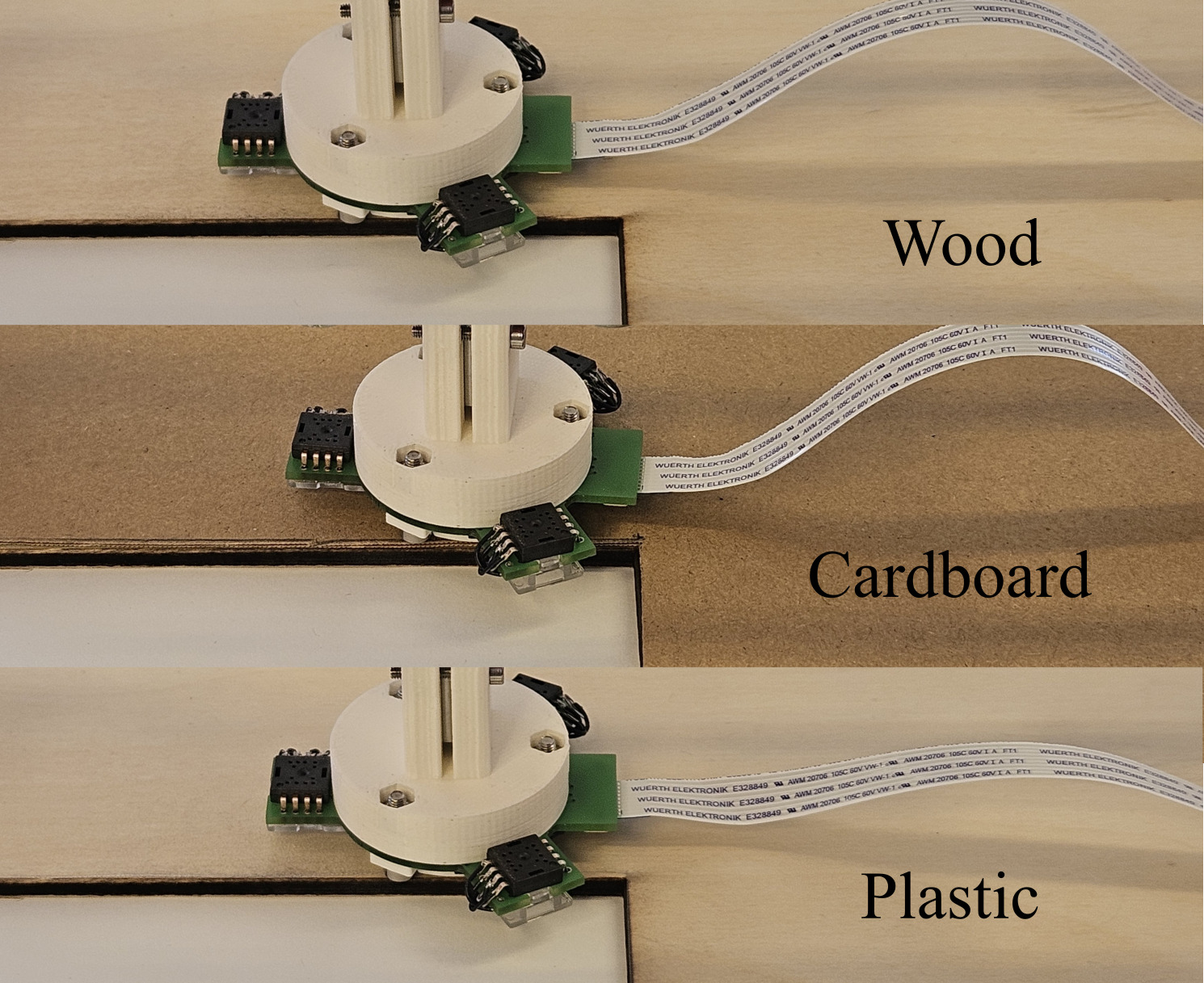}
    \caption{Three different surfaces: Wood, cardboard and clear plastic film. The figure also shows the cut out for testing the sensor rejection system.}
    \label{fig:surfaces}
    \vspace*{-0.5cm}
\end{figure} 

{
\renewcommand{\arraystretch}{1.2}
\begin{table}[pbb]
\scriptsize
\centering
\caption{Tracking error [mm] \& [deg], mean $\mu$ and standard deviation $\sigma$. \label{tab:tracking_error}}
\begin{tabular}{c|c|c|c}
\thickhline
Surface type and test & $x$ $(\mu, \sigma)$ & $y$ $(\mu, \sigma)$ & $\theta$ $(\mu, \sigma)$\\ \hline
\thickhline
Wood (Linear, 100 mm)  & (-2.80, 0.81) & (-3.69, 0.58) & (0.62, 2.31) \\ \hline
Cardboard (Linear, 100 mm)  & (-2.17, 0.37) & (0.47, 0.73) & (0.76, 0.91) \\ \hline
Plastic (Linear, 100 mm)  & (-1.79, 0.38) & (0.96, 0.27) & (-1.75, 0.55) \\ \hline
\thickhline
Wood (Rotational, 180 deg)& (0.55, 0.19) &  (1.82, 0.09) & (0.61, 0.56) \\ \hline
Cardboard (Rotational, 180 deg)& (-0.02, 0.22) & (0.51, 0.29) & (1.01, 0.56) \\ \hline
Plastic (Rotational, 180 deg)& (0.80, 0.29) & (2.18, 0.52) & (-0.50, 0.79)\\ \hline
\thickhline
Wood (Rejection, 100 mm) & (-2.21, 0.44) & (-3.44, 0.54) & (1.99, 0.76) \\ \hline
Cardboard (Rejection, 100 mm)& (-1.75, 0.34) & (0.44, 0.54) & (0.50, 0.66) \\ \hline
Plastic (Rejection, 100 mm) & (-2.15, 0.27) & (1.98, 0.35) & (-3.43, 0.63) \\ \hline
\end{tabular}
\end{table}
}
{
\renewcommand{\arraystretch}{1.2}
\begin{table}[pbb]
\scriptsize
\centering
\caption{Tracking error combined motion (100 mm \& 180 deg) , mean $\mu$ and standard deviation $\sigma$. \label{tab:tracking_error_combined}}
\begin{tabular}{c|c|c}
\thickhline
Surface type and test & distance $(\mu, \sigma)$ & $\theta$ $(\mu, \sigma)$\\ \hline
\thickhline
Wood  & (-4.24, 1.70) & (4.64, 4.42) \\ \hline
Cardboard  & (-2.05, 1.06) & (3.47, 1.47) \\ \hline
Plastic  & (0.94, 1.36) & (3.65, 2.19) \\ \hline
\end{tabular}
\end{table}
}

To estimate the accuracy of the velocity sensor in a controlled environment, four different tests were conducted using the test rig shown in Fig. 
 \ref{fig:testRig}. These tests were designed to evaluate the sensor's performance under various conditions: 
\begin{enumerate}
    \item \textbf{Linear:} The sensor was moved 100 mm across the surface in a straight line along the sensor's x-direction.
    \item \textbf{Rotation:} The sensor was rotated $180^\circ$ clockwise without any tangential movement.
    \item \textbf{Rejection:} The sensor was moved 100 mm linearly along the x-direction, but at the 50 mm mark, one of the sensors was positioned outside the surface, as depicted in Fig. \ref{fig:surfaces}. 
    \item \textbf{Combined:} The sensor underwent both 100 mm linear movement and 180-degree clockwise rotation simultaneously.
\end{enumerate}
During these experiments, the sensor's estimated velocity was integrated to calculate the displacement, which was then compared to the known displacement from the test rig. Each experiment was repeated 10 times across the three surfaces shown in Fig. \ref{fig:surfaces}. The results are presented in Tab. \ref{tab:tracking_error} and \ref{tab:tracking_error_combined}. 

The findings indicate that the sensors tracked well across all tested surfaces, with the velocity integrating to approximately a 2\% displacement error over a 100 mm distance. The estimation of angular velocity typically resulted in less than 1\% displacement error. Notably, the sensor's performance was not significantly disrupted when one of the sensors was outside the object, as demonstrated by the rejection test. Additionally, the sensor was capable of accurately estimating both angular and linear velocities simultaneously.

\subsection{Estimation of Contact Properties}\label{sec:friction_est_results}

\begin{figure}
    \centering
    \smallskip \includegraphics[width=\columnwidth]{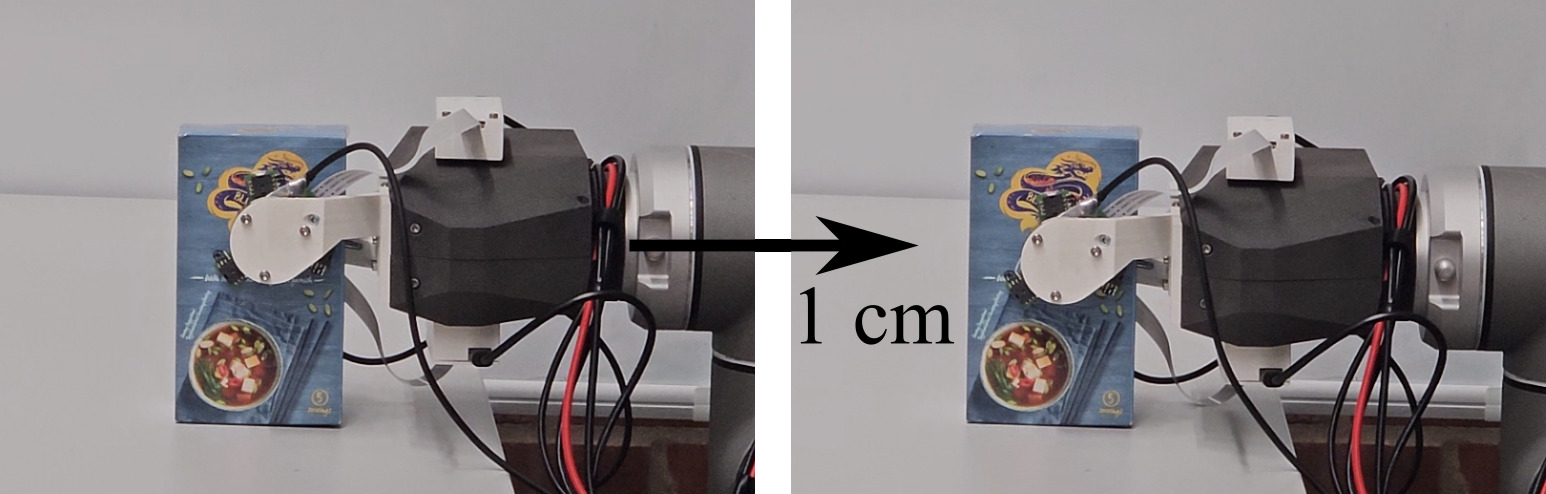}
    \caption{Linear exploration movement for friction estimation.}
    \label{fig:linear_exploration}
    \vspace*{-0.5cm}
\end{figure}

\begin{figure}
    \centering
    \smallskip 
    \includegraphics[width=\columnwidth]{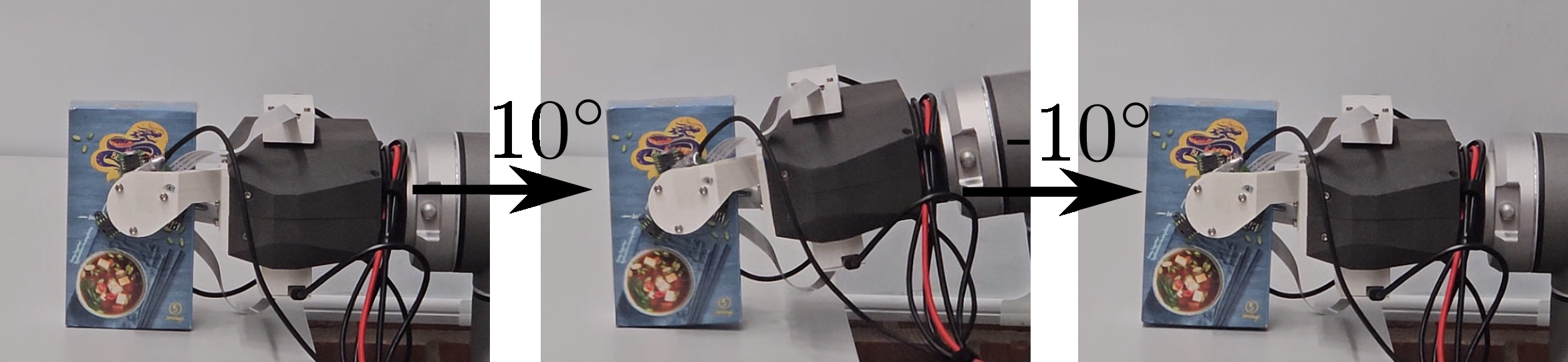}
    \caption{Rotational exploration movement for contact radius estimation.}
    \label{fig:rot_exploration}
    \vspace*{-0.5cm}
\end{figure}

The contact estimation procedure was evaluated using four different objects: wood, a spectacle case, plastic, and cardboard, see Fig. \ref{fig:objects}. The gripper, mounted on a UR10 robot (as shown in Fig. \ref{fig:linear_exploration}), grasped each object with a 5 N force, see Tab. \ref{tab:est_paramters} the contact exploration parameters. The procedure began with a linear exploration to estimate $\mu_s$, $\mu_c$ and $\mu_v$, see Fig. \ref{fig:linear_exploration}. Following the linear exploration, the gripper executed the rotational exploration, see Fig. \ref{fig:rot_exploration}. 
The estimation process, as detailed in Section \ref{sec:contact_est}, was repeated 10 times for each object, with the mean and standard deviations 
 presented in Tab. \ref{tab:friction_estimation}.

\renewcommand{\arraystretch}{1.2}
\begin{table}[pbb]
\scriptsize
\centering
\caption{Contact exploration parameters. \label{tab:est_paramters}}
\begin{tabular}{c|c||c|c||c|c}
\thickhline
\textbf{Parameter} & \textbf{Value} & \textbf{Parameter} & \textbf{Value} & \textbf{Parameter} & \textbf{Value} \\ \thickhline
$f_{g}$ & 5 N&  $t_{1}$ & 0.2 s & $t_{2}$ & 0.8 s  \\ \hline
$d_e$ & 10 mm &  $\theta_e$ & $10^\circ$ & $t_\theta$ & 0.5 s \\ \thickhline
\end{tabular}
\end{table}

{
\renewcommand{\arraystretch}{1.2}
\begin{table}[pbb]
\scriptsize
\centering
\caption{Estimation of contact properties, mean $\mu$ and standard deviation $\sigma$, $\cdot^{(i)}$ is for sensor $i$. \label{tab:friction_estimation}}
\resizebox{\columnwidth}{!}{%
\begin{tabular}{c|c|c|c|c}
\thickhline
Object & ($\mu_s^{(1)}$, $\mu_c^{(1)}$, $\mu_v^{(1)}$) & $r^{(1)}$ [mm] & ($\mu_s^{(2)}$, $\mu_c^{(2)}$, $\mu_v^{(2)}$) & $r^{(2)}$ [mm] \\ \hline
\thickhline
Cardboard ($\mu$) & (0.565, 0.538, 0.578) & 6.12 & (0.540, 0.511, 0.379) & 5.84 \\ 
Cardboard ($\sigma$) & (0.046, 0.072, 1.584) & 1.58 & (0.027, 0.049, 0.908) & 1.65\\ \hline
Case  ($\mu$) & (0.533, 0.511, 1.106) & 6.31 & (0.507, 0.479, 1.914) & 5.04\\ 
Case ($\sigma$) & (0.025, 0.028, 1.424) & 1.12 & (0.068, 0.033, 1.456) & 1.96 \\ \hline

Plastic ($\mu$)  & (0.411, 0.366, 11.694) & 7.64 & (0.399, 0.363, 3.60) & 7.40 \\ 
Plastic ($\sigma$)  & (0.014, 0.015, 1.884) & 1.58 & (0.020, 0.016, 0.647) & 1.68 \\ \hline
Wood ($\mu$) & (0.499, 0.444, -1,123) & 7.92 & (0.507, 0.432, -1.649) & 3.91\\ 
Wood  $\sigma$) & (0.023, 0.018, 0.885) & 1.49 & (0.065, 0.034, 0.735) & 2.94 \\ \hline

\end{tabular}
}
\end{table}
}
The experiments demonstrate that the gripper and sensor combination can estimate friction coefficients and contact radius in just 2 seconds of exploration. The procedure relies solely on in-hand sensing, without depending on the robot's forward kinematics. As shown in Tab. \ref{tab:friction_estimation}, both sensors produce similar friction coefficient estimates for the same material and object. The results reveal variations in friction coefficients across different objects, with cardboard exhibiting the highest friction and plastic the lowest. Notably, the viscous friction varies significantly between objects, and it is important to mention that the gripper's average velocity during exploration is only 0.0125 m/s and \textcolor{blue}{that the viscus friction estimate might only be accurate within the explored velocity range.}

\begin{figure}
    \centering
    \smallskip 
    \includegraphics[width=0.65\columnwidth]{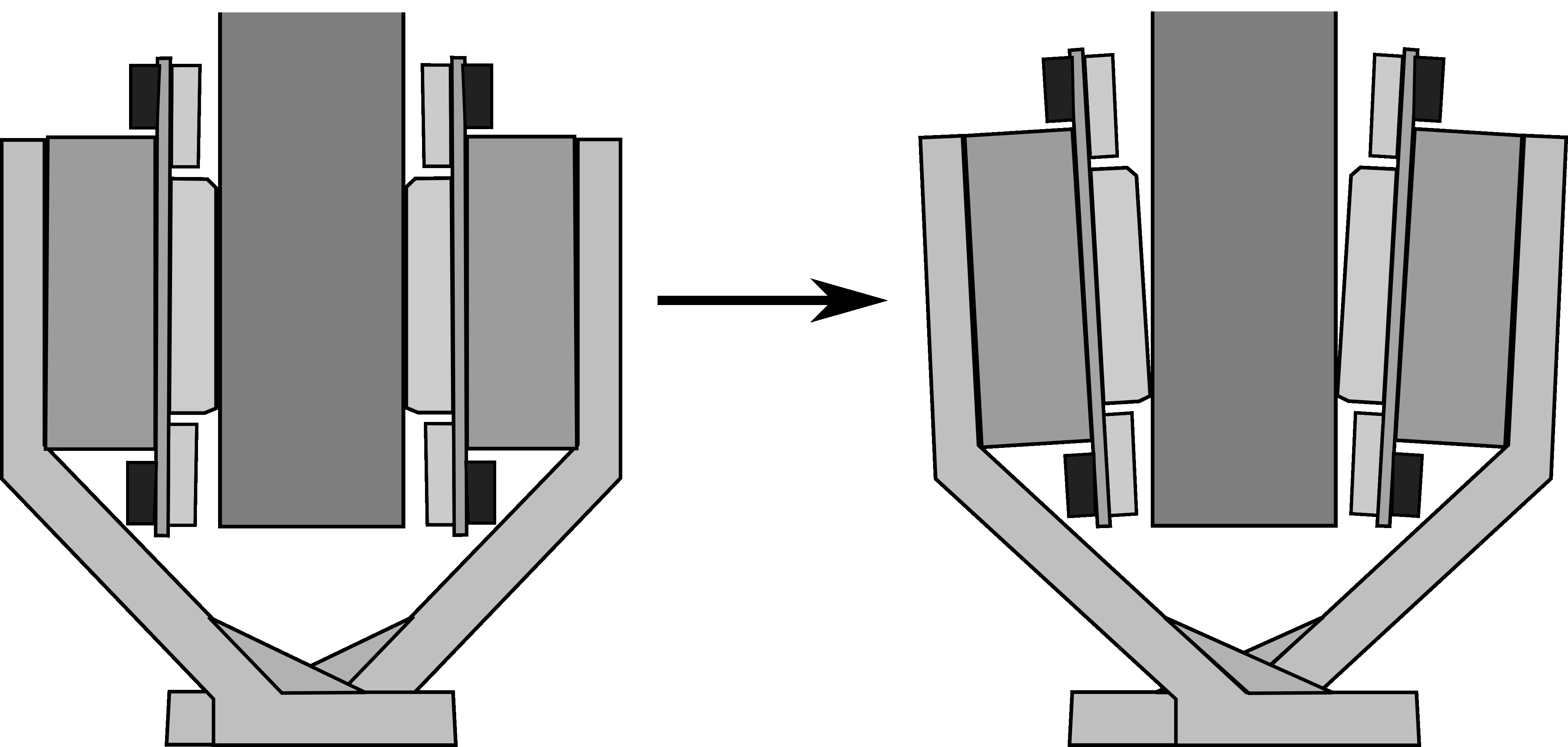}
    \caption{Illustration of finger flex under higher grasp force.}
    \label{fig:finger_flex}
    \vspace*{-0.5cm}
\end{figure}

The actual contact surface radius is 15 mm, and under the assumption of uniform pressure distribution, the equivalent rim contact radius would be 10 mm. However, the estimated rim contact radius is lower than the theoretical value for all objects. This discrepancy is mainly attributed to two factors: object surface deformation and finger flexion. Each object deforms differently under grasping, and since the gripper's contact pads are rigid, the assumption of uniform pressure distribution may be weak. The finger flexion during grasping is attributed as the primary factor influencing the results. Because the contact surface is rigid, even slight misalignment due to finger flexion or other causes can significantly impact the pressure distribution, as illustrated in Fig. \ref{fig:finger_flex}. This phenomenon particularly affects rigid objects more than deformable ones, as finger flexion reduces the contact area and limits the torque that can be generated.

\subsection{Slip-Avoidance}\label{sec:slip_avoidance}
{
\renewcommand{\arraystretch}{1.2}
\begin{table}[pbb]
\scriptsize
\centering
\caption{Contact properties used in sections \ref{sec:slip_avoidance} to \ref{sec:rot_slip}. \label{tab:object_est_last}}
\begin{tabular}{c|c|c|c|c}
\thickhline
Object & $\mu_c^{(1)}$ & $r^{(1)}$ &  $\mu_c^{(2)}$ & $r^{(2)}$ \\ \hline
\thickhline
Cardboard  & 0.514 & 4.84 & 0.473 & 6.30\\ \hline
Case  & 0.538 & 5.72 & 0.467 & 3.41 \\ \hline
Plastic  & 0.355 & 7.69 & 0.365 & 5.75 \\ \hline
Wood  & 0.459 & 7.15 & 0.404 & 1.90 \\ \hline
\end{tabular}
\end{table}
}

\begin{figure}
    \centering
    \smallskip 
    \includegraphics[width=0.99\columnwidth]{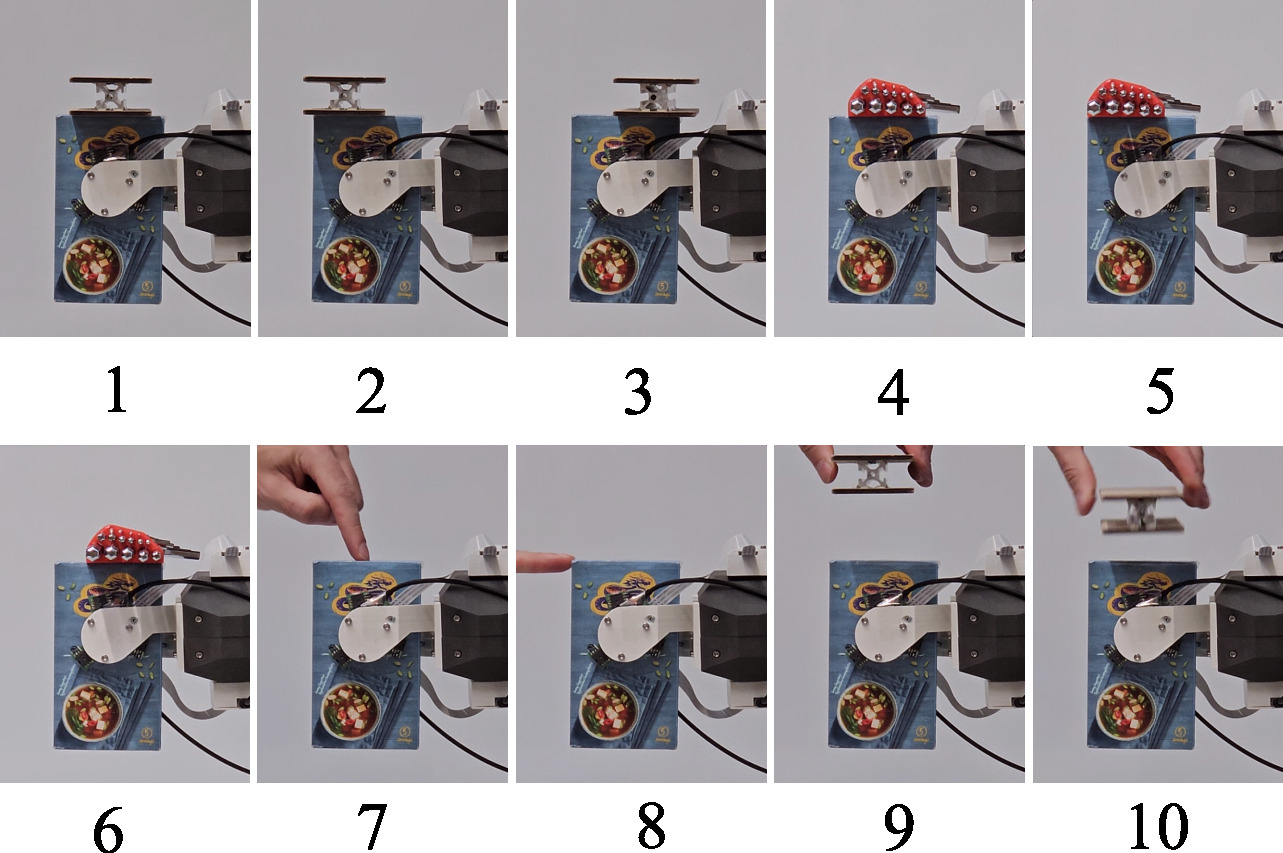}
    \caption{Disturbances to test the slip-avoidance controller, position 9 and 10 in the figure illustrates dropping a weight on the grasped object. The corresponding forces are presented in Fig. \ref{fig:hold_mode_plot}.}
    \label{fig:hold_mode_img}
    \vspace*{-0.5cm}
\end{figure}

The slip-avoidance controller is designed to prevent slippage by dynamically adjusting the grasp force in response to disturbances. To evaluate its effectiveness, an object was grasped while various disturbances were introduced, as depicted in Fig. \ref{fig:hold_mode_img}. The most recently estimated contact properties from the experiments in Section \ref{sec:friction_est_results} were used, see Tab. \ref{tab:object_est_last}. The parameters for the slip-aware controllers, tuned through experimentation, are presented in Tab. \ref{tab:slip_paramters}. 

\begin{figure}
    \centering
    \smallskip 
    \includegraphics[width=0.9\columnwidth]{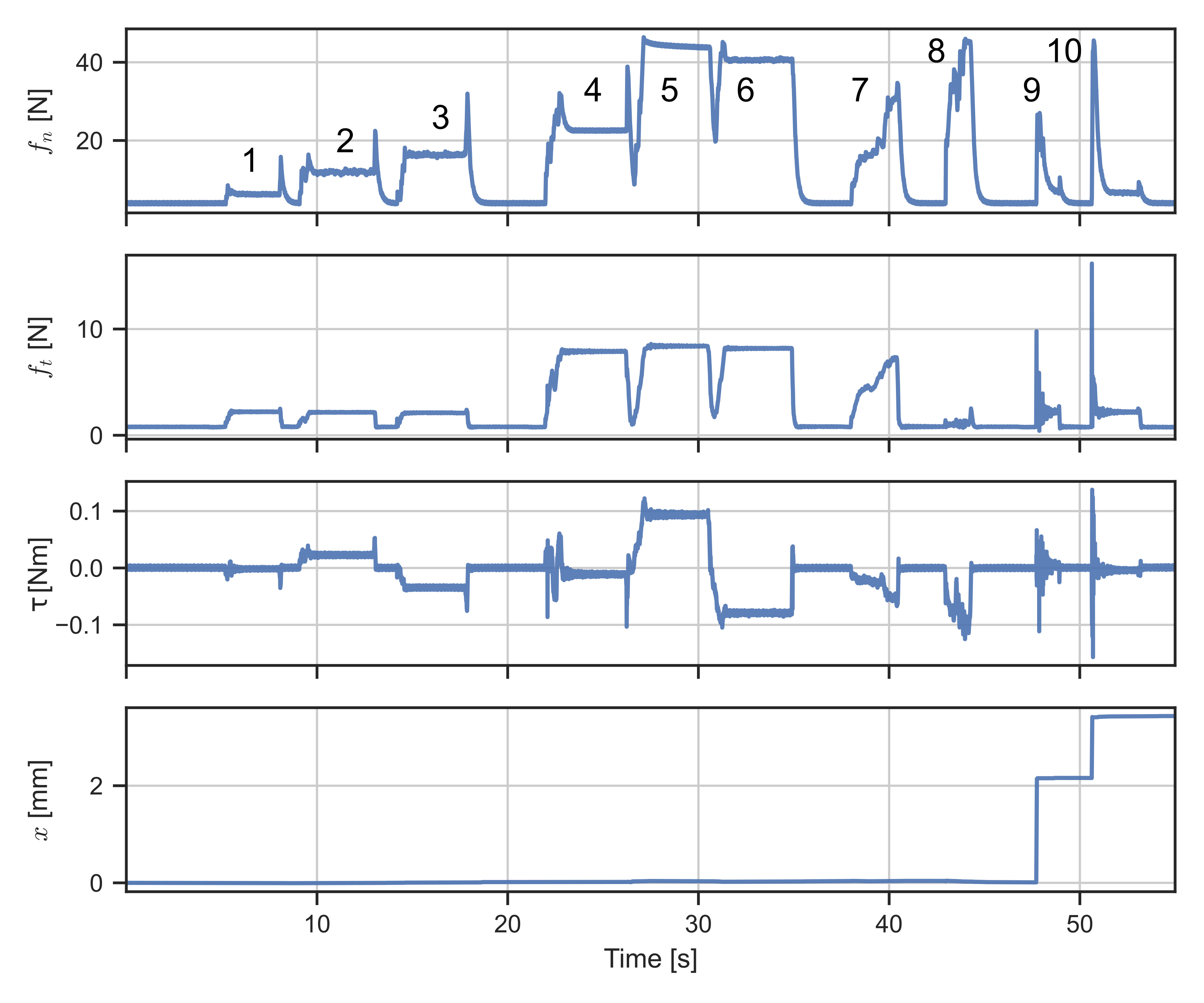}
    \caption{Forces and tangential velocity from the slip-avoidance controller.}
    \label{fig:hold_mode_plot}
    \vspace*{-0.4cm}
\end{figure}

The first disturbance involved adding a 141 g weight on top of the grasped object, followed by placing the weight on either side to create a torque disturbance. The resulting forces are shown in Fig. \ref{fig:hold_mode_plot}. When the weight was placed on the side of the object, the grasping force $f_n$ increased to compensate for the additional torque. Subsequently, a heavier weight of 727 g was placed on the object, as illustrated in positions 4 to 6 in Fig. \ref{fig:hold_mode_img}. The slip-avoidance controller responded by increasing the grasp force to compensate for the added weight. When the heavier weight was placed on the corner of the object, the grasp force reached its maximum, as seen in Fig. \ref{fig:hold_mode_plot}. Following this, a manual disturbance was introduced, as shown in positions 7 and 8 in Fig. \ref{fig:hold_mode_img}. Finally, to enforce slippage, the 141 g weight was dropped onto the grasped object, causing a sudden acceleration, depicted in positions 9 and 10 in \textcolor{blue}{Fig.} \ref{fig:hold_mode_img}. This sudden impact resulted in a spike in the grasp force until the object decelerated. The corresponding linear displacement $x$, as measured by the velocity sensors, is also presented in \textcolor{blue}{Fig. }\ref{fig:hold_mode_plot}. The experiment demonstrates that the slip-avoidance controller effectively adjusts the grasp force in real-time to prevent slippage, even under varying and sudden disturbances.

\renewcommand{\arraystretch}{1.2}
\begin{table}[pbb]
\scriptsize
\centering
\caption{Slip-aware control parameters. \label{tab:slip_paramters}}
\begin{tabular}{c|c||c|c||c|c}
\thickhline
\textbf{Parameter} & \textbf{Value} & \textbf{Parameter} & \textbf{Value} & \textbf{Parameter} & \textbf{Value} \\ \thickhline
$\gamma_s$ & 2 &  $k_{P,v}$ & 2 & $k_{P,\omega}$ & 2  \\ \hline
$\gamma_{h}$ & 1.4 &  $k_{P,l}$ & 50 & $k_{P,\tau}$ & 2 \\ \hline
$\alpha$ & 0.95  & $k_{I,l}$ & 1000 &  $k_{I,\tau}$ & 40  \\ \hline
 $f_{n, \textrm{min}}$ & 0.9 N & $ k_{D,l}$ & 0.1  & $k_{D,\tau}$ & 0.05 \\ \hline
$f_{n, s, \textrm{min}}$ & 2 N &  $k_{P, s}$ & 100 & $k_{P,h}$ & 100 \\ \thickhline
\end{tabular}
\end{table}

\subsection{Linear Slip Control}\label{sec:linear_slip}

The linear slippage controller was evaluated through four different experiments for each of the objects listed in Tab. \ref{tab:object_est_last}. The estimated contact properties and control parameters used are provided in Tab. \ref{tab:object_est_last} and \ref{tab:slip_paramters}, respectively. The experiments were designed to assess the controller's performance across a range of target displacements and slippage velocities:
\begin{enumerate}
    \item 20 mm linear displacement in 2 s.
    \item 20 mm linear displacement in 5 s.
    \item 40 mm linear displacement in 2 s.
    \item 40 mm linear displacement in 5 s.
\end{enumerate}
\begin{figure}
    \centering
    \begin{subfigure}[b]{0.49\columnwidth}
        \includegraphics[width=\columnwidth]{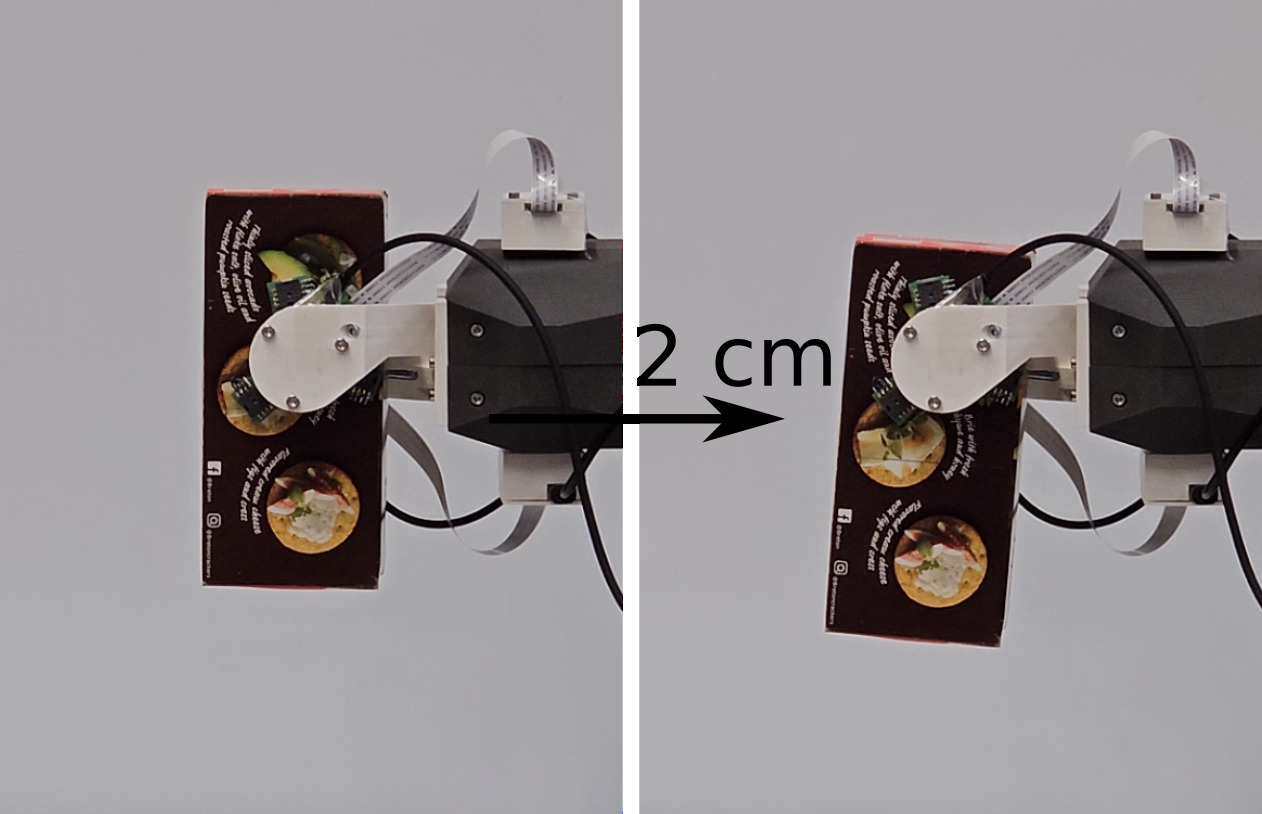}
        \caption{2 cm linear slip.}
        \label{fig:lin_slip_2cm}
    \end{subfigure}
    \begin{subfigure}[b]{0.49\columnwidth}
        \includegraphics[width=\columnwidth]{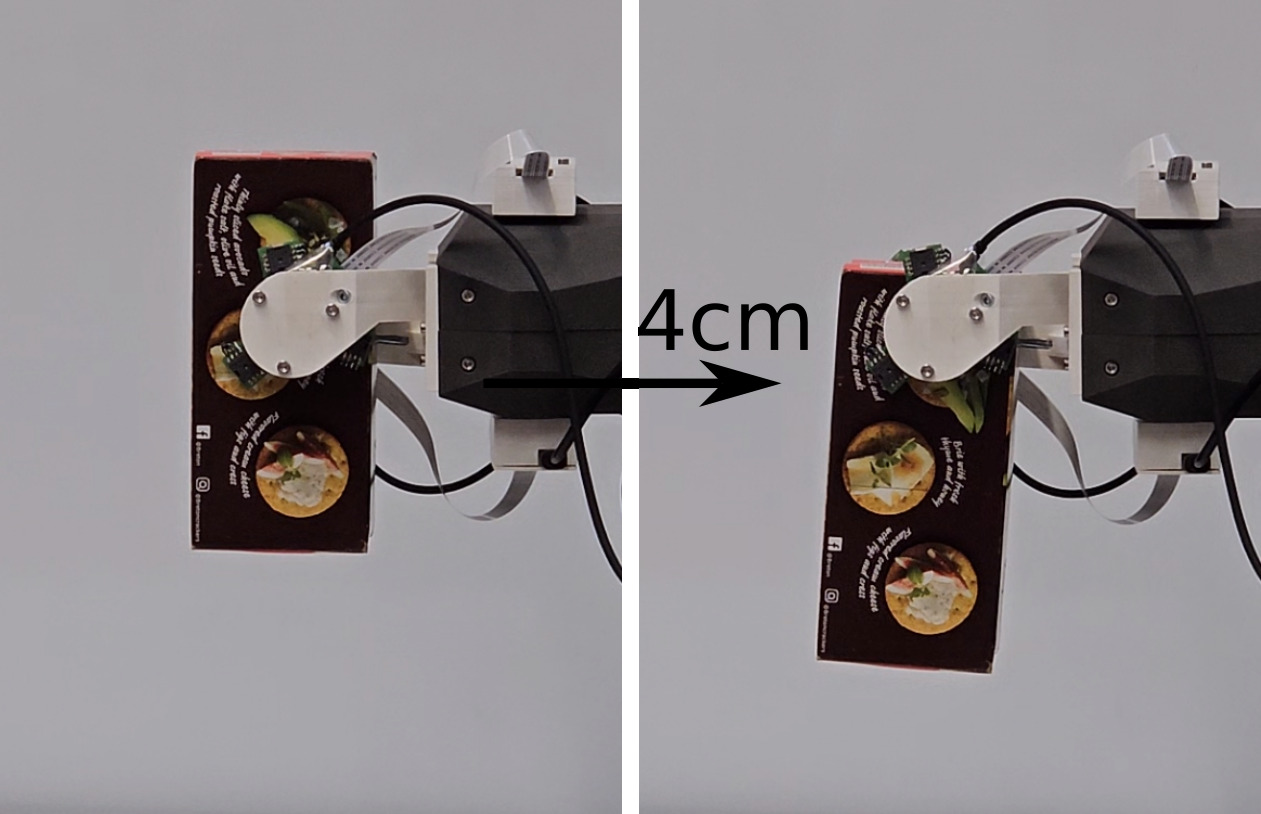}
        \caption{4 cm linear slip.}
        \label{fig:lin_slip_4cm}
    \end{subfigure}
    \caption{Snapshots of linear slippage experiment with cardboard object.} 
    \label{fig:linear_slip}        
    \vspace*{-0.5cm}
\end{figure}
\begin{figure}
    \centering
    \smallskip 
    \includegraphics[width=0.9\columnwidth]{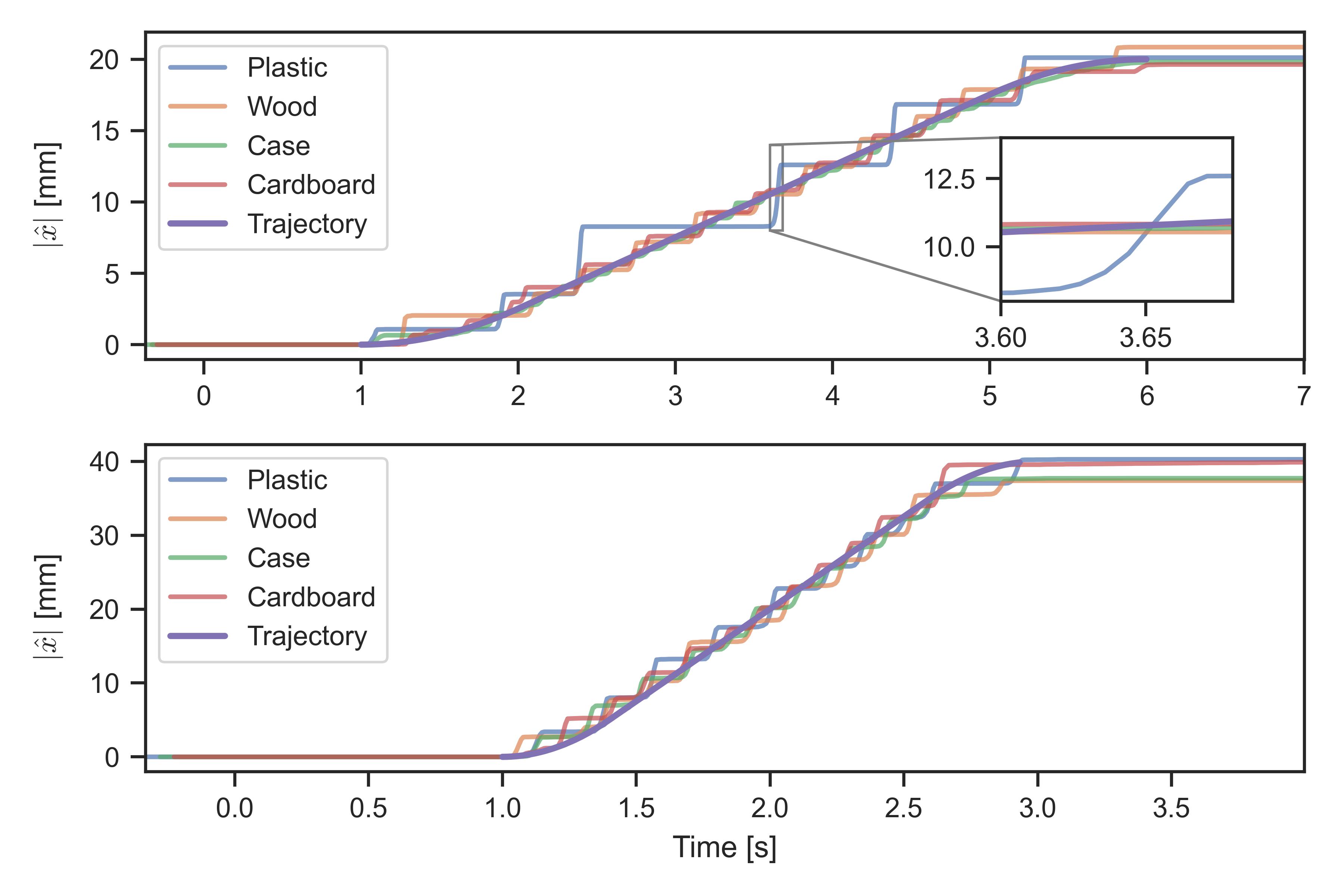}
    \caption{Linear slippage: top plot follows a 20 mm trajectory for 5 seconds, bottom plot shows a 40 mm trajectory for 2 second. $|\hat{x}|$ is the estimated displacement by the planar velocity sensors.}
    \label{fig:linear_plot_2cm}
    \vspace*{-0.5cm}
\end{figure}

Snapshots from these experiments are shown in Fig. \ref{fig:lin_slip_2cm} and \ref{fig:lin_slip_4cm}. The controller followed a trapezoidal trajectory, with both the trajectory and internal displacement estimation presented in Fig. \ref{fig:linear_plot_2cm} from two of the tests. The results demonstrate that the linear slip controller can achieve the desired final displacement; however, it experiences the stick-slip phenomenon during trajectory following.

\begin{figure}
    \centering
    \smallskip 
    \includegraphics[width=0.9\columnwidth]{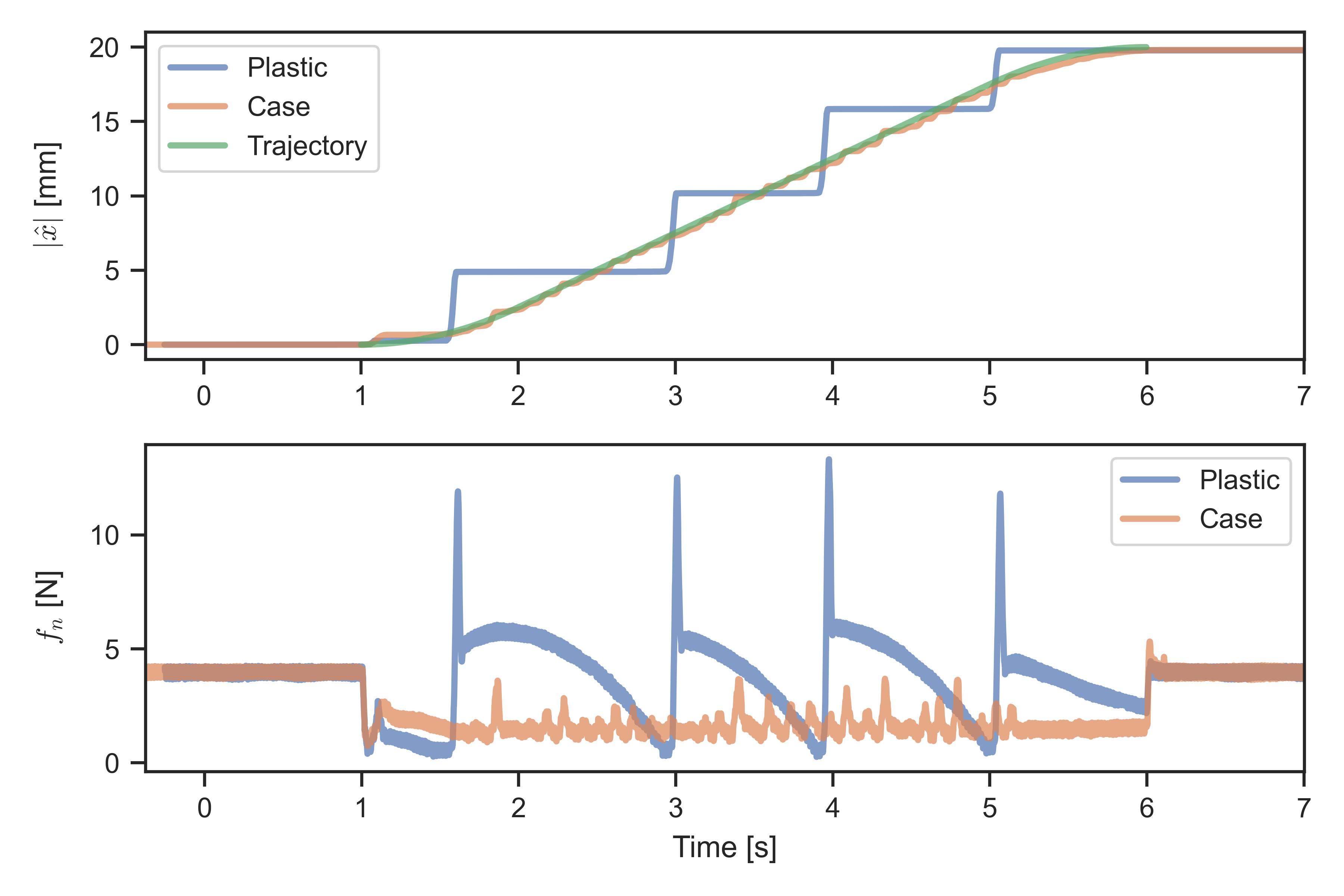}
    \caption{Highlights the two extremes in stick and slip phenomenon between the plastic and case object, with the associated measured grasp forces.}
    \label{fig:linear_slip_forces}
    \vspace*{-0.5cm}
\end{figure}

The entire slip-to-stick event can occur within $0.05$ seconds, as shown in Fig. \ref{fig:linear_plot_2cm}. When the object transitions from sticking to slipping, the friction force shifts from static to Coulomb friction, resulting in rapid acceleration. As the rigid contact pad provides minimal damping compared to soft fingers, the object's sliding velocity and acceleration become highly sensitive to the grasp force and contact properties. This sensitivity can cause the object to overshoot the intended trajectory upon slipping, leading the linear slip controller into a stick-slip cycle. Fig. \ref{fig:linear_slip_forces} illustrates two extremes in surface material behavior. The spectacle case with synthetic leather shows relatively smooth velocity transitions, with only minor adjustments to the grasp force. In contrast, the plastic object quickly overshoots the trajectory, triggering pronounced stick-slip motion, which is reflected in the grasp forces. Notably, the plastic object exhibits a larger difference between the static and Coulomb friction coefficients compared to the spectacle case (see Tab. \ref{tab:friction_estimation}), contributing to the pronounced stick-slip behavior observed during the experiment.

Each experiment was conducted 10 times per object, with ground-truth displacement measured using a caliper. The statistical analysis of the results is presented in Tab. \ref{tab:linear_slipage}. The findings demonstrate that the proposed gripper, sensors, and controllers can accurately reposition objects through linear slippage. For plastic, cardboard and wood objects, the tracking error in the intended setting was comparable to the measured tracking error using the test rig described in \ref{sec:vel_sensors_results}.

The synthetic leather case was a notable outlier, as it on average overshot the target displacement significantly further then the other objects. As shown in Tab. \ref{tab:linear_slipage}, this overshoot can be attributed to tracking inaccuracies, where the sensors underestimated the actual distance traveled. However, the tracking accuracy for the spectacle case improved at slower velocities, such as 20 mm in 5 seconds. It should be noted that during the 40 mm in 2 seconds experiment, there was one instance where the spectacle case fell out of the gripper.

{
\renewcommand{\arraystretch}{1.2}
\begin{table}[pbb]
\scriptsize
\centering
\caption{Linear slippage, distance $x$ travelled, mean $\mu$ and standard deviation $\sigma$.\label{tab:linear_slipage}}
\resizebox{\columnwidth}{!}{%
\begin{tabular}{c|c|c|c|c}
\thickhline
Object & $x$ $(t=2)$  & $x-\hat{x}$ $(t=2)$ & $x$ $(t=5)$  & $x-\hat{x}$ $(t=5)$   \\ 
 &  ($\mu$, $\sigma$) & ($\mu$, $\sigma$)  &  ($\mu$, $\sigma$) & ($\mu$, $\sigma$)\\ \hline
\thickhline
Cardboard (20 mm) & (20.85, 0.97) & (1.24, 1.04) &  (20.61, 0.74) & (0.71, 0.74)  \\ \hline
Case (20 mm) & (23.46, 1.07) & (3.67, 1.08) & (21.30, 0.63) & (1.26, 0.71) \\ \hline
Plastic (20 mm) & (20.40, 0.92) & (0.27, 1.05) & (20.86, 0.69) & (0.51, 0.69) \\ \hline
Wood (20 mm) & (20.19, 0.71) & (0.42, 0.65) & (20.59, 0.61) & (0.58, 0.44) \\ \hline
\thickhline
Cardboard (40 mm) & (42.93, 1.83) & (3.36, 1.47) & (41.67, 1.70) & (0.98, 1.52)  \\ \hline
Case (40 mm) & (48.67, 5.68) & (8.27, 5.78) & (47.15, 4.17) & (6.93, 4.03) \\ \hline
Plastic (40 mm) & (41.05, 1.68) & (1.74, 1.62) & (42.35, 1.07) & (1.84, 0.73) \\ \hline
Wood (40 mm) & (39.44, 0.72) & (0.27, 0.83) & (40.61, 0.84) & (0.70, 0.57) \\ \hline
\end{tabular}
}
\end{table}
}

\subsection{Rotational Slip Control}\label{sec:rot_slip}

\begin{figure}
    \centering
    \smallskip 
    \includegraphics[width=0.9\columnwidth]{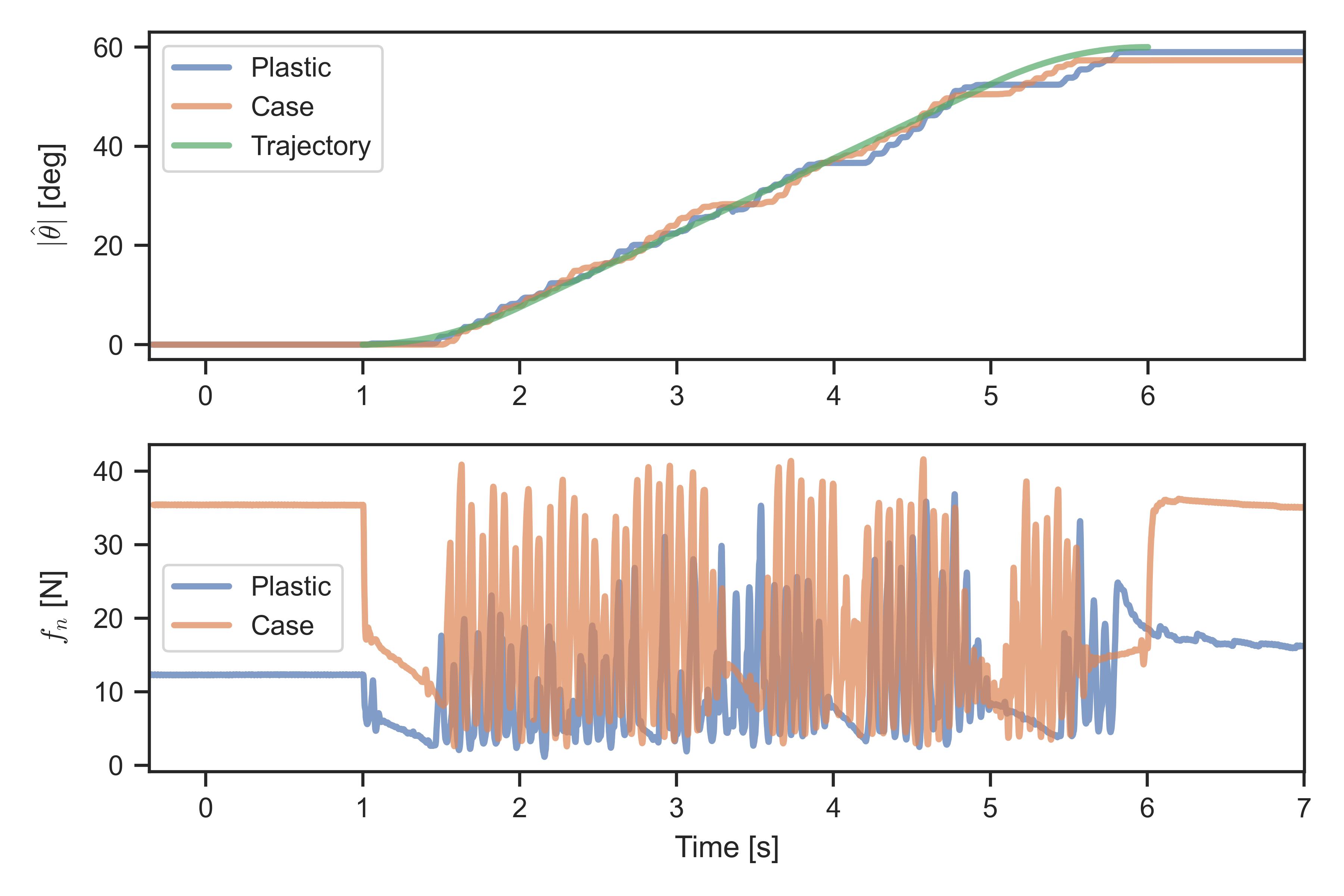}
    \caption{Highlights the grasp force and trajectory following for the plastic and case object.}
    \label{fig:rotation_forces}
    \vspace*{-0.5cm}
\end{figure}

The rotational slip controller leverages gravity to reorient the object by following specific trajectories, which were evaluated under different target displacements and velocities:
\begin{enumerate}
    \item \textbf{$45^\circ$ in 2 s:} Starting from $15^\circ$ off vertical. 
    \item \textbf{$45^\circ$ in 5 s:} Starting from $15^\circ$ off vertical. 
    \item \textbf{$60^\circ$ in 2 s:} Starting from $60^\circ$ off vertical.  
    \item \textbf{$60^\circ$ in 5 s:} Starting from $60^\circ$ off vertical. 
\end{enumerate}
\begin{figure}
    \centering
    \begin{subfigure}[b]{0.49\columnwidth}
        \includegraphics[width=\columnwidth]{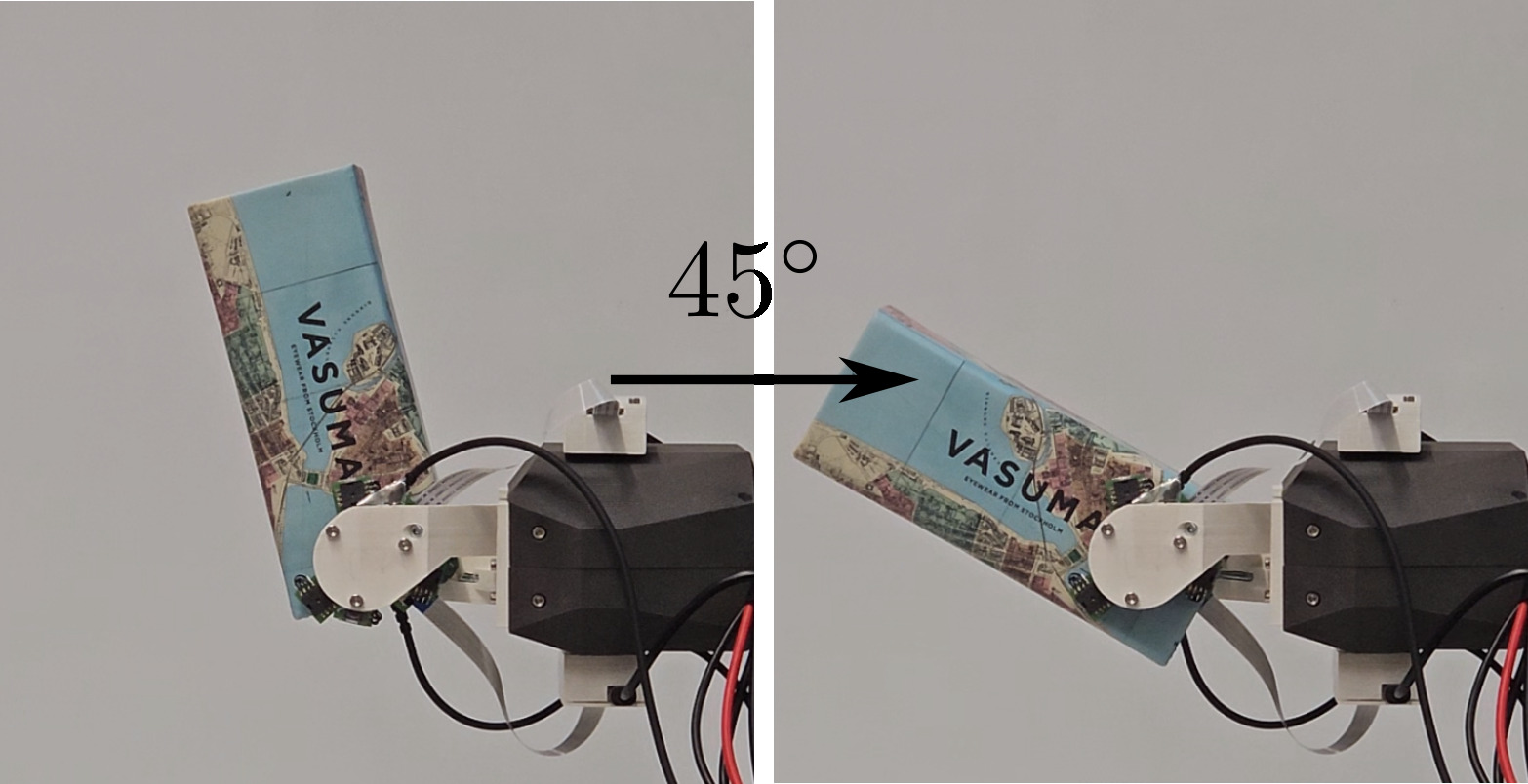}
        \caption{ $45^\circ$ rotation.}
        \label{fig:rot_slip_45}
    \end{subfigure}
    \begin{subfigure}[b]{0.49\columnwidth}
        \includegraphics[width=\columnwidth]{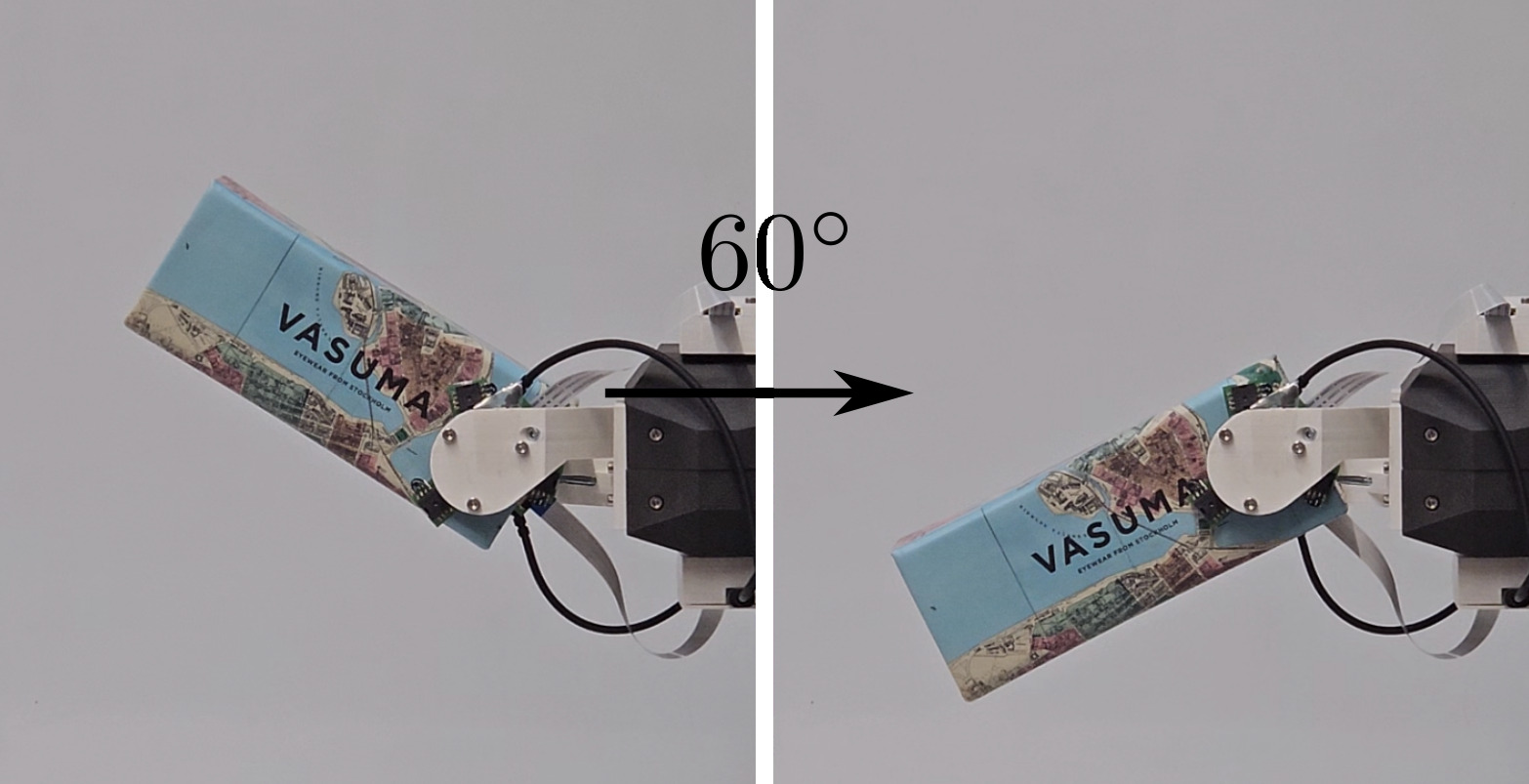}
        \caption{$60^\circ$ rotation.}
        \label{fig:rot_slip_60}
    \end{subfigure}
    \caption{Snapshots of rotational slippage experiment.} 
    \label{fig:rotation_slip}     
    \vspace*{-0.5cm}
\end{figure}
\begin{figure}
    \centering
    \smallskip 
    \includegraphics[width=0.9\columnwidth]{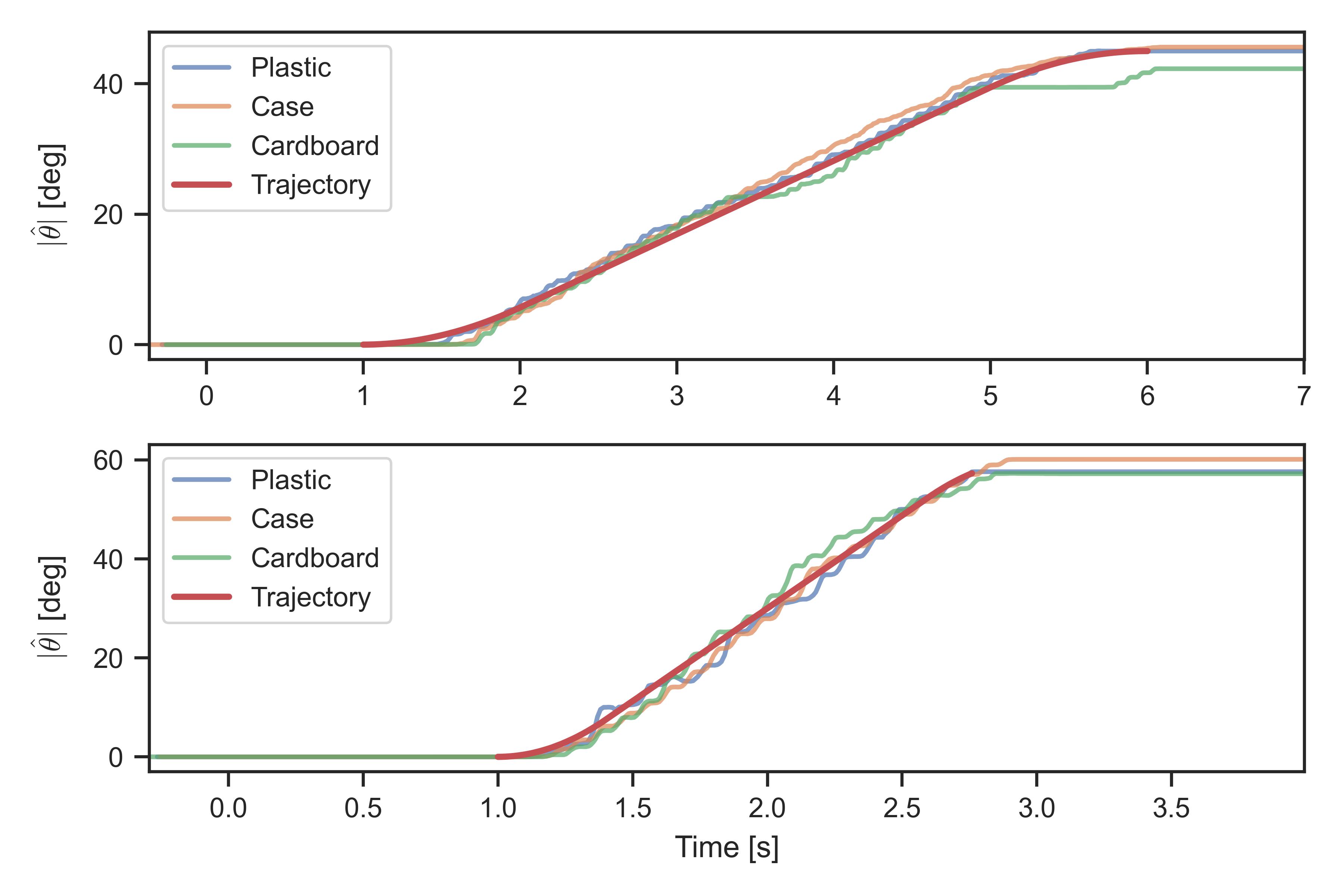}
    \caption{Rotational slippage for $45^\circ$, top plot shows a 5 second trajectory and bottom plot shows a 2 second trajectory. $|\hat{\theta}|$ is the estimated displacement by the planar velocity sensors.}
    \label{fig:rotation_plot_45}
    \vspace*{-0.5cm}
\end{figure}
The initial and final poses of the rotational slip experiments are shown in Figs. \ref{fig:rot_slip_45} and \ref{fig:rot_slip_60}. The internally measured rotational displacements compared to the desired trajectories are presented in Fig. \ref{fig:rotation_plot_45}. The trajectory following in the rotational slip experiments appears smoother compared to the linear slippage experiments; the slip-stick motion is less pronounced. This smoother response can be partially attributed to the limited control authority for generating torque. The grasp forces between the linear and rotational slippage experiments differ significantly, as seen by comparing the forces in Figs. \ref{fig:linear_slip_forces} and \ref{fig:rotation_forces}. Higher grasp forces cause more pronounced finger flexion, reducing the effective surface contact area, as illustrated in Figure \ref{fig:finger_flex}, which reduces torque. This finger flexion can result in a nonlinear torque response to the grasp force. However, it is worth noting that the grasp force oscillates more in the rotational slip experiments than in the linear ones. The parameters used for the rotational controller, listed in Tab. \ref{tab:slip_paramters}, were tuned for trajectory following rather than smooth grasp forces. Additionally, the wooden object saturated the rotational controller and was therefore not tested further.

{
\renewcommand{\arraystretch}{1.2}
\begin{table}[pbb]
\scriptsize
\centering
\caption{Rotational slippage, angle $\theta$ rotated, mean $\mu$ and standard deviation $\sigma$.  \label{tab:rotation_slippage}}
\resizebox{\columnwidth}{!}{%
\begin{tabular}{c|c|c|c|c}
\thickhline
Object & $\theta$ $(t=2)$  & $\theta-\hat{\theta}$ $(t=2)$ & $\theta$ $(t=5)$  & $\theta-\hat{\theta}$ $(t=5)$   \\ 
 &  ($\mu$, $\sigma$) & ($\mu$, $\sigma$)  &  ($\mu$, $\sigma$) & ($\mu$, $\sigma$)\\ \hline
\thickhline
Cardboard ($45^\circ$) & (54.0, 6.3) & (8.6, 5.9) & (52.9, 8.6) & (9.5, 8.4)  \\ \hline
Case ($45^\circ$) & (50.0, 6.4) & (3.7, 6.2) & (48.7, 4.0) & (3.7, 4.4) \\ \hline
Plastic ($45^\circ$) & (58.9, 6.3) & (13.4, 6.2) & (56.2, 4.5) & (12.6, 5.6) \\ \hline
\thickhline
Cardboard ($60^\circ$)  & (63.9, 3.9) & (5.3, 3.8) & (74.0, 12.7) & (16.5, 11.7)  \\ \hline
Case ($60^\circ$) & (65.1, 3.9) & (4.9, 4.1) & (63.9, 5.4) & (6.8, 7.2) \\ \hline
Plastic ($60^\circ$) & (70.8, 3.4) & (10.5, 3.8) & (74.9, 7.6) & (17.1, 7.0) \\ \hline
\end{tabular}
}
\end{table}
}

Each experiment was conducted 10 times for each object, with the ground truth rotation measured with a digital level meter. The statistical results are presented in Tab.  \ref{tab:rotation_slippage}. The data indicates that the rotational slippage controller consistently overshoots the target rotation on average, primarily due to tracking errors. According to the internal rotation estimates, the controller comes close to the target, but real-world measurements typically show an overshoot of approximately $10^\circ$. During linear slippage tests, the estimated displacement was closely aligned with real-world measurements. However, two factors contribute to the discrepancies in rotational slippage: higher grasp forces and a higher likelihood of multiple optical sensors being outside the object.

A higher grasp force increases finger flexion, which alters the distance between the sensor and the object surface. Optical mouse sensors are generally sensitive to vertical distance, potentially impacting accuracy. The authors attribute most of the tracking error to situations where multiple optical sensors are outside the object simultaneously. The current gripper design has limited space for accommodating object rotation without causing collisions, often leading to object placements where more than one sensor could be tracking outside of the object during the motion. When only one of three optical sensor accurately tracks velocity, displacement is likely to be underestimated.

\subsection{Hinge Control}
{
\renewcommand{\arraystretch}{1.2}
\begin{table}[pbb]
\scriptsize
\centering
\caption{Hinge mode, object change of orientation, mean $\mu$ and standard deviation $\sigma$.  \label{tab:hinge_mode}}
\begin{tabular}{c|c|c}
\thickhline
Object & $\mu$ [deg] & $\sigma$ \\ \hline
\thickhline
Cardboard  & 8.4 & 4.9 \\ \hline
Case  & 7.2 & 1.9 \\ \hline
Plastic  & 11.5 & 6.8 \\ \hline
Wood  & 4.8 & 1.5 \\ \hline
\end{tabular}
\end{table}
}

\begin{figure}
    \centering
    \smallskip 
    \includegraphics[width=0.9\columnwidth]{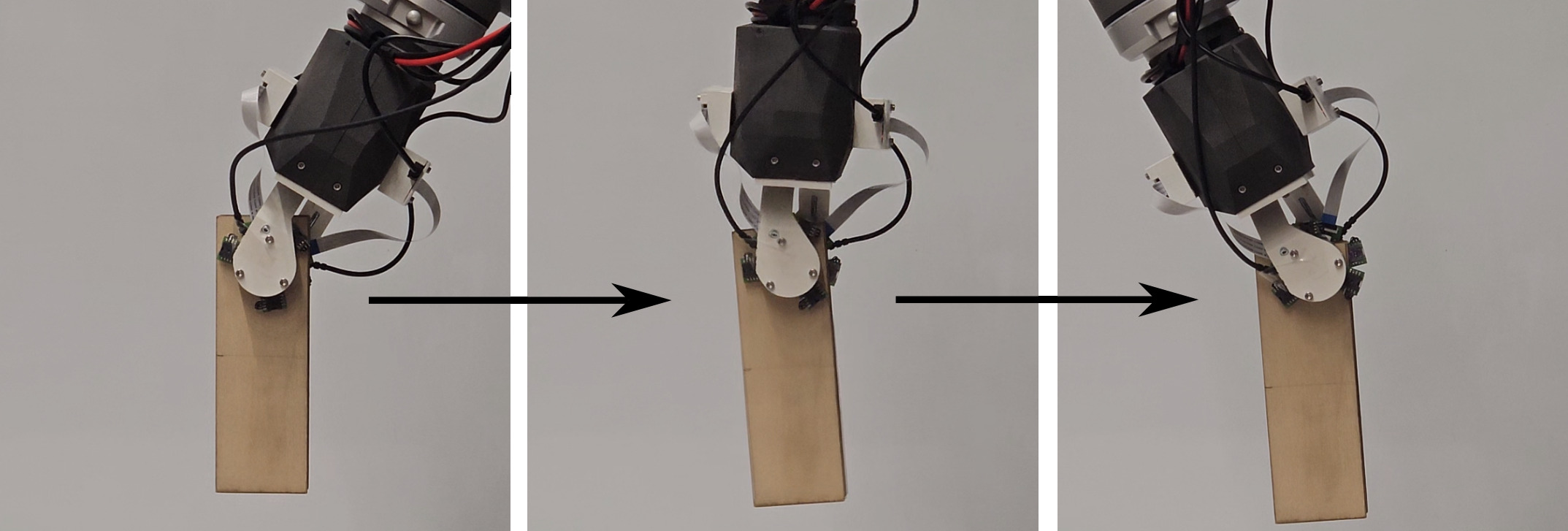}
    \caption{Hinge mode experiment.}
    \label{fig:hinge_mode}
    \vspace*{-0.5cm}
\end{figure}

The hinge controller enables the gripper to rotate while keeping the CoG of the object directly beneath the grasp point, as shown in Fig. \ref{fig:hinge_mode}. In the experiment, the gripper is initially rotated 30° away from the vertical axis, with the object hanging vertically from the gripper. The gripper, with hinge control activated, is then rotated 60° in the opposite direction, while the object is expected to remain in a vertical orientation.

The final orientation of the object relative to vertical is measured using a digital level meter, and the results are summarized in Tab. \ref{tab:hinge_mode}. These results demonstrate that the hinge controller effectively maintains the object's orientation as the gripper rotates. However, it should be noted that the cardboard object fell out of the gripper once during the trials, and that the robot introduced significant shaking during the rotational motion. 

\begin{figure}
    \centering
    \includegraphics[width=0.7\linewidth]{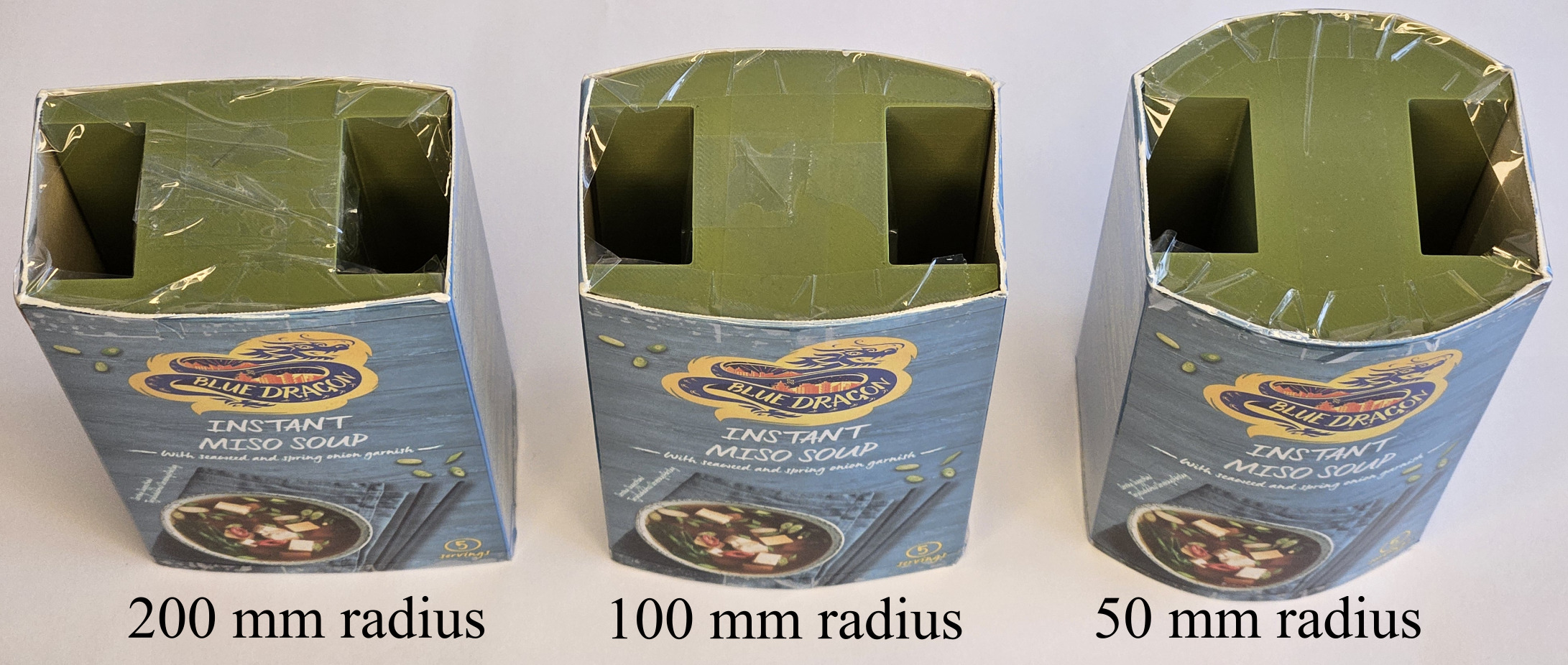}
    \caption{\textcolor{blue}{Plastic objects with curved surfaces.}}
    \label{fig:curved_surface}
    \vspace*{-0.6cm}
\end{figure}

\subsection{Testing the Limits} \label{sec:test_limits}

\begin{figure*} 
    \centering
    \smallskip 
    \includegraphics[width=1.0\textwidth]{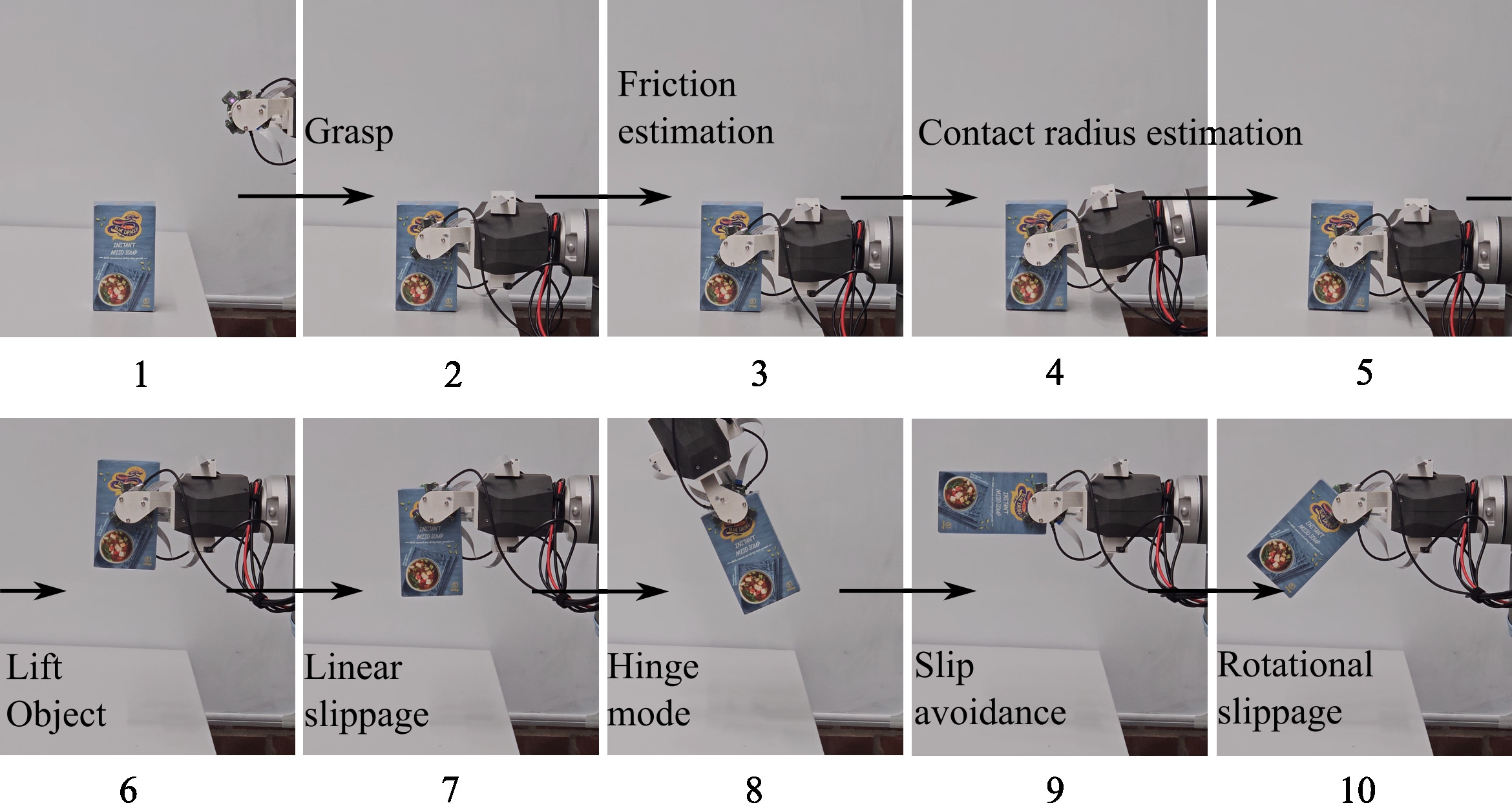}
    \caption{In-hand manipulation demonstration.}
    \label{fig:demo}
    \vspace*{-0.5cm}
\end{figure*}

\textcolor{blue}{Two edge cases are evaluated: (i) the maximum speed at which an object can be repositioned within the gripper, and (ii) the effect of surface curvature on performance. To simplify the analysis, only plastic objects are considered}

\textcolor{blue}{Case (i): the object is commanded to slide 20 mm over various target durations (see Tab. \ref{tab:test_lim}). For the target time of 1 s the performance is similar to section \ref{sec:linear_slip}. For target times of 0.5 s and 0.25 s, the controller overshoots the desired displacement by approximately 5 mm. However, for the shortest duration (0.1 s), the overshoot is reduced to 2.7 mm. This improvement is attributed to the absence of stick-slip effects, as the object's maximum acceleration and velocity are lower than the commanded values, resulting in no trajectory overshoot within the specified duration. Notably, the 0.1 s trajectory takes an average of 0.17 s to complete.}

\textcolor{blue}{Case (ii): While the planar velocity sensors are optimized for flat surfaces, this experiment evaluates performance degradation with increasing surface curvature. The tested objects, shown in Fig. \ref{fig:curved_surface}, all match the weight of the plastic object in Tab. \ref{tab:object_properites}. Each object is commanded to slide 20 mm in 2 s, with results summarized in Tab. \ref{tab:test_lim}. As curvature increases, the velocity tracking progressively underestimates the true velocity. This degradation is attributed to the sensor moving away from its optimal sensing distance as the surface curves. On the 200 mm radius surface, the object fell from the gripper in 1 out of 10 trials and exhibited significant overshoot. For the 100 mm and 50 mm radius surfaces, the object fell out in all trials, with estimated displacements decreasing further as curvature increased.}

\renewcommand{\arraystretch}{1.2}
\begin{table}[pbb]
\scriptsize
\centering
\caption{\textcolor{blue}{Testing the limits with plastic object for linear slippage of 20 mm with different trajectory times and surface curvatures. Distance $x$ traveled, mean $\mu$, standard deviation $\sigma$ and actual time $t$. A fail is considered if the object falls out of the gripper. } \label{tab:test_lim}}
\resizebox{\columnwidth}{!}{%
\begin{tabular}{c|c||c|c|c|c}
\thickhline
    Trajectory & x & Surface & $x$ & $\hat{x}$ & Fail   \\ 
    time & ($\mu$, $\sigma$, $t$) & radius  & ($\mu$, $\sigma$) & $\mu$ &  rate \\ \thickhline
    1 s  & (21.87, 1.80, 0.98) & 200 mm & (56.28, 7.81) & 19.66 & 10 \% \\ \hline    
  0.5 s  & (25.40, 4.60, 0.41) & 100 mm & (N/A)& 17.84 & 100 \% \\ \hline
  0.25 s & (25.96, 3.64, 0.24) & 50 mm & (N/A)& 13.39 & 100 \% \\ \hline
  0.1 s  & (22.70, 1.78, 0.17) & & &  & \\ \hline
\end{tabular}
}
\end{table}

\subsection{In-Hand Manipulation Demonstration} \label{sec:final_demo}

The final demonstration integrates all of the proposed methods into a single cohesive performance, as shown in Fig. \ref{fig:demo}. The sequence begins by grasping an object in a known pose, followed by the estimation of the contact properties. Once the object is securely grasped, the slip-avoidance controller is activated, and the object is lifted. Next, the linear slip controller is employed to move the object 30 mm downward over 3 seconds, after which the hinge controller is used to reorient the gripper relative to the object. The slip-avoidance controller then reorients both the gripper and the object, setting the stage for the rotational slip controller to rotate the object by $30^\circ$ within 3 seconds.

This demonstration was conducted ten times using the plastic object, and successfully performing all tasks 9 times without dropping the object, achieving a 90 \% success rate. The demonstration highlights the capabilities of the proposed system and illustrates how gravity-assisted slip control can endow parallel jaw grippers with the ability to manipulate an object and change the gripper-object configuration.

\section{Conclusions} \label{sec:conclusions}

In this paper, we have presented a comprehensive system for advanced in-hand slip control, featuring several key components: a parallel gripper that enables fast and precise force control, relative velocity sensors mounted on force-torque sensors that accurately measure both tangential and rotational velocities, and a fast contact property estimation process. Additionally, we introduced four slip-aware controllers: slip-avoidance, trajectory-following for linear and rotational slippage, and hinge control.
The gripper, equipped with these sensors, was mounted on a UR10 robot, and our system was rigorously tested across a wide range of scenarios. The results demonstrate that using solely in-hand sensing, our approach advances the state of the art for in-hand slip control.  

This study demonstrates the significant potential of the proposed hardware system.  However, our immediate focus will be on enhancing the sensing capabilities and improving the physical contact properties \textcolor{blue1}{to relax the constraint of parallel surfaces or a rigid contact pad, thereby enabling interaction with more complex object geometries.} Beyond this, the research could branch into various promising directions. For instance, future work could explore dynamic movements and manipulation, smoother slippage control, online friction estimation, extend the effective reach of manipulators, slip-aware multi-arm manipulation, manipulation based on external contacts, sensor fusion with external sensors, planning for slip-aware systems, and advancing human-robot interactions.

\begin{acks}
This work was funded by the Wallenberg AI, Autonomous Systems and Software Program (WASP) funded by the Knut and Alice Wallenberg Foundation.\\
Website: \url{https://wasp-sweden.org/}
\end{acks}
\theendnotes
\bibliographystyle{SageH}
\bibliography{references.bib}

\end{document}